\documentclass[twoside,11pt]{article}

\usepackage{jmlr2e}
\usepackage{hyperref}
\usepackage{url}
\usepackage{dsfont}
\usepackage{bm}
\usepackage{amsmath}
\usepackage{amssymb}
\usepackage{graphicx}
\usepackage{wrapfig}
\usepackage{caption}
\usepackage{algorithm2e}
\usepackage{multirow}
\usepackage{lscape}
\usepackage{rotating}

\jmlrheading{1}{2014}{1-48}{4/00}{10/00}{Daniel Hern\'andez-Lobato}

\ShortHeadings{Non-linear Causal Inference using Gaussianity Measures}{Hern\'andez-Lobato
\emph{et al.}}
\firstpageno{1}

\begin{document}

\title{Non-linear Causal Inference using Gaussianity Measures}

\author{%
  \name Daniel Hern\'andez-Lobato \email daniel.hernandez@uam.es \\
  \addr Universidad Aut\'onoma de Madrid\\
  Calle Francisco Tom\'as y Valiente 11,  \\
  Madrid, 28049, Spain 
  \AND
  \name Pablo Morales-Mombiela \email pablo.morales@estudiante.uam.es \\
  \addr Quantitative Risk Research\\
  Calle Faraday 7,  \\
  Madrid, 28049, Spain
  \AND
  \name David Lopez-Paz \email  david@lopezpaz.org\\
  \addr  Max Planck Institute for Intelligent Systems, T\"ubingen, Germany \\
  \addr  University of Cambridge, United Kingdom
  \AND
  \name Alberto Su\'arez \email alberto.suarez@uam.es \\
  \addr Universidad Aut\'onoma de Madrid\\
  Calle Francisco Tom\'as y Valiente 11, \\
  Madrid, 28049, Spain 
}

\editor{TBD}

\maketitle

\begin{abstract}%
We provide theoretical and empirical evidence for a type of
asymmetry between causes and effects that is present when these are 
related via linear models contaminated with additive non-Gaussian noise.
Assuming that the causes and the effects  have the same distribution, 
we show that the distribution of the residuals of a linear fit in the anti-causal 
direction is closer to a Gaussian than the distribution of the residuals 
in the causal direction. This Gaussianization effect is
characterized by reduction of the magnitude of the high-order cumulants 
and by an increment of the differential entropy of the residuals. 
The problem of non-linear causal inference 
is addressed by performing an embedding in an expanded feature space, in which 
the relation between causes and effects can be assumed to be linear. 
The effectiveness of a method to discriminate between causes and effects
based on this type of asymmetry is illustrated in a variety of experiments
using different measures of Gaussianity. The proposed method is shown to 
be competitive with state-of-the-art techniques for causal inference.
\end{abstract}


\section{Introduction}

The inference of causal relationships from data is one of the current areas of
interest in the artificial intelligence community, \emph{e.g.}  \citep{ChenZCS14,
Janzing12,morales13}. The reason for this surge of interest is 
that discovering the causal structure of a complex system provides an
explicit description of the mechanisms that generate the data, and allows us
to understand the consequences of interventions in the system
\citep{Pearl2000}. More precisely, automatic causal inference can be 
used to determine how modifications of the value of certain relevant
variables (the causes) influence the values of other related variables (the effects).
Therefore, understanding cause-effect relations is of paramount importance to
control the behavior of complex systems and has applications in industrial processes, 
medicine, genetics, economics, social sciences or meteorology.

Causal relations can be determined in complex systems in three different ways.
First, they can be inferred from domain knowledge provided by an expert, and
incorporated in an ad-hoc manner in the description of the system.
Second, they can be discovered by performing interventions in the system. These
are controlled experiments in which one or several variables of the system are
forced to take particular values. Interventions constitute a primary tool
for identifying causal relationships. However, in many situations they are 
unethical, expensive, or technically infeasible.
Third, they can be estimated using causal discovery algorithms that use as 
input purely uncontrolled and static data.

This last approach for causal discovery has recently received much
attention from the machine learning community \citep{Shimizu06,Hoyeretal08,zhang2009}. 
These  methods assume a particular model for the \emph{mapping mechanisms} 
that link causes to effects. 
By specifying particular conditions on the mapping mechanism and
the distributions of the cause and noise variables, the causal direction becomes
identifiable \citet{ChenZCS14}.
For instance, \citet{Hoyeretal08} assume that the effect is a 
non-linear transformation of the cause plus some independent additive
noise. A potential drawback of these methods is that the
assumptions made by the particular model considered 
could be unrealistic for the data under study.

In this paper we propose a general method for causal inference that belongs to
the third of the categories described above. Specifically, we assume that
the cause and the effect variables have the same distribution and 
are linked by a linear relationship contaminated with non-Gaussian noise. 
For the univariate case we prove that, under these assumptions, the magnitude of
the cumulants of the residuals of order higher than two is smaller 
for the linear fit in the anti-causal direction than in the causal one. 
Since the Gaussian is the only distribution whose cumulants of 
order higher than 2 are zero, statistical tests based 
on measures of Gaussianity can be used for causal inference.
An antecedent of this result is the observation 
that, when cause and effect have the same distribution,
the residuals of a fit in the anti-causal direction have 
higher entropy than in the causal direction \citep{HyvarinenS13,KpotufeSJS14}. 
Since the residuals of the causal and anti-causal
linear models have the same variance and the 
Gaussian is the distribution that maximizes the entropy 
for a fixed variance, this means that the distribution of the latter
is more Gaussian than the former.  

For multivariate cause-effect pairs that have the same distribution and
are related by a linear model with additive non-Gaussian noise
the proof given by \citep{HyvarinenS13,KpotufeSJS14} 
can be extended to show that the entropy of the vector of residuals of a linear fit 
in the anti-causal direction is larger than the corresponding residuals of a  
linear fit in the causal direction.
We conjecture that also in this case there is a reduction of the magnitude of 
the tensor cumulants of the anti-causal multivariate residuals 
and provide some numerical evidence of this effect in two dimensions. 

The problem of non-linear causal inference is addressed 
by embedding the original problem in an expanded feature space. 
We then make  the assumption that the non-linear relation between 
causes and effects in the original space is linear in the expanded feature  space. 
The computations required to make inference on the causal direction 
based on this embedding can be readily carried out using kernel methods.

In summary, the proposed method for causal inference
proceeds by first making a transformation of the original variables
so that causes and effects have the same distribution.
Then we perform kernel ridge regression in both the causal 
and the anti-causal directions. The dependence between 
causes and effects, which is non-linear in the original space, is 
assumed to be linear in the kernel-induced feature space.
A statistical test is then used to quantify the degree 
of similarity between the distributions of these residuals and 
a Gaussian distribution with the same variance. 
Finally, the direction in which the residuals are 
less Gaussian is identified as the causal one. 

The performance of this method 
is evaluated in both synthetic and real-world cause-effect pairs.
From the results obtained it is apparent that the 
anti-causal residuals of a linear fit in the expanded feature 
space are more Gaussian than the causal residuals.
In general, it is difficult to estimate the entropy from a finite sample
\citep{beirlant++_1997_nonparametric}. Empirical estimators 
of high order cumulants involve high 
order moments, which means they often have large variance.
As an alternative, we propose to use 
statistical tests based on the \emph{energy distance}
to characterize the Gaussianization effect for the residuals of 
linear fits in the causal and anti-causal directions. Tests
based on the energy distance were analyzed in depth in \citep{Szekely2005}.
They have been shown to be related to homogeneity tests based on 
embeddings in a Reproducing Kernel Hilbert Space in \citep{Gretton2012}. 
An advantage of energy distance-based statistics is that
they can be readily estimated from a sample by computing
expectations of pairwise Euclidean distances. 
The energy distance generally provides better 
results than the entropy or cumulant-based Gaussianity measures. 
In the problems investigated, the accuracy of
the proposed method, using the energy distance to the Gaussian,
is comparable to other state-of-the-art 
techniques for causal discovery.

The rest of the paper is organized as follows: Section \ref{sec:asymmetry}
illustrates that, under certain conditions, the residuals of a linear regression 
fit are closer to a Gaussian in the anti-causal direction than in the causal one, 
based on a reduction of the high-order cumulants and on an increment of the entropy. 
This section considers both the univariate and multivariate cases. 
Section \ref{sec:feature_expansion} adopts a kernel approach to carry out a
feature expansion that can be used to detect non-linear causal relationships.
We also show here how to compute the residuals in the expanded feature 
space, and how to choose the different hyper-parameters of the proposed method.
Section \ref{sec:algorithm} contains a detailed description of the implementation.
In section \ref{sec:experiments} we present the results of an empirical assessment 
of the proposed method in both synthetic and real-world cause-effect data pairs. 
Finally, Section \ref{sec:conclusions} summarizes the conclusions 
and puts forth some ideas for future research.

\section{Asymmetry based on the non-Gaussianity of the residuals of linear models}
\label{sec:asymmetry}

Let $\mathcal{X}$ and $\mathcal{Y}$ be two random variables that are causally related. The direction of the causal relation is not known. 
Our goal is to determine whether $\mathcal{X}$ causes 
$\mathcal{Y}$, \emph{i.e.}, $\mathcal{X} \rightarrow \mathcal{Y}$ or, 
alternatively $\mathcal{Y}$ causes $\mathcal{X}$, \emph{i.e.},
$\mathcal{Y} \rightarrow \mathcal{X}$. 
For this purpose, we exploit an asymmetry between causes 
an effects. This type of asymmetry can be uncovered using 
statistical tests that measure the non-Gaussianity of 
the residuals of linear regression models obtained from 
fits in the causal and in the anti-causal direction. 

To motivate the methodology that we have developed, we will proceed in stepwise 
manner. First we analyze a special case in one dimension: 
We assume that $\mathcal{X}$ and  $\mathcal{Y}$  have the same distribution 
and are related via a linear model contaminated with  additive i.i.d. non-Gaussian noise.  
The noise is independent of the cause.  
Under these assumptions we  show that the distribution of the residuals of a 
linear fit in  the incorrect (anti-causal) direction is closer to 
a Gaussian distribution than the distribution of the residuals in 
the correct (causal) direction. 
For this, we use an argument based on the reduction of the magnitude of the
cumulants of order higher than 2. The cumulants are defined as the 
derivatives of  the logarithm of the moment-generating function evaluated at zero 
\citep{cornish+fisher_1938_moments,mccullagh87}. 

The Gaussianization effect can be characterized also in terms of an 
increase of the entropy. The proof is based on the results of \citep{HyvarinenS13}, 
which are extended in this paper to the multivariate case.
In particular, we show that the entropy of the 
residuals of a linear fit in the anti-causal direction is larger or equal than 
the entropy of the residuals of a linear fit in the causal direction. 
Since the Gaussian it the distribution that has maximum entropy, 
given a particular covariance matrix, an increase of the entropy of 
the residuals means that their distribution becomes closer to the Gaussian.

Finally, we note that it is easy to guarantee that $\mathcal{X}$ and $\mathcal{Y}$  
have the same distribution in the case that these variables are unidimensional 
and continuous. To this end we only have to transform one of
the variables (typically the cause random variable) using the probability 
integral transform, as described in Section \ref{sec:algorithm}. 
However, after the data have been transformed, the relation 
between the variables will no longer be linear in general. 
Thus, to address  non-linear cause-effect problems involving
 univariate random variables the linear model 
is formulated in an expanded feature space, where 
the multivariate analysis of the Gaussianization effect 
is also applicable.  In this feature space all the  
computations required for causal inference
can be formulated in terms of kernels. This can be used 
to detect non-linear causal relations in the original input space 
and allows for an efficient implementation of the method. 
The only assumption is that the non-linear relation in the original 
input space is linear in the expanded feature space induced by 
the selected kernel.

\subsection{Analysis of the univariate case based on cumulants}
\label{sec:asym_univa}

Let $\mathcal{X}$ and $\mathcal{Y}$ be one-dimensional  
random variables that have the same distribution. Without further 
loss of generality, we will assume that they have zero mean and unit variance.  Let $\mathbf{x}=(x_1,\ldots,x_N)^\text{T}$ and $\mathbf{y} =
(y_1,\ldots,y_N)^\text{T}$ be $N$ paired samples drawn i.i.d. from
$P(\mathcal{X},\mathcal{Y})$. Assume that the 
causal direction is $\mathcal{X} \rightarrow \mathcal{Y}$ and that the measurements  are related  by a linear model
\begin{align}
y_i & = w  x_i + \epsilon_i\,,\quad \epsilon_i \, \bot \, x_i\,, \quad \forall
i\,,
\label{eq:model_one_d}
\end{align}
where $w=\text{corr}(\mathcal{X},\mathcal{Y}) \in [-1,1]$  and $\epsilon_i$ is independent i.d. non-Gaussian additive noise.

A linear model in the opposite direction, \emph{i.e.}, $\mathcal{Y}\rightarrow \mathcal{X}$, can be built using least squares  
\begin{equation}
x_i = w y_i  + \tilde{\epsilon}_i,
\end{equation}
where $w = \text{corr}(\mathcal{Y},\mathcal{X})$ is the same coefficient
as in the previous model. The residuals of this reversed linear model are defined as $\tilde{\epsilon}_i = x_i - w y_i $. 

\begin{figure}[tb]
\begin{center}
\begin{tabular}{cc}
\includegraphics[width=0.49\textwidth]{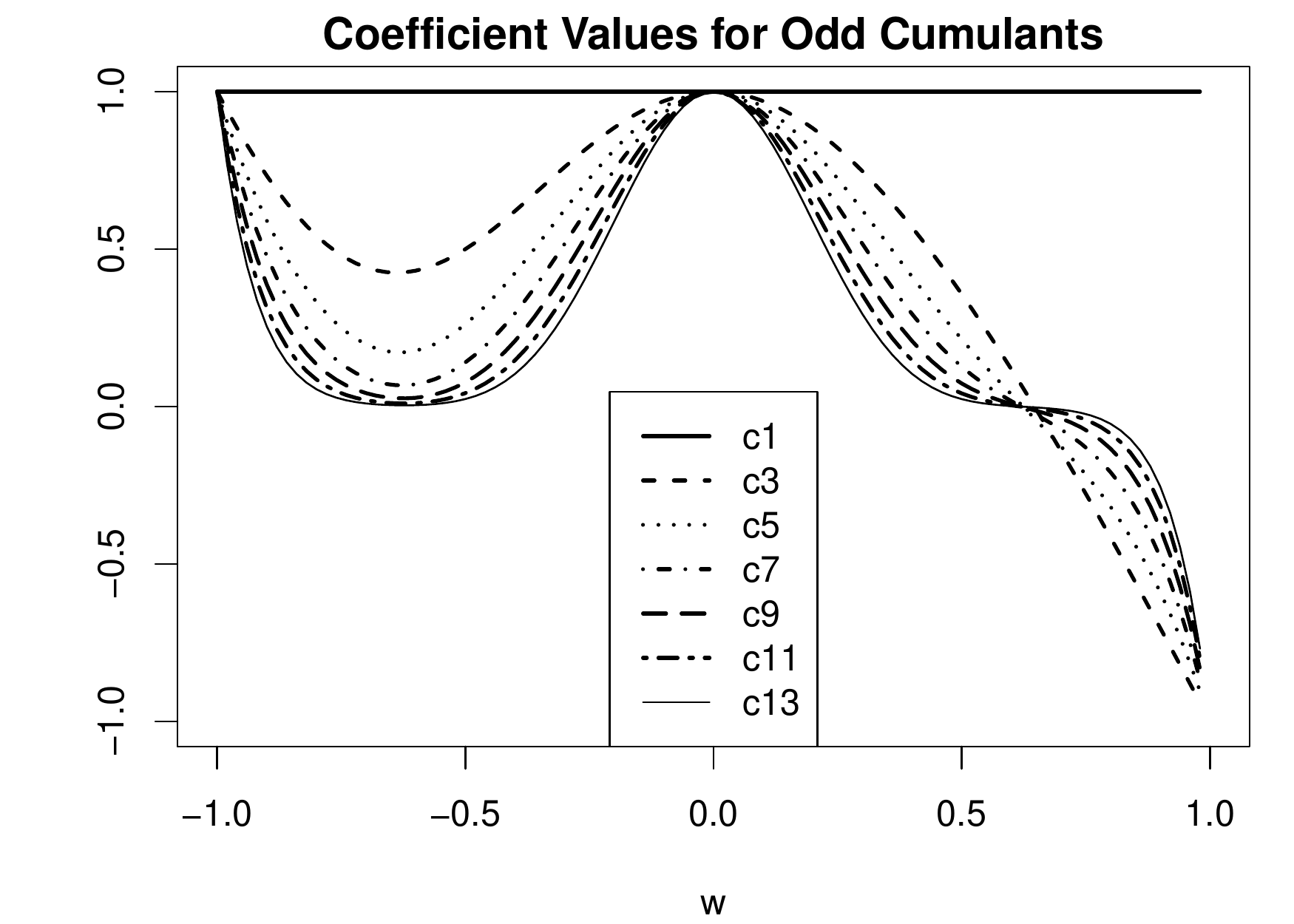} &
\includegraphics[width=0.49\textwidth]{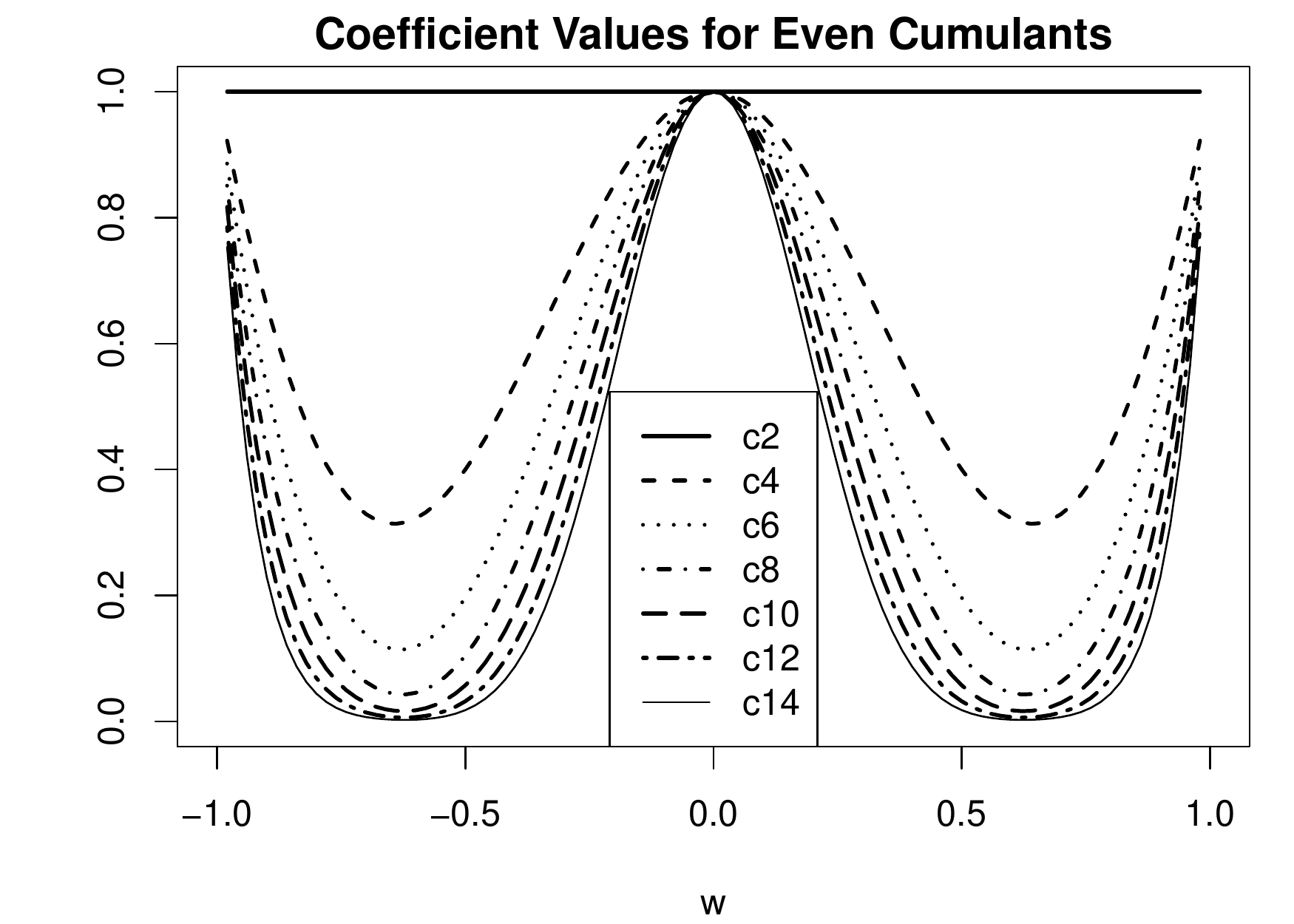} 
\end{tabular}
\end{center}
\caption{Values of the function $c_n(\cdot)$ as a function of $w$ for each 
	cumulant number $n$ (odd in the left plot, even in the right plot).
	All values for $c_n(\cdot)$ lie in the interval $[-1,1]$.}
\label{fig:cumulants}
\end{figure}

Following an argument similar to that of \cite{hernandezLobato11} we show that
the residuals $\{\tilde{\epsilon}_i\}_{i=1}^N$ in the anti-causal direction are more Gaussian than 
the residuals $\{\epsilon_i\}_{i=1}^N$ in the actual causal direction $\mathcal{X}
\rightarrow \mathcal{Y}$ based on a reduction of the magnitude of the cumulants. The proof is based 
on establishing a relation between  the cumulants 
of the distribution of the residuals in both the causal 
and the anti-causal direction.  
First, we show that $\kappa_n(y_i) $, 
the $n$-th order cumulant of $\mathcal{Y}$, can be
expressed in terms of $\kappa_n(\epsilon_i)$, 
the $n$-th order cumulant of the residuals:
\begin{align}
\kappa_n(y_i) & = w^n \kappa_n(x_i) + \kappa_n(\epsilon_i) 
 = w^n \kappa_n(y_i) + \kappa_n(\epsilon_i)
 = \frac{1}{1-w^n} \kappa_n(\epsilon_i)\,.
\label{eq:rel_one_d}
\end{align}
To derive this relation we have used (\ref{eq:model_one_d}), 
that $x_i$ and $y_i$ have the same distribution (and hence have the same cumulants),
and standard properties of cumulants \citep{cornish+fisher_1938_moments,mccullagh87}. 
Furthermore,
\begin{align}
\kappa_n(\tilde{\epsilon}_i) & = \kappa_n(x_i - w y_i) 
= \kappa_n(x_i - w^2 x_i - w \epsilon_i) 
= (1 - w^2)^n \kappa_n(x_i) + (- w)^n \kappa_n(\epsilon_i) \nonumber \\
& = (1 - w^2)^n \kappa_n(y_i) + (- w)^n \kappa_n(\epsilon_i) 
= \frac{(1 - w^2)^n}{1-w^n} \kappa_n(\epsilon_i) + (- w)^n \kappa_n(\epsilon_i) 
\nonumber \\
& = c_n(w) \kappa_n(\epsilon_i) \,,
\label{eq:formula_1d}
\end{align}
where we  have used the definition of $\tilde{\epsilon_i}$ and (\ref{eq:rel_one_d})
In Figure \ref{fig:cumulants} the value of
\begin{equation}
c_n(w) = \frac{(1 - w^2)^n}{1-w^n} + (- w)^n.
\end{equation}
is displayed as a function of $w \in [-1,1]$.
Note that $c_1(w) = c_2(w) = 1$ independently of the value of $w$. This
means that the mean and the variance of the residuals are the same
in both the causal and anti-causal directions. 
For  $n>2$, $|c_n(w)| \le 1$ with equality only for $w = 0 $ and $w = \pm 1$.
The result is that the high-order cumulants of the residuals 
in the anti-causal direction are
smaller in magnitude that the corresponding cumulants in the causal direction. 
Using the observation that all the cumulants of the Gaussian distribution of 
order higher than two are zero \citep{marcinkiewicz_1938_propriete},
we conclude that the 
distribution of the residuals  in the anti-causal direction 
is closer to the Gaussian distribution than in the causal direction. 

In summary, we can infer the causal direction by (i) fitting a linear model
in each possible direction, \emph{i.e.}, $\mathcal{X} \rightarrow \mathcal{Y}$ and $\mathcal{Y}\rightarrow\mathcal{X}$,
and (ii) carrying out statistical tests to detect the level of Gaussianity of the two corresponding 
residuals. The direction in which the residuals are less Gaussian is expected to be the correct one.

\subsection{Analysis of the multivariate case based on cumulants}
\label{sec:mul-cum}

In this section we argue that the Gaussianization effect of the residuals in
the anti-causal direction also takes place when the two random variables
$\mathcal{X}$ and $\mathcal{Y}$ are multidimensional. We
will assume that these variables follow the same distribution
and, without further loss of generality,
that they have been whitened (\emph{i.e.}, they have a zero
mean vector and the identity matrix as the covariance matrix). 
Let  $\mathbf{X}=(\mathbf{x}_1,\ldots,\mathbf{x}_N)^\text{T}$ and $\mathbf{Y} =
(\mathbf{y}_1,\ldots,\mathbf{y}_N)^\text{T}$ be $N$ paired samples drawn i.i.d.
from $P(\mathcal{X}, \mathcal{Y})$.  In this case, the model assumed for the
actual causal relation is
\begin{align}
\mathbf{y}_i = \mathbf{A} \mathbf{x}_i + \bm{\epsilon}_i \,, \quad
\bm{\epsilon}_i \,\bot\, \mathbf{x}_i\,,\quad \forall i\,,
\label{eq:model_multi}
\end{align}
where $\mathbf{A}=\text{corr}(\mathcal{Y},\mathcal{X})$ is a $d \times d$ matrix
of model coefficients and $\bm{\epsilon}_i$ is i.i.d. non-Gaussian additive noise. The model in the 
anti-causal direction is the one that results from the least squares fit:
\begin{align}
\mathbf{x}_i &= \tilde{\mathbf{A}} \mathbf{y}_i + \tilde{\bm{\epsilon}}_i\,,
\end{align}
where we have defined $\tilde{\bm{\epsilon}}_i=\mathbf{x}_i -
\tilde{\mathbf{A}} \mathbf{y}_i$ and $\tilde{\mathbf{A}} =
\text{corr}(\mathcal{X},\mathcal{Y})=\mathbf{A}^\text{T}$.

As in the univariate case, we start by expressing the cumulants of
$\mathcal{Y}$ in terms of the cumulants of the residuals. However,  
the cumulants are now tensors \citep{mccullagh87}:
\begin{align}
\kappa_n(\mathbf{y}_i) &= \kappa_n(\mathbf{A} \mathbf{x}_i) +
\kappa_n(\bm{\epsilon}_i)\,.
\end{align}
In what follows, the notation $\text{vect}(\cdot)$ stands for the vectorization
of a tensor.  For example, in the case of a tensor $\mathbf{T}$ with dimensions $d
\times d \times d$
$$\text{vect}(\mathbf{T})=(T_{1,1,1},T_{2,1,1},\cdots,T_{d,1,1},
T_{1,2,1},\cdots,T_{d,d,d})^\text{T}.$$ 
Using this notation we obtain
\begin{align}
\text{vect}(\kappa_n(\mathbf{y}_i)) &= \mathbf{A}^n
\text{vect}(\kappa_n(\mathbf{x}_i)) + \text{vect}(\kappa_n(\bm{\epsilon})) =
(\mathbf{I} - \mathbf{A}^n)^{-1}\text{vect}(\kappa_n(\bm{\epsilon}))\,,
\label{eq:cum_mul_rel}
\end{align}
where $\mathbf{A}^n=\mathbf{A} \otimes \mathbf{A} \otimes \mathbf{A} \cdots \otimes \mathbf{A}$, $n$ times, 
is computed using the Kronecker matrix product. To derive this 
expression we have used (\ref{eq:model_multi}), 
the fact that $\mathcal{Y}$ and $\mathcal{X}$ are equally distributed 
and hence have the same cumulants. We also have used
the properties of the tensor cumulants  
$\text{vect}(\kappa_n(\mathbf{A}\mathbf{x}_i))=\mathbf{A}^n\text{vect}(\kappa_n(\mathbf{x}_i))$, where 
the powers of the matrix $\mathbf{A}$ are computed using the Kronecker product \citep{mccullagh87}. 

Similarly, for the reversed linear model
\begin{align}
\kappa_n(\tilde{\bm{\epsilon}}_i) &= \kappa_n(\mathbf{x}_i -
\mathbf{A}^\text{T} \mathbf{y}_i)
   = \kappa_n(\mathbf{x}_i - \mathbf{A}^\text{T} \mathbf{A} \mathbf{x}_i -
   \mathbf{A}^\text{T} \bm{\epsilon}_i)
   = \kappa_n((\mathbf{I} - \mathbf{A}^\text{T} \mathbf{A}) \mathbf{x}_i -
   \mathbf{A}^\text{T} \bm{\epsilon}_i)\nonumber \\
   &= \kappa_n((\mathbf{I} - \mathbf{A}^\text{T} \mathbf{A}) \mathbf{x}_i) +
   \kappa_n(-\mathbf{A}^\text{T} \bm{\epsilon}_i)\,.
\end{align}
Using again the notation for the vectorized tensor cumulants 
\begin{align}
\text{vect}(\kappa_n(\tilde{\bm{\epsilon}}_i)) &= (\mathbf{I} -
\mathbf{A}^\text{T}\mathbf{A})^n \text{vect}(\kappa_n(\mathbf{x}_i)) + (-1)^n
(\mathbf{A}^\text{T})^n \text{vect}(\kappa_n(\bm{\epsilon}_i))\nonumber \\
&= (\mathbf{I} - \mathbf{A}^\text{T} \mathbf{A})^n (\mathbf{I} -
\mathbf{A}^n)^{-1} \text{vect}(\kappa_n(\bm{\epsilon}_i)) + (-1)^n
(\mathbf{A}^\text{T})^n \text{vect}(\kappa_n(\bm{\epsilon}_i))\nonumber \\
&= \left((\mathbf{I} - \mathbf{A}^\text{T} \mathbf{A})^n (\mathbf{I} -
\mathbf{A}^n)^{-1} + (-1)^n (\mathbf{A}^\text{T})^n
\right)\text{vect}(\kappa_n(\bm{\epsilon}_i))\,,
\label{eq:relation_multi}
\end{align}
where the powers of matrices are computed using the Kronecker product as well, and where we have
used (\ref{eq:cum_mul_rel}) and that $\mathcal{Y}$ and $\mathcal{X}$ are equally distributed and have the same cumulants.

We now give some evidence to support that the magnitude of
$\text{vect}(\kappa_n(\tilde{\bm{\epsilon}}_i))$ is smaller than the magnitude
of $\text{vect}(\kappa_n(\bm{\epsilon}_i))$ in terms of the $\ell_2$-norm, for cumulants of 
order higher than 2. That is, the tensors corresponding to high-order cumulants become
closer to a tensor with all its components equal to zero. For this, we introduce the following definition:

\begin{definition}
The operator norm of a matrix  $\mathbf{M}$ induced by the $\ell_p$ vector norm 
is $||\mathbf{M}||_\text{op} = \text{min}\{c \geq 0:\,
||\mathbf{M}\mathbf{v}||_{p} \leq c ||\mathbf{v}||_{p}\,, \forall
\mathbf{v}\}$, where $||\cdot||_p$ denotes the $\ell_p$-norm for vectors.
\end{definition}

The consequence is that $||\mathbf{M}||_\text{op} \geq ||\mathbf{M}\mathbf{v}||_{p} / ||\mathbf{v}||_{p}$,
 $\forall \mathbf{v}$. This means that $||\mathbf{M}||_p$ can be understood as 
a measure of the size of the matrix $\mathbf{M}$. 
In the case of the $\ell_2$-norm, the operator norm of a matrix
$\mathbf{M}$ is equal to its largest singular value or, equivalently, to the
square root of the largest eigenvalue of $\mathbf{M}^\text{T}\mathbf{M}$.
Let $\mathbf{M}_n=(\mathbf{I} - \mathbf{A}^\text{T} \mathbf{A})^n (\mathbf{I} -
\mathbf{A}^n)^{-1} + (-1)^n (\mathbf{A}^\text{T})^n$. That is, $\mathbf{M}_n$
is the matrix that relates the cumulants of order $n$ of the residuals  
in the causal and  anti-causal directions in (\ref{eq:relation_multi}). 
We now evaluate 
$||\mathbf{M}_n||_\text{op}$, and show that in most cases
its value is smaller
than one for high-order cumulants $\kappa_n(\cdot)$, 
leading to a Gaussianization of the residuals 
in the anti-causal direction.  From (\ref{eq:relation_multi}) and the definition given above, 
we know that $||\mathbf{M}_n||_\text{op}\geq ||\text{vect}(\kappa_n(\tilde{\bm{\epsilon}}_i))||_2 / ||\text{vect}(\kappa_n(\bm{\epsilon}_i))||_2$.
This means that if $||\mathbf{M}_n||_\text{op}<1$ the cumulants of the residuals in the incorrect 
causal direction are shrunk to the origin. 
Because the multivariate Gaussian 
distribution has all cumulants of order higher than two equal to zero 
\citep{mccullagh87}, this translates into a distribution for the residuals in the anti-causal 
direction that is closer to the Gaussian distribution.

In the causal direction, we have that
$\mathds{E}[\mathbf{y}\mathbf{y}^T]= \mathds{E}[(\mathbf{A}\mathbf{x}_i +
\bm{\epsilon}_i)(\mathbf{A}\mathbf{x}_i + \bm{\epsilon}_i)^\text{T}]=
\mathbf{A}\mathbf{A}^T + \mathbf{C} = \mathbf{I}$, where $\mathbf{C}$ is the
positive definite covariance matrix of the actual residuals\footnote{We assume
such matrix exists and that it is positive definite.}.  Thus,
$\mathbf{A}\mathbf{A}^T = \mathbf{I} - \mathbf{C}$ and hence the singular
values of $\mathbf{A}$, denoted $\sigma_1,\ldots,\sigma_d$, satisfy $0\leq
\sigma_i = \sqrt{1 - \alpha_i} \leq 1$, where $\alpha_i$ is the corresponding
positive eigenvalue of $\mathbf{C}$.  Assume that $\mathbf{A}$ is symmetric
(this also means that $\mathbf{M}_n$ is symmetric).  Denote by
$\lambda_1,\ldots,\lambda_d$ to the eigenvalues of $\mathbf{A}$.  That is,
$\sigma_i = \sqrt{\lambda_i^2}$ and $0\leq \lambda_i^2 \leq 1$, $i=1,\ldots,d$.
For a fixed cumulant of order $n$ we have that
\begin{align}
||\mathbf{M}_n||_\text{op} & = 
\underset{\mathbf{v} \in \mathcal{S}}{\text{max}} 
\quad 
\left|
\prod_{j=1}^n \frac{\left[1 -  \lambda_{v_j}^2\right]}{1 -  \prod_{i=1}^n
\lambda_{v_i}} + (-1)^n
\prod_{j=1}^n \lambda_{v_j}
\right|
\,,
\end{align}
where $\mathcal{S}=\{1,\ldots,d\}^n$, $|\cdot|$ denotes absolute value, and we
have employed standard properties of the Kronecker product about eigenvalues
and eigenvectors \citep{Laub2004}. Note that this expression does not depend on
the eigenvectors of $\mathbf{A}$, but only on its eigenvalues.

\begin{figure}[tb]
\begin{center}
\begin{tabular}{ccc}
\includegraphics[width=0.31\textwidth]{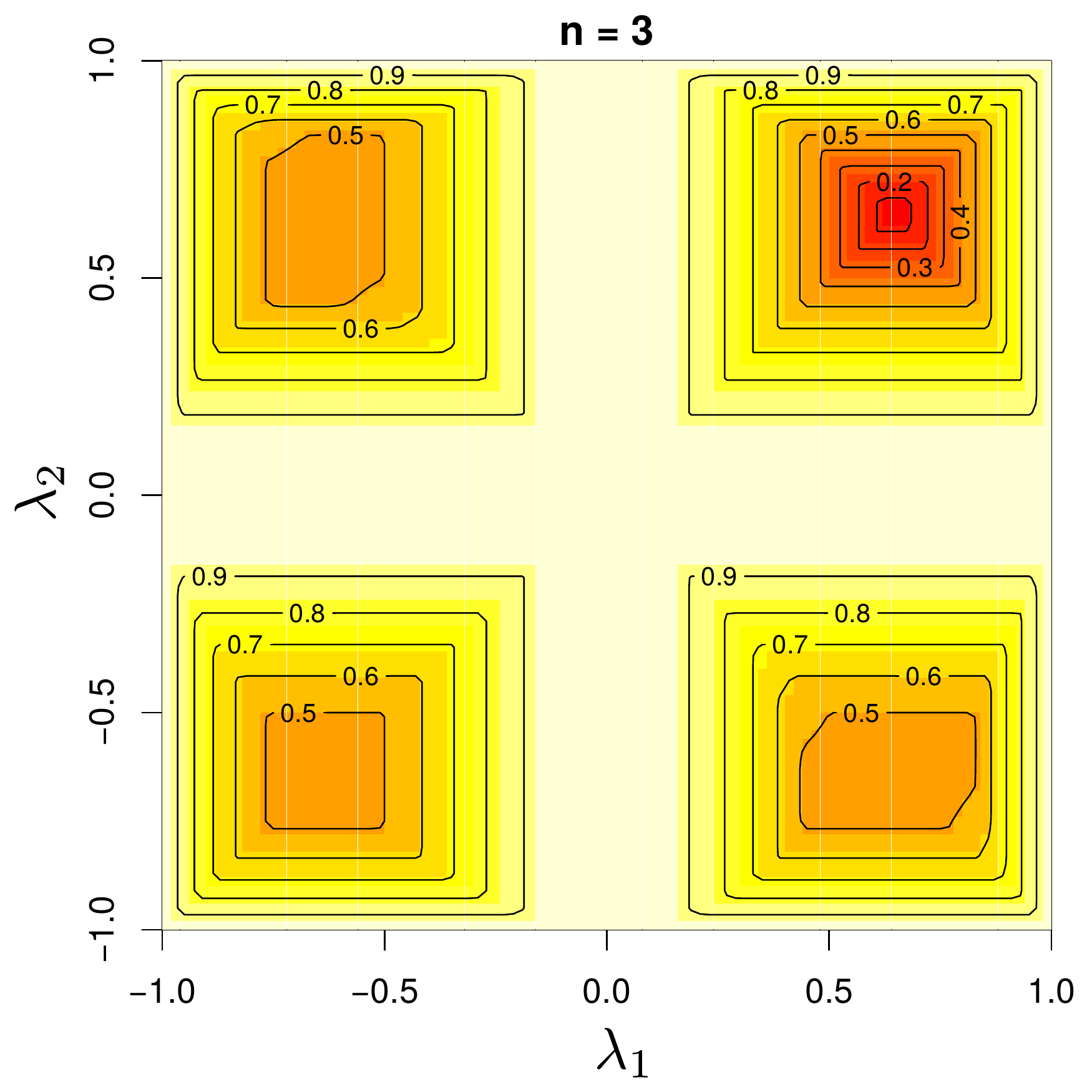} &
\includegraphics[width=0.31\textwidth]{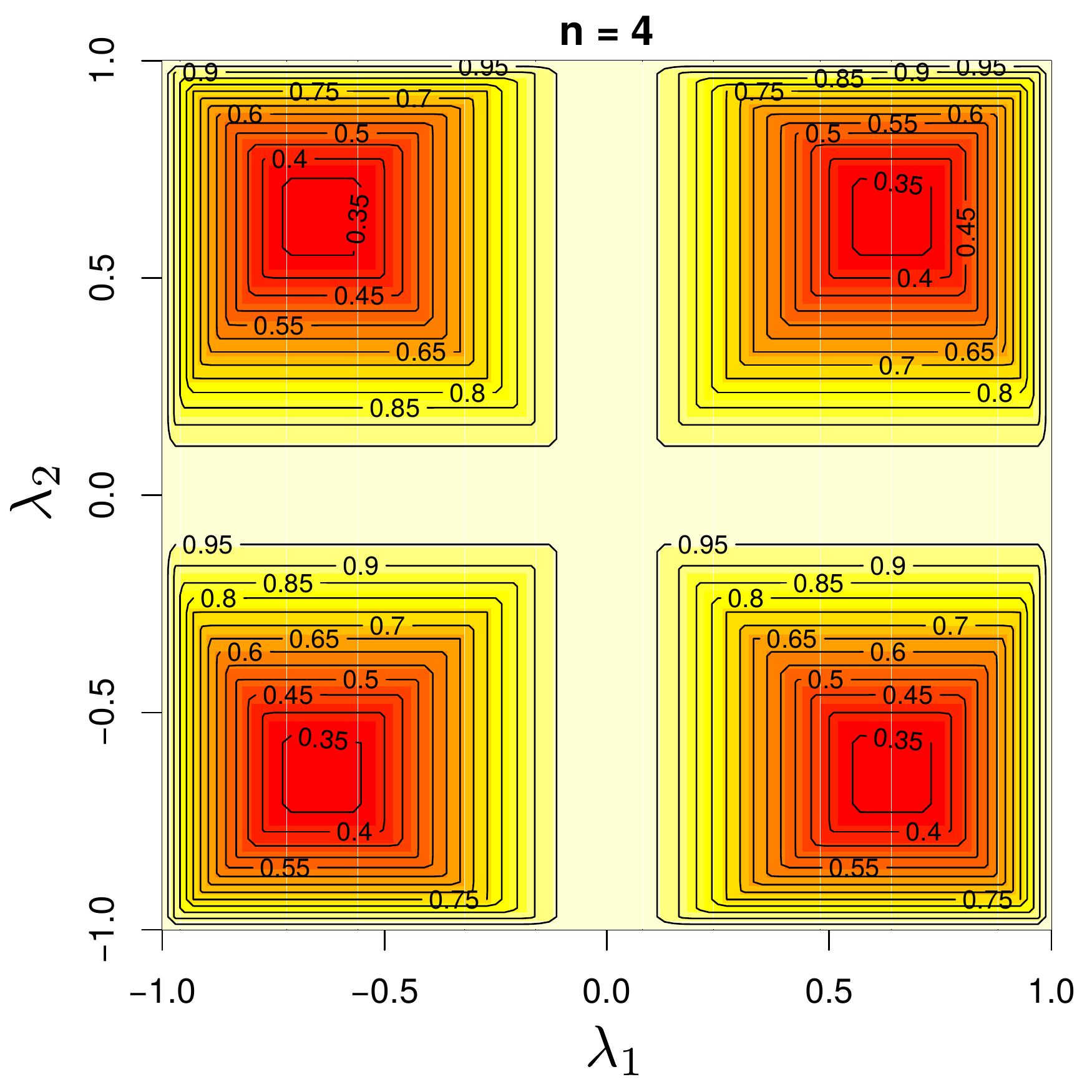} &
\includegraphics[width=0.31\textwidth]{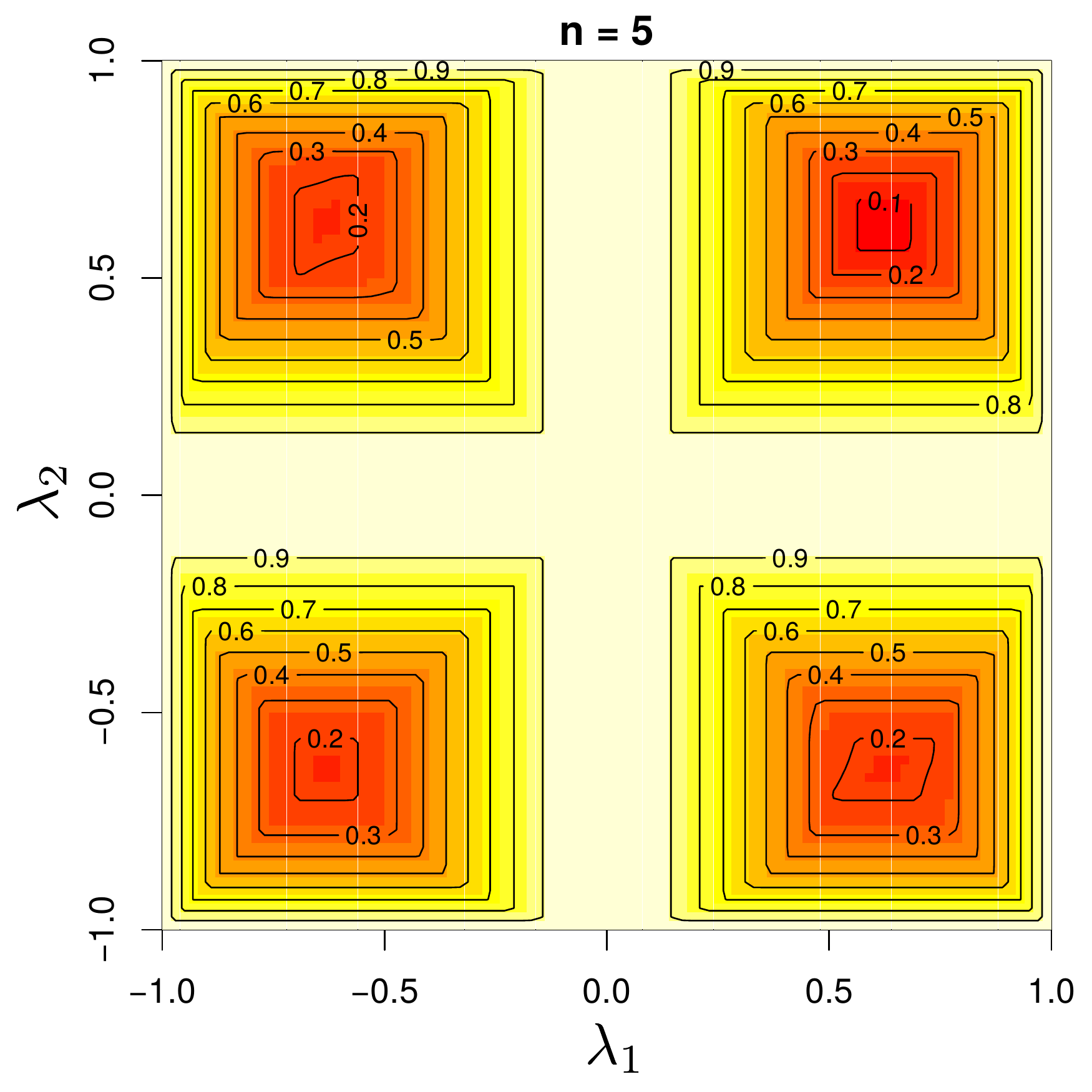} \\
\includegraphics[width=0.31\textwidth]{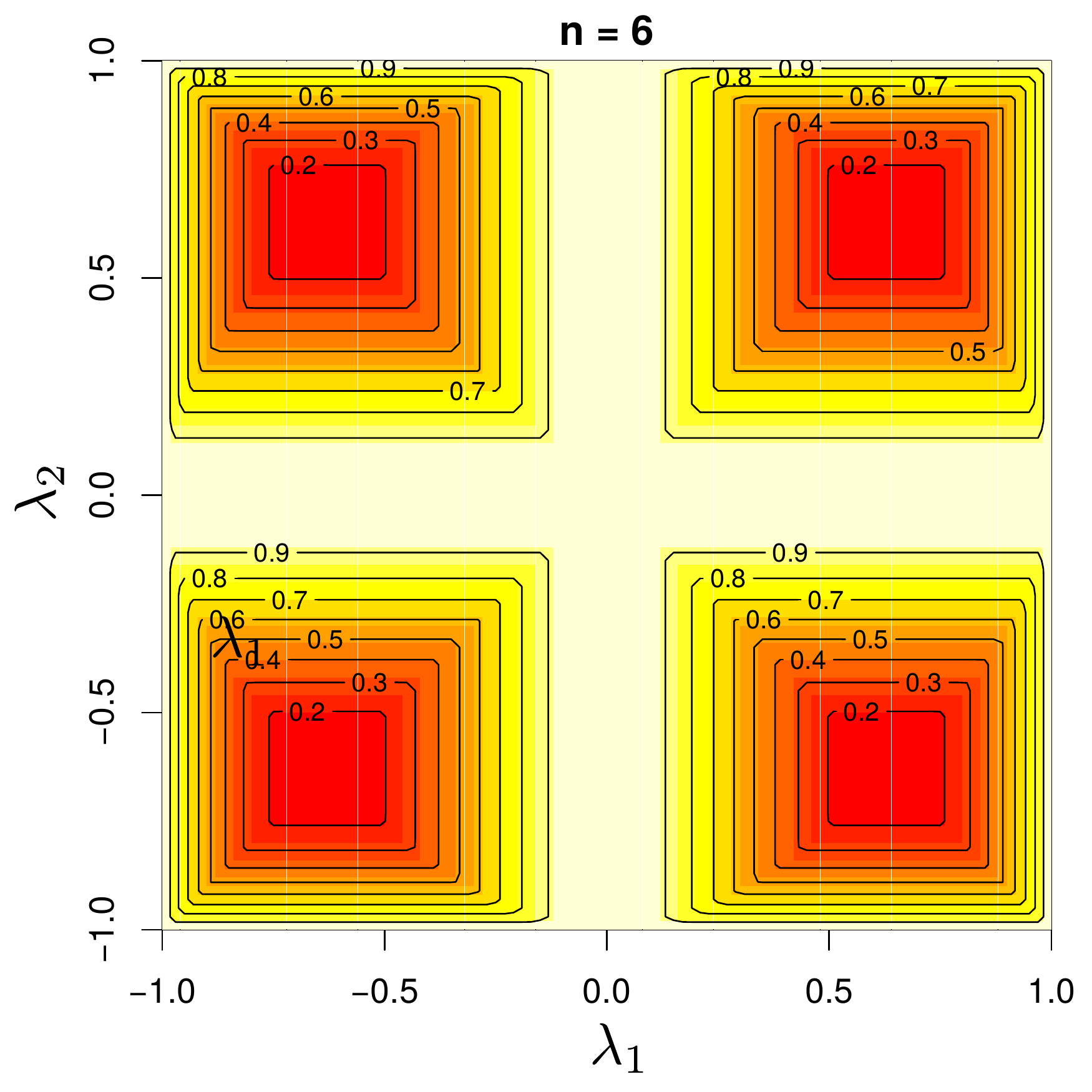} &
\includegraphics[width=0.31\textwidth]{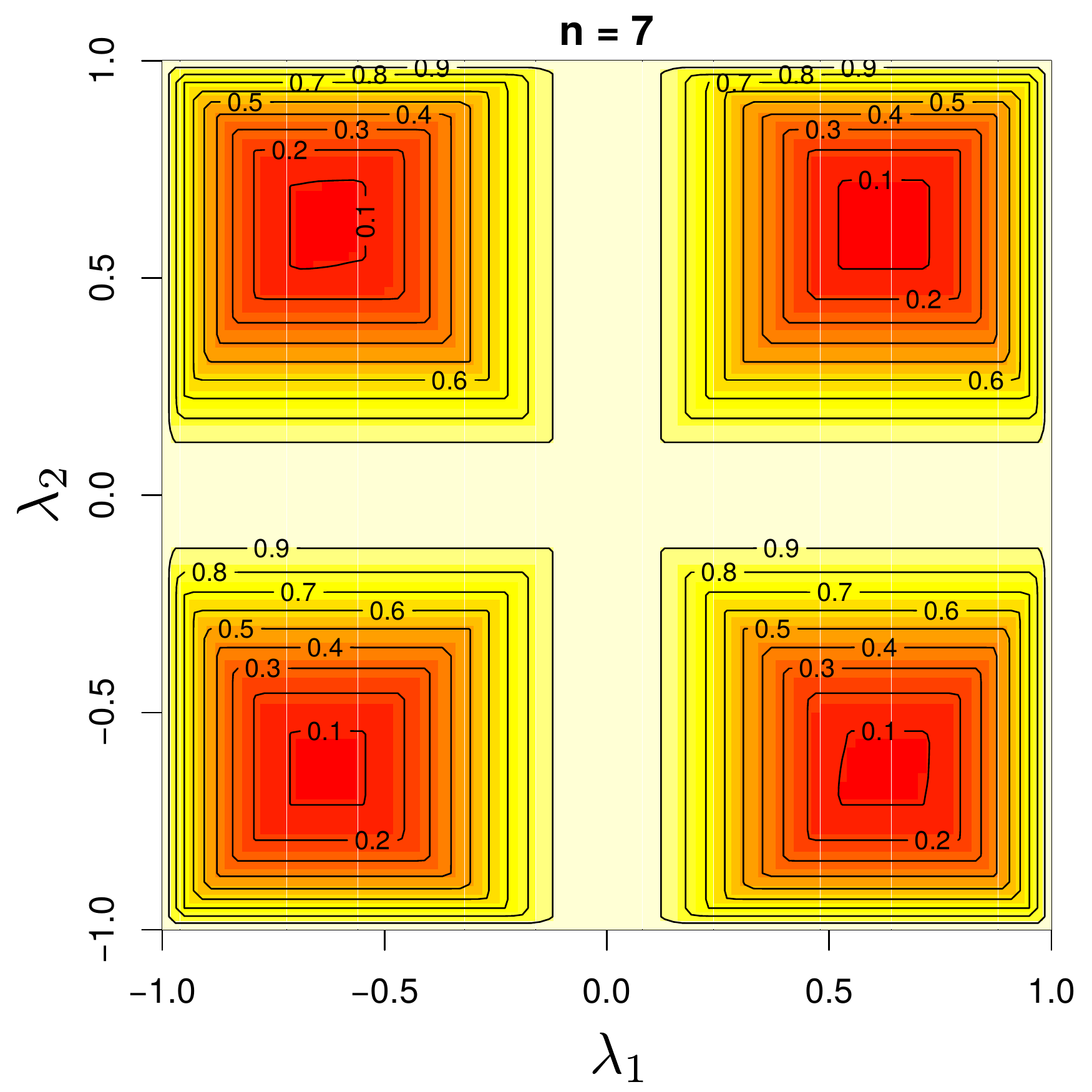} &
\includegraphics[width=0.31\textwidth]{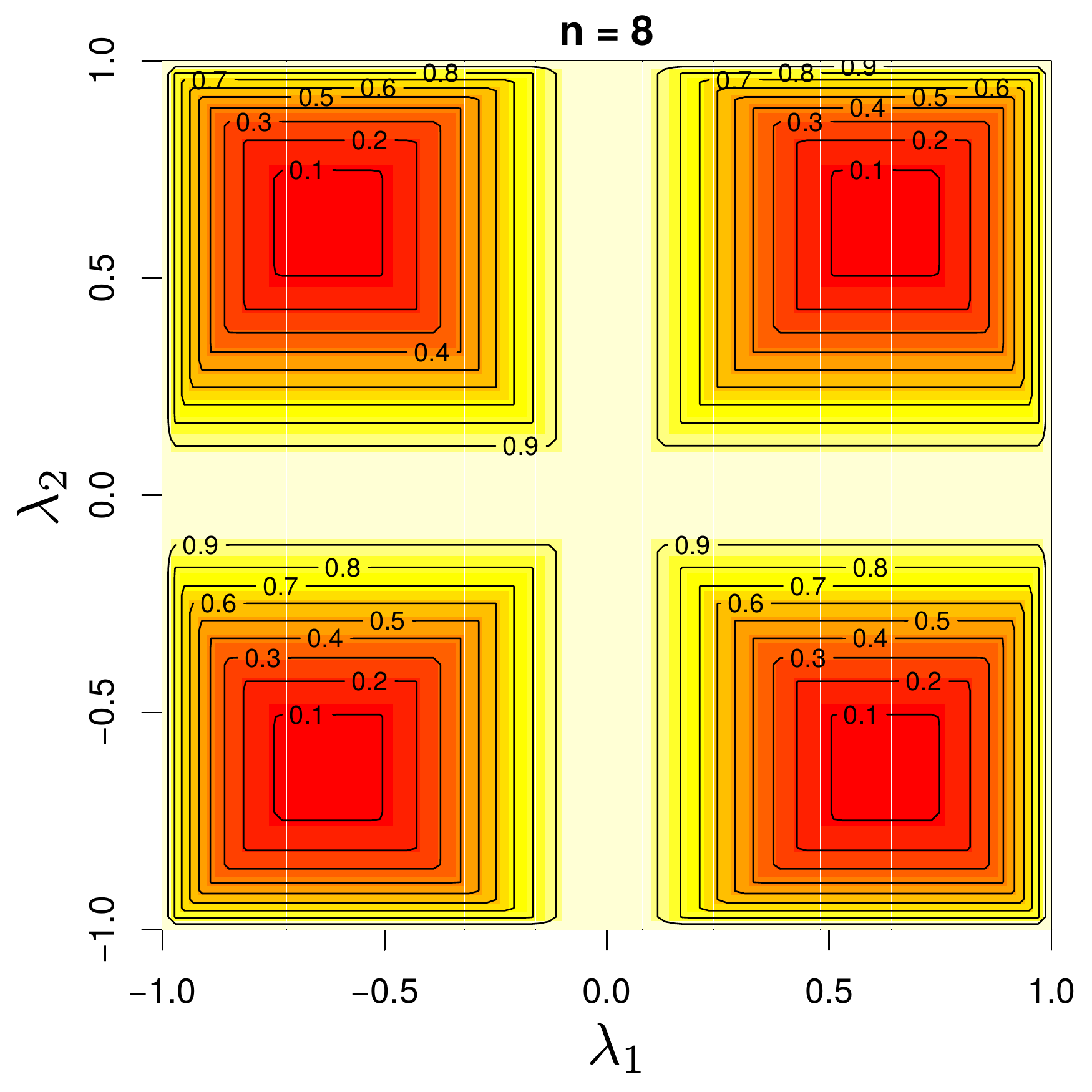} 
\end{tabular}
\end{center}
\caption{Contour curves of the values of $||\mathbf{M}_n||_\text{op}$ for $d=2$
and for $n > 2$ as a function of $\lambda_1$ and $\lambda_2$, \emph{i.e.}, the
two eigenvalues of $\mathbf{A}$. $\mathbf{A}$ is assumed to be symmetric.
Similar results are obtained for higher-order cumulants.}
\label{fig:operator_norm}
\end{figure}

Figure \ref{fig:operator_norm} shows, for symmetric $\mathbf{A}$, the value of
$||\mathbf{M}_n||_\text{op}$ for $n=3,\ldots,8$, and $d=2$ when the two
eigenvalues of $\mathbf{A}$ range in the interval $(-1,1)$. We observe that
$||\mathbf{M}_n||_\text{op}$ is always smaller than one. As described before, this will lead to a
reduction in the $\ell_2$-norm of the cumulants in the anti-causal
direction due to (\ref{eq:relation_multi}), and will in consequence produce a
Gaussianization effect on the distribution of the residuals. 
For $n \le 2$ it can be readily shown that $||\mathbf{M}_1||_\text{op} = ||\mathbf{M}_2||_\text{op} = 1$.

\begin{figure}[tb]
\begin{center}
\begin{tabular}{cccc}
\includegraphics[width=0.31\textwidth]{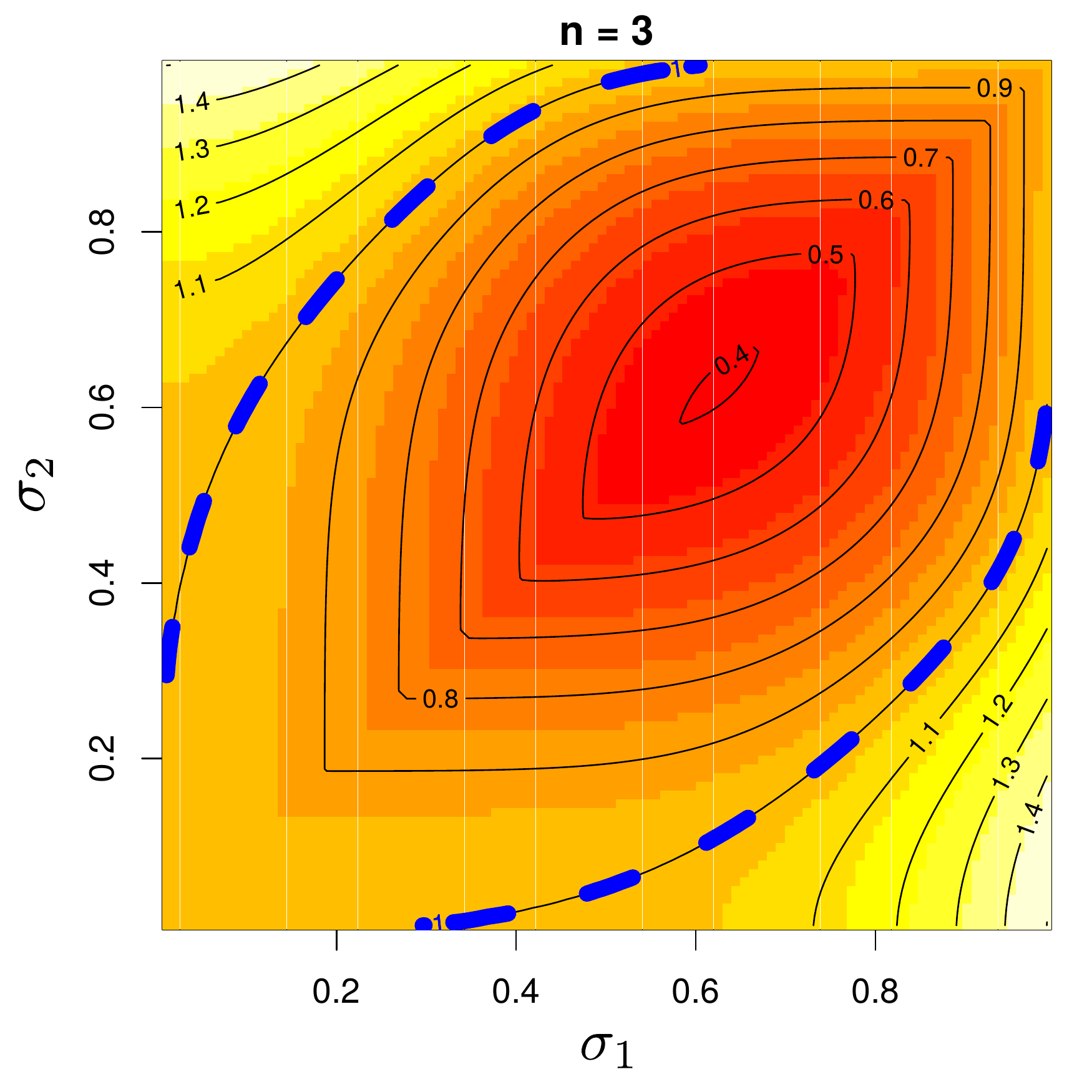} &
\includegraphics[width=0.31\textwidth]{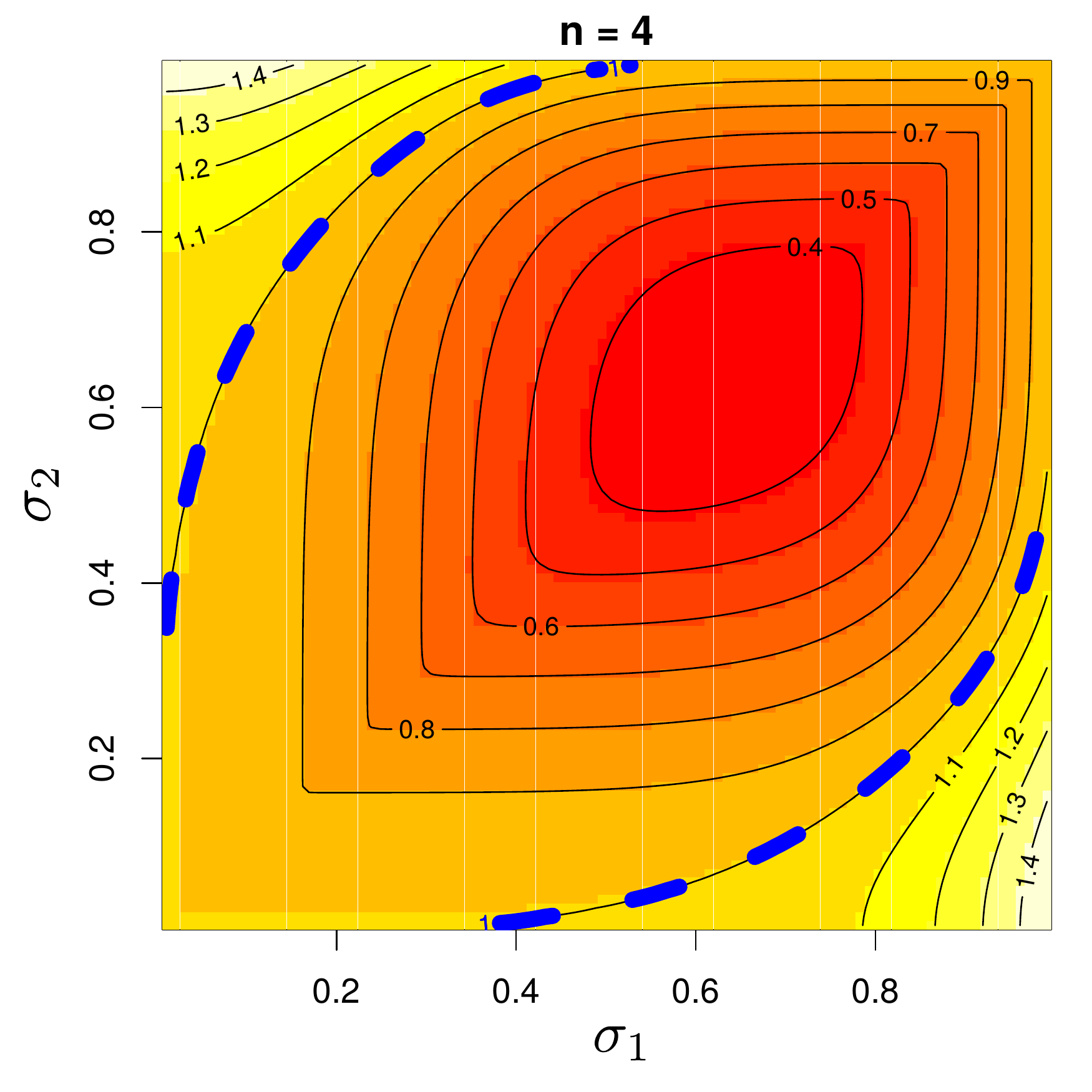} &
\includegraphics[width=0.31\textwidth]{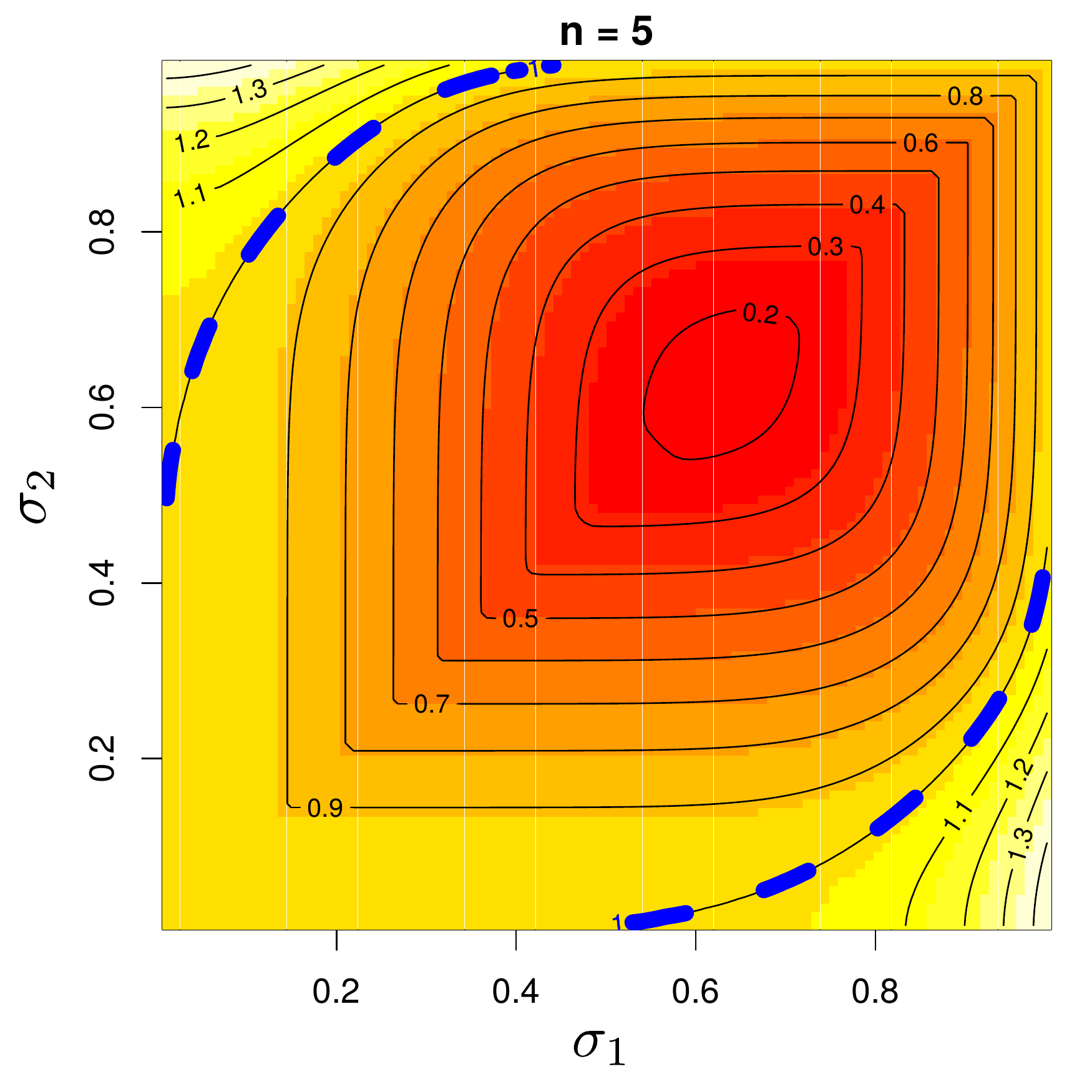} \\
\includegraphics[width=0.31\textwidth]{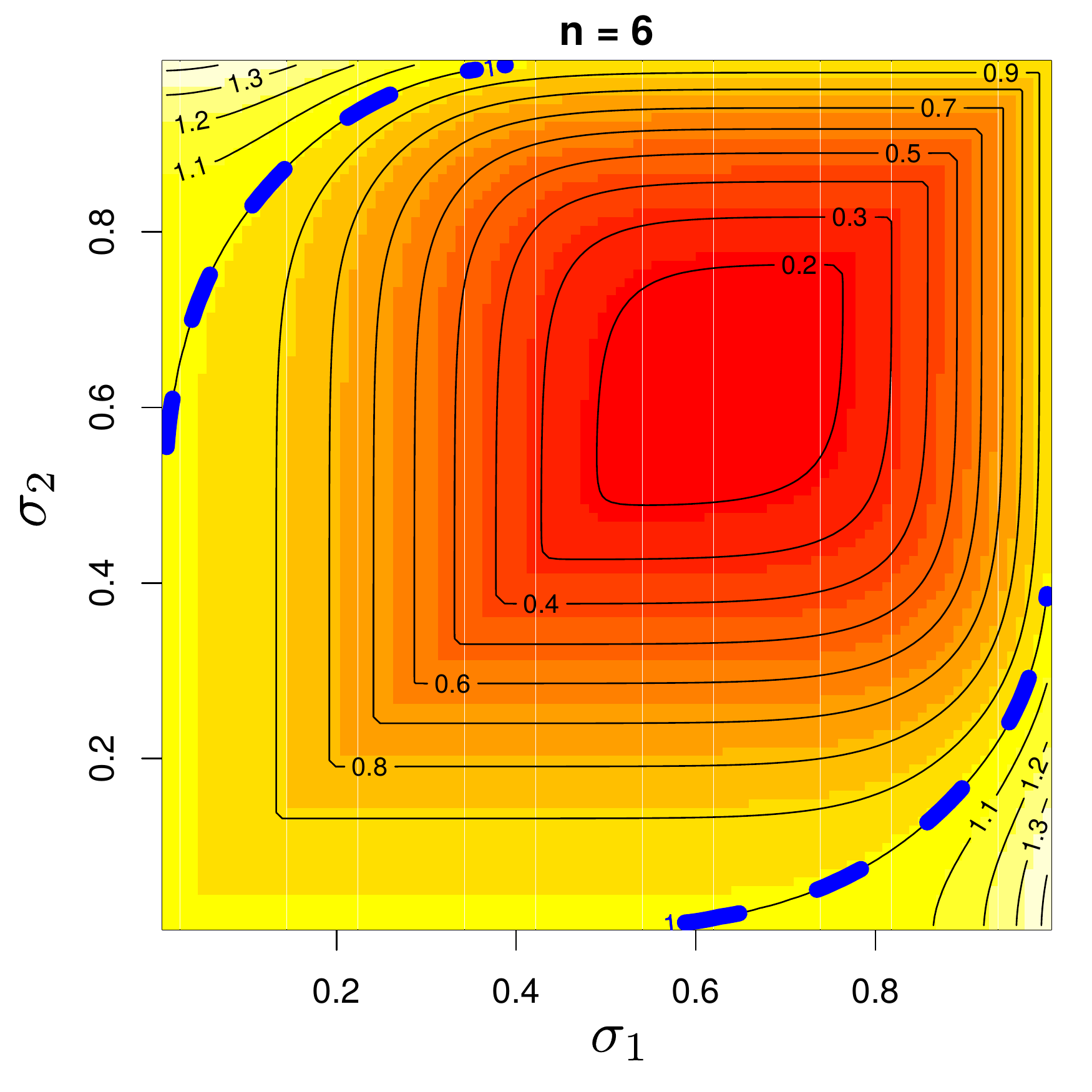} &
\includegraphics[width=0.31\textwidth]{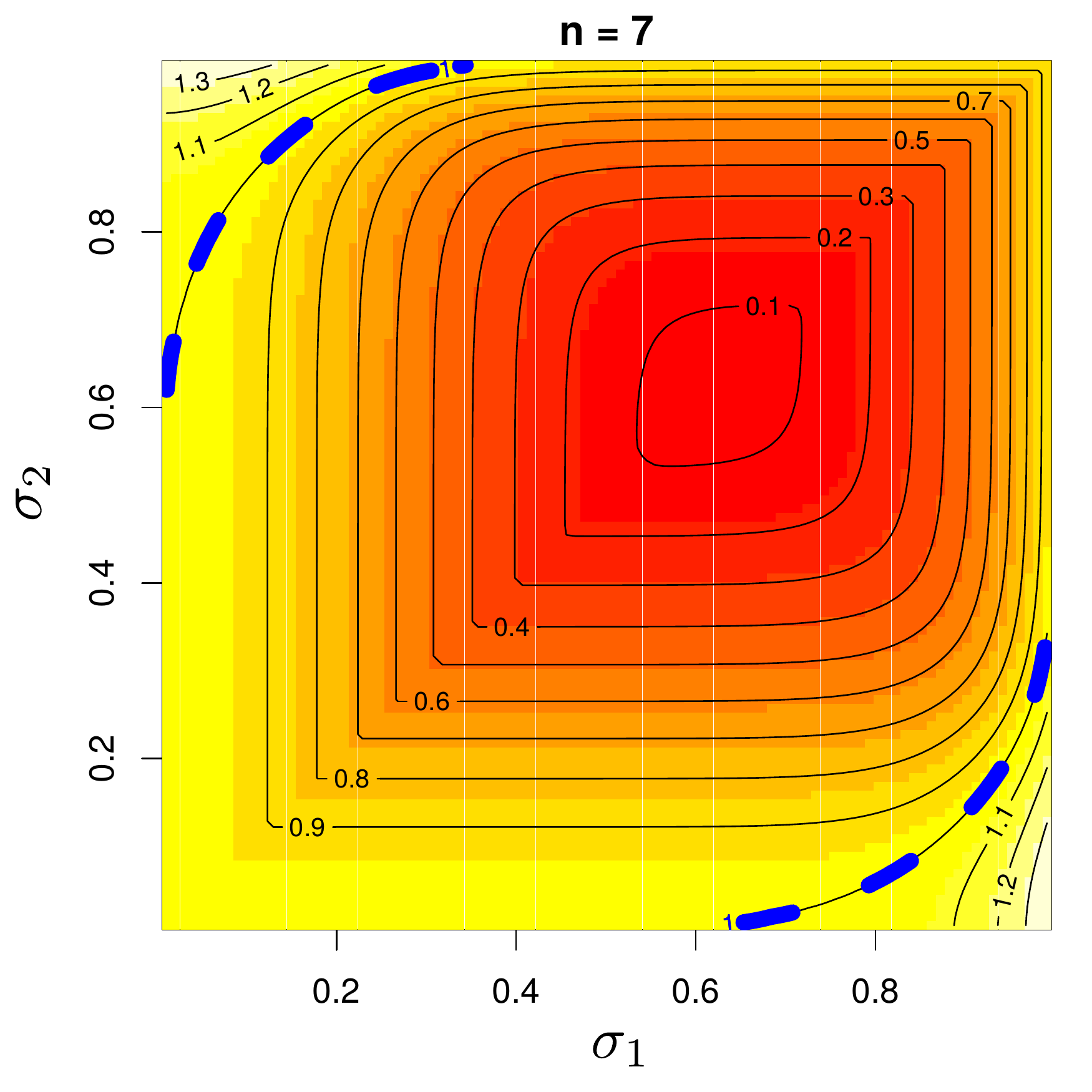} &
\includegraphics[width=0.31\textwidth]{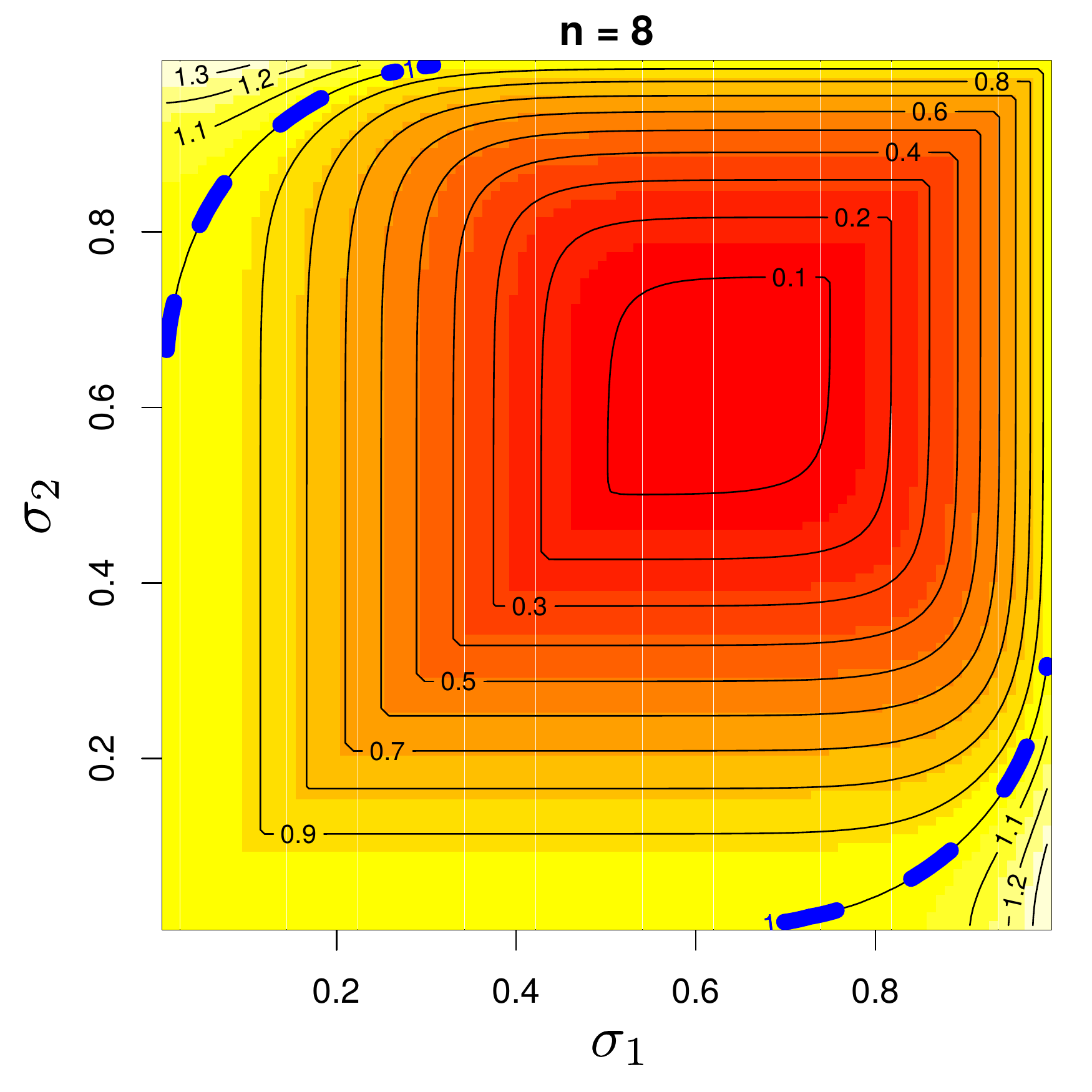} 
\end{tabular}
\end{center}
\caption{Contour curves of the values of $||\mathbf{M}_n||_\text{op}$ for $d=2$
and for $n > 2$ as a function of $\sigma_1$ and $\sigma_2$, \emph{i.e.}, the
two singular values of $\mathbf{A}$. $\mathbf{A}$ is not assumed to be
symmetric.  The singular vectors of $\mathbf{A}$ are chosen at random. 
A dashed blue line highlights the boundary of the region where 
$||\mathbf{M}_n||_\text{op}$ is strictly smaller than one. }
\label{fig:operator_norm_non_symmetric}
\end{figure}

In general, the matrix $\mathbf{A}$ need not be symmetric. In this case,
$||\mathbf{M}_1||_\text{op} = ||\mathbf{M}_2||_\text{op} = 1$ as well. However,
the evaluation of $||\mathbf{M}_n||_\text{op}$ for $n>2$
is more difficult, but feasible
for small $n$.  Figure \ref{fig:operator_norm_non_symmetric} displays the values of $||\mathbf{M}_n||_\text{op}$, for $d=2$,
as the two singular values of $\mathbf{A}$, $\sigma_1$ and $\sigma_2$, vary
in the interval $(0,1)$. The left singular vectors and the right singular
vectors of $\mathbf{A}$ are chosen at random.
In this figure a dashed blue line highlights the boundary of the region 
where $||\mathbf{M}_n||_\text{op}$ is strictly smaller than one. 
We observe that for most values of $\sigma_1$ and $\sigma_2$,
$||\mathbf{M}_n||_\text{op}$ is smaller than one, leading to a
Gaussianization effect in the distribution of the residuals in the anti-causal direction.
However, for some singular values, $||\mathbf{M}_n||_\text{op}$ is strictly larger than
one. Of course, this does not mean that there is not such a Gaussianization effect also in those cases.  The definition given for $||\mathbf{M}_n||_\text{op}$ assumes that all
potential vectors $\mathbf{v}$ represent valid cumulants of a probability distribution, which need not be the case 
in practice. For example, it is well known that cumulants exhibit some form of symmetry \citep{mccullagh87}.
This can be seen in the second order cumulant, which is a covariance matrix.
The consequence is that $||\mathbf{M}_n||_\text{op}$ is simply an upper bound on the reduction of the
$\ell_2$-norm of the cumulants in the anti-causal model. Thus, we also
expect a Gaussianization effect to occur also for these cases. 
Furthermore, the numerical simulations presented in 
Section \ref{sec:experiments} provide evidence of this effect for asymmetric $\mathbf{A}$.

\begin{figure}[tb]
\begin{center}
\begin{tabular}{cc}
\includegraphics[height=4.5cm]{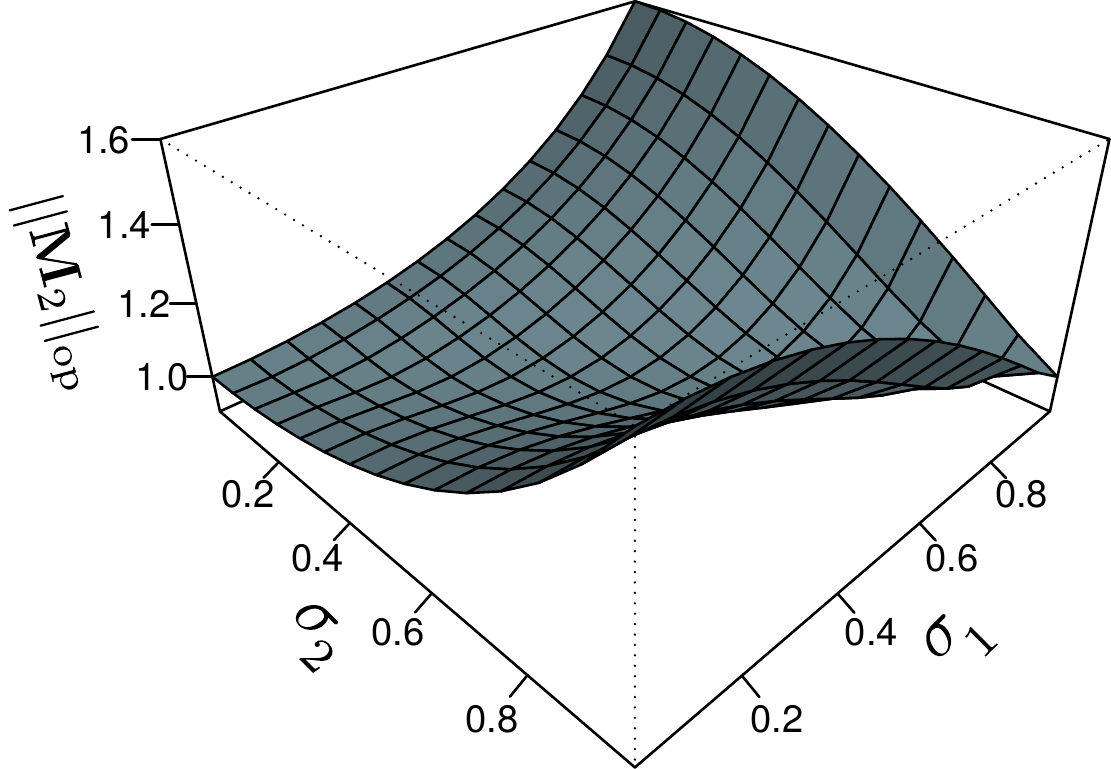} &
\includegraphics[height=4.5cm]{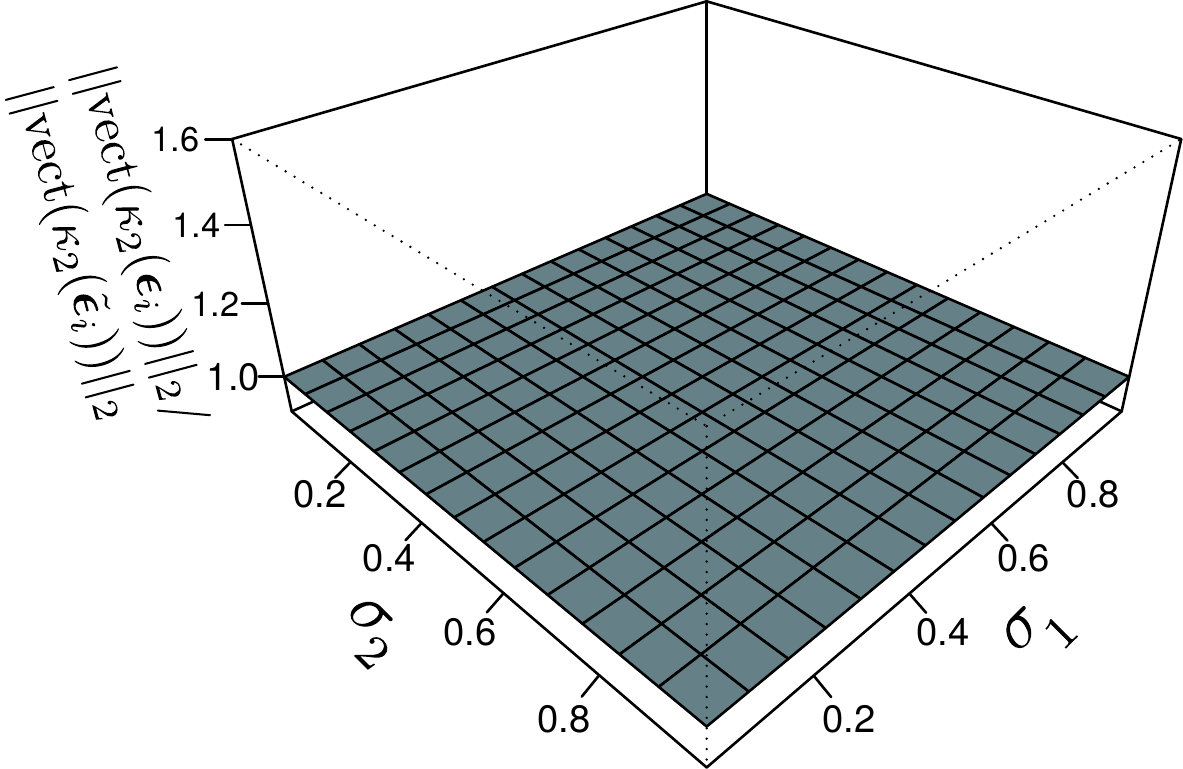} 
\end{tabular}
\end{center}
\caption{(left) Value of $||\mathbf{M}_2||_\text{op}$,
for $d=2$, as a function of $\sigma_1$ and $\sigma_2$, \emph{i.e.}, the
two singular values of $\mathbf{A}$. 
(right) Actual ratio between $||\text{vect}(\kappa_2(\bm{\epsilon}_i))||_2$ and
$||\text{vect}(\kappa_2(\tilde{\bm{\epsilon}}_i))||_2$,
for $d=2$, as a function of $\sigma_1$ and $\sigma_2$.
$\mathbf{A}$ is not assumed to be symmetric. The singular vectors of $\mathbf{A}$ are chosen at random. }
\label{fig:operator_norm_bound}
\end{figure}

The fact that $||\mathbf{M}_n||_\text{op}$ is only an upper bound is illustrated in 
Figure \ref{fig:operator_norm_bound}. This figure considers the particular case of 
the second cumulant $\kappa_2(\cdot)$, which can be analyzed in detail. 
On the left plot the value of $||\mathbf{M}_2||_\text{op}$ is displayed as 
a function of $\sigma_1$ and $\sigma_2$, the two singular values of $\mathbf{A}$.
We observe that $||\mathbf{M}_2||_\text{op}$ takes values that are larger than one.
In this case it is possible to evaluate 
in closed form the $\ell_2$-norm of $\text{vect}(\kappa_2(\bm{\epsilon}_i))$ and
$\text{vect}(\kappa_2(\tilde{\bm{\epsilon}}_i))$, \emph{i.e.}, the vectors that
contain the second order cumulant of the residuals in each direction.
In particular, it is well known that the second order cumulant is equal to the covariance matrix 
\citep{mccullagh87}. In the causal direction, the covariance matrix of 
the residuals is $\mathbf{C}=\mathbf{I}-\mathbf{A}\mathbf{A}^\text{T}$, as shown in the previous 
paragraphs. The covariance matrix of the residuals in the anti-causal direction, 
denoted by $\tilde{\mathbf{C}}$, can be computed similarly. Namely, $\tilde{\mathbf{C}} = \mathbf{I}-\mathbf{A}^\text{T}\mathbf{A}$. 
These two matrices, \emph{i.e.}, $\mathbf{C}$ and $\tilde{\mathbf{C}}$, 
respectively give $k_2(\bm{\epsilon}_i)$ and $k_2(\tilde{\bm{\epsilon}}_i)$.
Furthermore, they have the same singular values. This means that
$||\text{vect}(k_2(\bm{\epsilon}_i))||_2 /  ||\text{vect}(k_2(\tilde{\bm{\epsilon}}_i))||_2 = 1$, 
as illustrated by the right plot in Figure \ref{fig:operator_norm_bound}. 
Thus, $||\mathbf{M}_2||_\text{op}$ is simply an upper bound on the actual reduction 
of the $\ell_2$-norm of the second order cumulant of the residuals in the anti-causal 
direction. The same behavior is expected for $||\mathbf{M}_n||_\text{op}$, with $n>2$.
In consequence, one should expect that the cumulants of the distribution of the
residuals of a model fitted in the anti-causal direction are smaller in magnitude.
This will lead to an increased level of Gaussianity measured in terms of a reduction of 
the magnitude of the high-order cumulants.

\subsection{Analysis of the multivariate case based on information theory}
\label{sec:information_theory}

The analysis of the multivariate case carried out in the previous section is illustrative. Nevertheless,
it does not prove that the distribution of the residuals in the anti-casual direction is more Gaussian 
based on a reduction of the magnitude of the cumulants. Further evidence of this increased level of 
Gaussianity can be obtained based on an increase of the entropy obtained by using information theory. 
In this case we also assume that the multi-dimensional variables $\mathcal{X}$ and $\mathcal{Y}$ follow 
the same distribution, but unlike in the previous section, they need not be whitened, only centered.
In this section we closely follow \citep[Sec. 2.4]{HyvarinenS13} and extend their results to the multivariate case.

Under the assumptions specified earlier,
the model in the causal direction is $\mathbf{y}_i = \mathbf{A} \mathbf{x}_i + \bm{\epsilon}_i$, with
$\mathbf{x}_i \bot \bm{\epsilon}_i$ and $\mathbf{A}=\text{Cov}(\mathcal{Y},\mathcal{X})\text{Cov}(\mathcal{X},\mathcal{X})^{-1}$.
Similarly, the model in the anti-causal direction is $\mathbf{x}_i = \tilde{\mathbf{A}} \mathbf{y}_i + \tilde{\bm{\epsilon}}_i$
with $\tilde{\mathbf{A}}=\text{Cov}(\mathcal{X},\mathcal{Y})\text{Cov}(\mathcal{Y},\mathcal{Y})^{-1}$.
By making use of the causal model it is possible to show that 
$\tilde{\bm{\epsilon}}_i = 
(\mathbf{I}-\tilde{\mathbf{A}} \mathbf{A})\mathbf{x}_i - \tilde{\mathbf{A}} \bm{\epsilon}_i$, where 
$\mathbf{I}$ is the identity matrix. Thus, the following equations are satisfied:
\begin{align}
\left(\begin{array}{c} \mathbf{x}_i \\
\mathbf{y}_i \end{array}\right) & = \mathbf{P} \left(\begin{array}{c} \mathbf{x}_i \\
\bm{\epsilon}_i \end{array}\right) = \left(\begin{array}{cc} \mathbf{I} & \mathbf{0} \\
\mathbf{A} & \mathbf{I} \end{array}\right) \left(\begin{array}{c} \mathbf{x}_i \\
\bm{\epsilon}_i \end{array}\right)\,,
\label{eq:relation_entropy_1}
\end{align}
and
\begin{align}
\left(\begin{array}{c}
\mathbf{y}_i \\
\tilde{\bm{\epsilon}}_i
\end{array}\right) & = \tilde{\mathbf{P}}
\left(\begin{array}{c}
\mathbf{x}_i \\
\bm{\epsilon}_i
\end{array}\right) = 
\left(\begin{array}{cc}
\mathbf{A} & \mathbf{I} \\
\mathbf{I}-\tilde{\mathbf{A}}\mathbf{A} & -\tilde{\mathbf{A}}
\end{array}\right)
\left(\begin{array}{c}
\mathbf{x}_i \\
\bm{\epsilon}_i
\end{array}\right)\,.
\label{eq:relation_entropy_2}
\end{align}

Let $H(\mathbf{x}_i,\mathbf{y}_i)$ be the entropy of the joint distribution of the two 
random variables associated with samples $\mathbf{x}_i$ and $\mathbf{y}_i$. Because (\ref{eq:relation_entropy_1}) 
is a linear transformation, we can use the entropy transformation formula \citep{HyvarinenS13} to get that
$H(\mathbf{x}_i,\mathbf{y}_i)$ = $H(\mathbf{x}_i,\bm{\epsilon}_i) + \log|\text{det}\mathbf{P}|$,
where $\text{det}\mathbf{P}=\text{det}\mathbf{I}=1$. Thus, we have that
$H(\mathbf{x}_i,\mathbf{y}_i)$ = $H(\mathbf{x}_i,\bm{\epsilon}_i)$.
Conversely, if we use (\ref{eq:relation_entropy_2}) we have
$H(\mathbf{y}_i,\bm{\epsilon}_i)$ = $H(\mathbf{x}_i,\bm{\epsilon}_i) + \log|\text{det}\tilde{\mathbf{P}}|$,
where $\text{det}\tilde{\mathbf{P}}=\text{det}\mathbf{A} 
\cdot \text{det}(-\tilde{\mathbf{A}} - (\mathbf{I}-\tilde{\mathbf{A}} \mathbf{A}) \mathbf{A}^{-1} \mathbf{I}) = -\text{det}\mathbf{I}=-1$,
under the assumption that $\mathbf{A}$ is invertible.
The result is that $H(\mathbf{x}_i,\mathbf{y}_i)=H(\mathbf{y}_i,\bm{\epsilon}_i)$ = $H(\mathbf{x}_i,\bm{\epsilon}_i)$.

Denote the mutual information between the cause and the noise with $I(\mathbf{x}_i,\bm{\epsilon}_i)$. 
Similarly, let $I(\mathbf{y}_i,\tilde{\bm{\epsilon}}_i)$ be the mutual information between the 
random variables corresponding to the observations $\mathbf{y}_i$ and $\tilde{\bm{\epsilon}}_i$. Then,
\begin{align}
I(\mathbf{x}_i,\bm{\epsilon}_i) - I(\mathbf{y}_i,\tilde{\bm{\epsilon}}_i) & = 
H(\mathbf{x}_i) + H(\bm{\epsilon}_i) - H(\mathbf{x}_i,\bm{\epsilon}_i) - 
H(\mathbf{y}_i) - H(\tilde{\bm{\epsilon}}_i) + H(\mathbf{y}_i, \tilde{\bm{\epsilon}}_i)
\nonumber \\
&=  H(\mathbf{x}_i) + H(\bm{\epsilon}_i) - H(\mathbf{y}_i) - H(\tilde{\bm{\epsilon}}_i)\,.
\end{align}
Furthermore, from the actual causal model assumed we know that $I(\mathbf{x}_i,\bm{\epsilon}_i) = 0$. By contrast,
we have that $I(\mathbf{y}_i,\tilde{\bm{\epsilon}}_i) \geq 0$, since both 
$\mathbf{y}_i$ and $\tilde{\bm{\epsilon}}_i$ depend on $\mathbf{x}_i$ and $\bm{\epsilon}_i$. 
We also know that $H(\mathbf{x}_i)=H(\mathbf{y}_i)$ because we have made the hypothesis that both $\mathcal{X}$ and
$\mathcal{Y}$ follow the same distribution.  The result is that:
\begin{align}
H(\bm{\epsilon}_i)  \leq H(\tilde{\bm{\epsilon}}_i)\,,
\end{align}
with equality iff the residuals are Gaussian. 
We note that an alternative but equivalent way to obtain this last result is to consider 
a multivariate version of Lemma 1 in \citep{KpotufeSJS14}, under the assumption of the same 
distribution for the cause and the effect. In particular, even though \cite{KpotufeSJS14} assume 
univariate random variables, their work can be easily generalized to multiple variables. 

Although the random variables corresponding to $\bm{\epsilon}_i$ and $\tilde{\bm{\epsilon}}_i$ have both zero mean, they need 
not have the same covariance matrix. Denote with $\text{Cov}(\bm{\epsilon}_i)$ and $\text{Cov}(\tilde{\bm{\epsilon}}_i)$ to these 
matrices and let $\hat{\bm{\epsilon}}_i$ and $\hat{\tilde{\bm{\epsilon}}}_i$ be the whitened residuals 
(\emph{i.e.}, the residuals multiplied by the Cholesky factor of the inverse of the corresponding covariance matrix). Then,
\begin{align}
H(\hat{\bm{\epsilon}}_i) & \leq H(\hat{\tilde{\bm{\epsilon}}}_i) - 
	\frac{1}{2} \log |\text{det}\text{Cov}(\tilde{\bm{\epsilon}}_i)| + \frac{1}{2}\log |\text{det}\text{Cov}(\bm{\epsilon}_i)|\,.
\end{align}
As shown in Appendix \ref{Appendix:A}, although not equal, 
the matrices $\text{Cov}(\bm{\epsilon}_i)$ and $\text{Cov}(\tilde{\bm{\epsilon}}_i)$ 
have the same determinant.  Thus, the two determinants cancel in the equation above. This gives,
\begin{align}
H(\hat{\bm{\epsilon}}_i) \leq H(\hat{\tilde{\bm{\epsilon}}}_i)\,.
\end{align}

The consequence is that the entropy of the whitened residuals in the anti-causal direction is expected to be higher or equal than 
the entropy of the whitened residuals in the causal direction. Because the Gaussian distribution is the continuous distribution with 
the highest entropy for a fixed covariance matrix, we conclude that the level of Gaussianity of the residuals in the 
anti-causal direction, measured in terms of differential entropy, has to be larger or equal to the level of Gaussianity 
of the residuals in the causal direction.

In summary, if the causal relation between the two identically distributed
random variables $\mathcal{X}$ and $\mathcal{Y}$ is linear the residuals of
a least squares fit  in the anti-causal direction are
more Gaussian than those of a linear fit in the causal direction. 
This Gaussianization can be characterized by a reduction of the magnitude 
of the corresponding high-order cumulants (although we have not formally proved this, 
we have provided some evidence that this is the case) and by an increase 
of the entropy (we have proved this in this section). 
A causal inference method can take advantage of this 
asymmetry to determine the causal direction. In particular, statistical tests 
based on measures of Gaussianity can be used for this purpose.
However and importantly, these measures of Gaussianity need not be 
estimates of the differential  entropy or the high-order cumulants. In particular, the entropy is a quantity 
that is particularly difficult to estimate in practice \citep{beirlant++_1997_nonparametric}.
The same occurs with the high-order cumulants. Their estimators involve high-order 
moments, and hence suffer from high variance.

When the distribution of the residuals in (\ref{eq:model_multi}) is Gaussian, the 
causal direction cannot be identified. In this case, it is possible to show that
the distribution of the reversed residuals $\tilde{\bm{\epsilon}}_i$, the cause $\mathbf{x}_i$, and 
the effect $\mathbf{y}_i$, is Gaussian as a consequence of (\ref{eq:cum_mul_rel}) and (\ref{eq:relation_multi}). 
This non-identifiability agrees with the general result of \cite{Shimizu06}, which indicates 
that non-Gaussian distributions are strictly required in the disturbance variables to carry 
out causal inference in linear models with additive independent noise.

Finally, the fact that the Gaussianization effect is also expected in the
multivariate case suggests a method to address causal inference
problems in which the relationship between cause and
effect is non-linear: It consists in mapping each observation $x_i$ and $y_i$ 
to a vector in an expanded feature space.
One can them assume that the non-linear relation in the original input 
space is linear in the expanded feature space and compute
the residuals of  kernel ridge regressions in both directions. 
The direction in which the residuals are less Gaussian is identified
as the causal one.

\section{A feature expansion to address non-linear causal inference problems}

\label{sec:feature_expansion}

We now proceed to relax the assumption that the causal relationship between the 
unidimensional random variables $\mathcal{X}$ and $\mathcal{Y}$ is linear. 
For this purpose, instead of working in the original space 
in which the samples $ \left\{ (x_i, y_i)\right\}_{i=1}^N$ are observed, 
we will  assume that the model is linear in some expanded feature space
$ \left\{ \left(\phi(x_i), \phi(y_i) \right)\right\}_{i=1}^N$  for some mapping function $\phi(\cdot):\mathds{R} \rightarrow \mathds{R}^d$. 
Importantly, this map preserves the property that if $x_i$ and $y_i$ are equally distributed, so 
will be $\phi(x_i)$ and $\phi(y_i)$.  According to the analysis presented in the previous section, the 
residuals of a linear model in the expanded space should be more Gaussian in the anti-causal direction 
than the residuals of a linear model in the causal direction, based on an increment of the differential
entropy and on a reduction of the magnitude of the cumulants. The assumption we make is that the
non-linear relation between $\mathcal{X}$ and $\mathcal{Y}$ in the original input space is linear 
in the expanded feature space.

In this section we focus on obtaining the normalized residuals of
linear models formulated in the expanded
feature space. For this purpose, we assume that a kernel function $k(\cdot,\cdot)$
can be used to evaluate dot products in the expanded feature space. In particular,
$k(x_i,x_j)  = \phi(x_i)^\text{T} \phi(x_j)$ and $k(y_i,y_j) =
\phi(y_i)^\text{T} \phi(y_j)$ for arbitrary $x_i$ and $x_j$ and $y_i$ and
$y_j$. Furthermore, we will not assume in general that $\phi(x_i)$ and
$\phi(y_i)$ have been whitened, only centered. Whitening is a linear
transformation which is not expected to affect to the level of Gaussianity of
the residuals. However, once these residuals have been obtained they will
be whitened in the expanded feature space. Later on we describe how to center
the data in the expanded feature space. For now on, we will assume this step
has already been done.

\subsection{Non-linear model description and fitting process}

Assume that the relation between
$\mathcal{X}$ and $\mathcal{Y}$ is linear in an expanded feature space
\begin{align}
\phi(y_i) = \mathbf{A} \phi(x_i) + \bm{\epsilon}_i\,, \quad \bm{\epsilon}_i \perp x_i,
\end{align}
where  $\bm{\epsilon}_i$ is i.i.d. non-Gaussian additive noise.

Given $N$ paired observations $\{(x_i,y_i)\}_{i=1}^N$ drawn i.i.d. from
$P(\mathcal{X}, \mathcal{Y})$, define the matrices
$\bm{\Phi}_x=(\phi(x_1),\ldots,\phi(x_N))$ and
$\bm{\Phi}_y=(\phi(y_1),\ldots,\phi(y_N))$ of size $d \times N$.  The estimate of $\mathbf{A}$ that minimizes the
sum of squared errors is $\hat{\mathbf{A}} = \bm{\Gamma}\bm{\Sigma}^{-1}$,
where $\bm{\Gamma}  = \bm{\Phi}_y\bm{\Phi}_x^\text{T}$ and $\bm{\Sigma} =
\bm{\Phi}_x\bm{\Phi}_x^\text{T}$.  Unfortunately, when $d > N$, where $d$ is
the number of variables in the feature expansion, the matrix $\bm{\Sigma}^{-1}$
does not exist and $\hat{\mathbf{A}}$ is not unique.  This means that there is
an infinite number of solutions for $\hat{\mathbf{A}}$  with zero squared
error.

To avoid the indetermination described above and also to alleviate
over-fitting, we propose a regularized estimator. Namely,
\begin{align}
\mathcal{L}(\mathbf{A}) &= \sum_{i=1}^N \frac{1}{2}||\phi(y_i) -
\mathbf{A}\phi(x_i)||_2^2 + \tau \frac{1}{2} ||\mathbf{A}||_\mathcal{F}^2\,,
\label{eq:optim}
\end{align}
where $||\cdot||_2$ denotes the $\ell_2$-norm and $||\cdot||_\mathcal{F}$
denotes the Frobenius norm.  In this last expression $\tau>0$ is a parameter
that controls the amount of regularization.  The minimizer of (\ref{eq:optim})
is $\hat{\mathbf{A}} = \bm{\Gamma} \bm{\Sigma}^{-1}$, where $\bm{\Gamma} =
\bm{\Phi}_y\bm{\Phi}_x^\text{T}$ and $\bm{\Sigma} = \tau \mathbf{I} +
\bm{\Phi}_x\bm{\Phi}_x^\text{T}$.  The larger the value of $\tau$, the closer
the entries of $\hat{\mathbf{A}}$ are to zero.  Furthermore, using the matrix
inversion lemma we have that $\bm{\Sigma}^{-1} = \tau^{-1} \mathbf{I} -
\tau^{-1} \bm{\Phi}_x \mathbf{V}^{-1} \bm{\Phi}_x^\text{T}$, where
$\mathbf{V}=\left(\tau\mathbf{I}+\mathbf{K}_{x,x}\right)^{-1}$ and
$\mathbf{K}_{x,x}=\bm{\Phi}_x^\text{T}\bm{\Phi}_x$ is a kernel matrix whose
entries are given by $k(x_i,x_j)$. After some algebra it is possible to show
that
\begin{align}
\hat{\mathbf{A}} & = \bm{\Gamma}\bm{\Sigma}^{-1} = \bm{\Phi}_y\mathbf{V}
\bm{\Phi}_x^\text{T}\,,
\label{eq:matrix_A}
\end{align}
which depends only on the matrix $\mathbf{V}$. This matrix be computed with cost
$\mathcal{O}(N^3)$. We note that the estimate obtained in (\ref{eq:matrix_A}) coincides with 
the kernel conditional embedding operator described by \cite{song2013} for mapping 
conditional distributions into infinite dimensional feature spaces using kernels.

\subsection{Obtaining the matrix of inner products of the residuals}

A first step towards obtaining the whitened residuals in feature space
(which will be required for the estimation of their level of Gaussianity) is to
compute the matrix of inner products of these residuals (kernel matrix).  For
this, we define $\bm{\epsilon}_i = \phi(y_i) - \hat{\mathbf{A}} \phi(x_i)$. Thus,
\begin{align}
\bm{\epsilon}_i^\text{T} \bm{\epsilon}_j &= \left[ \phi(y_i) - \hat{\mathbf{A}}
\phi(x_i)\right]^\text{T} \left[ \phi(y_j) - \hat{\mathbf{A}} \phi(x_j)\right]
\nonumber \\
& = \phi(y_i)^\text{T}  \phi(y_j) - \phi(y_i)^\text{T} \hat{\mathbf{A}}
\phi(x_j)
  - \phi(x_i)^\text{T} \hat{\mathbf{A}} \phi(y_j) + \phi(x_i)^\text{T}
    \hat{\mathbf{A}} \hat{\mathbf{A}} \phi(x_j)\,,
\end{align}
for two arbitrary residuals $\bm{\epsilon}_i$ and $\bm{\epsilon}_j$ in feature
  space.  In general, if we denote with $\mathbf{K}_\epsilon$ to the matrix
  whose entries are given by $\bm{\epsilon}_i^\text{T} \bm{\epsilon}_j$ and
  define $\mathbf{K}_{y,y}=\bm{\Phi}_y^\text{T} \bm{\Phi}_y$, we have that
\begin{align}
\mathbf{K}_\epsilon &= \mathbf{K}_{y,y} - 
	\mathbf{K}_{y,y}\mathbf{V} \mathbf{K}_{x,x} - 
	\mathbf{K}_{x,x}\mathbf{V} \mathbf{K}_{y,y} + 
	\mathbf{K}_{x,x}\mathbf{V} \mathbf{K}_{y,y} \mathbf{V} \mathbf{K}_{x,x}
	\,,
\end{align}
where we have used the definition of $\hat{\mathbf{A}}$ in (\ref{eq:matrix_A}).
This expression only depends on the kernel matrices $\mathbf{K}_{x,x}$ and
$\mathbf{K}_{y,y}$ and the matrix $\mathbf{V}$, and can be computed with cost $\mathcal{O}(N^3)$.

\subsection{Centering the input data and centering and whitening the residuals}

An assumption made in Section \ref{sec:asymmetry} was that the samples of the
random variables $\mathcal{X}$ and $\mathcal{Y}$ are centered, \emph{i.e.}, they have
zero mean. In this section we show how to carry out this centering process in 
feature space. Furthermore, we also show how to center the residuals of the
fitting process, which are also whitened. Whitening is a standard procedure in
which the data are transformed to have the identity matrix as the covariance
matrix. It also corresponds to projecting the data onto all the principal
components, and scaling them to have unit standard deviation.

We show how to center the data in feature space. For this, we follow
\citep{scholkopf1997kernel} and work with:
\begin{align}
\tilde{\phi}(x_i) &= \phi(x_i) - \frac{1}{N} \sum_{j=1}^N \phi(x_j)\,, &
\tilde{\phi}(y_i) &= \phi(y_i) - \frac{1}{N} \sum_{j=1}^N \phi(y_j)\,.
\end{align}
The consequence is that now the kernel matrices $\mathbf{K}_{x,x}$ and
$\mathbf{K}_{y,y}$ are replaced by
\begin{align}
\tilde{\mathbf{K}}_{x,x} &= \mathbf{K}_{x,x} - \mathbf{1}_N \mathbf{K}_{x,x}
-  \mathbf{K}_{x,x} \mathbf{1}_N + \mathbf{1}_N \mathbf{K}_{x,x} \mathbf{1}_N
   \,, \nonumber \\
\tilde{\mathbf{K}}_{y,y} &= \mathbf{K}_{y,y} - \mathbf{1}_N \mathbf{K}_{y,y}
-  \mathbf{K}_{y,y} \mathbf{1}_N + \mathbf{1}_N \mathbf{K}_{y,y} \mathbf{1}_N\,,
\end{align}
where  $\mathbf{1}_N$ is a $N \times N$ matrix with all entries equal to $1/N$.
The residuals can be centered also in a similar way. Namely,
$\tilde{\mathbf{K}}_\epsilon = \mathbf{K}_\epsilon - \mathbf{1}_N \mathbf{K}_\epsilon
-  \mathbf{K}_\epsilon \mathbf{1}_N + \mathbf{1}_N \mathbf{K}_\epsilon \mathbf{1}_N$. 

We now explain the whitening of the residuals, which are now assumed to be
centered.  This process involves the computation of the eigenvalues and
eigenvectors of the $d \times d$ covariance matrix $\mathbf{C}$ of the
residuals. This is done as in kernel PCA \citep{scholkopf1997kernel}.  Denote
by $\tilde{\bm{\epsilon}}_i$ to the centered residuals.  The covariance matrix
is $\mathbf{C} = N^{-1} \sum_{i=1}^N \tilde{\bm{\epsilon}}_i
\tilde{\bm{\epsilon}}_i^\text{T}$.  The eigenvector expansion implies that
$\mathbf{C}\mathbf{v}_i = \lambda_i \mathbf{v}_i$, where $\mathbf{v}_i$ denotes
the $i$-th eigenvector and $\lambda_i$ the $i$-th eigenvalue. The consequence
is that $N^{-1}\sum_{k=1}^N \tilde{\bm{\epsilon}}_k
\tilde{\bm{\epsilon}}_k^\text{T} \mathbf{v}_i  = \lambda_i \mathbf{v}_i$.
Thus, the eigenvectors can be expressed as a combination of the residuals.
Namely, $\mathbf{v}_i = \sum_{j=1}^N b_{i,j} \tilde{\bm{\epsilon}}_j$, where
$b_{i,j} = N^{-1} \tilde{\bm{\epsilon}}_j^\text{T} \mathbf{v}_i$.  Substituting
this result in the previous equation we have that $N^{-1} \sum_{k=1}^N
\tilde{\bm{\epsilon}}_k\tilde{\bm{\epsilon}}_k^\text{T} \sum_{j=1}^N b_{i,j}
\tilde{\bm{\epsilon}}_j  = \lambda_i \sum_{j=1}^N b_{i,j}
\tilde{\bm{\epsilon}}_j$.  When we multiply both sides by
$\tilde{\bm{\epsilon}}_l^\text{T}$ we obtain $N^{-1} \sum_{k=1}^N
\tilde{\bm{\epsilon}}_l^\text{T}
\tilde{\bm{\epsilon}}_k\tilde{\bm{\epsilon}}_k^\text{T} \sum_{j=1}^N b_{i,j}
\tilde{\bm{\epsilon}}_j = \lambda_i \sum_{j=1}^N b_{i,j}
\tilde{\bm{\epsilon}}_l^\text{T}\tilde{\bm{\epsilon}}_j$, for $l=1,\ldots d$,
which is written in terms of kernels as $\tilde{\mathbf{K}}_\epsilon
\tilde{\mathbf{K}}_\epsilon \mathbf{b}_i = \lambda_i N
\tilde{\mathbf{K}}_\epsilon \mathbf{b}_i$, where $\mathbf{b}_i =
(b_{i,1},\ldots,b_{i,N})^\text{T}$.  A solution to this problem is found by
solving the eigenvalue problem $\tilde{\mathbf{K}}_\epsilon \mathbf{b}_i =
\lambda_i N \mathbf{b}_i$.  We also require that the eigenvectors have unit
norm. Thus, $1  = \mathbf{v}_i^\text{T} \mathbf{v}_i = \sum_{j=1}^N
\sum_{k=1}^N b_{i,j} b_{i,k} \tilde{\bm{\epsilon}}_j^\text{T}
\tilde{\bm{\epsilon}}_k = \mathbf{b}_i^\text{T} \tilde{\mathbf{K}}_\epsilon
\mathbf{b}_i = \lambda_i N  \mathbf{b}_i^\text{T}  \mathbf{b}_i$, which means
that $\mathbf{b}_i$ has norm $1 / \sqrt{\lambda_i N}$.  Consider now that
$\tilde{\mathbf{b}}_i$ is one eigenvector of $\tilde{\mathbf{K}}_\epsilon$.
Then, $\mathbf{b}_i = 1/\sqrt{\lambda_iN} \tilde{\mathbf{b}}_i$. Similarly, let
$\tilde{\lambda}_i$ be an eigenvalue of $\tilde{\mathbf{K}}_\epsilon$. Then
$\lambda_i = \tilde{\lambda}_i / N$.  In summary, $\lambda_i$ and $b_{i,j}$,
with $i=1,\ldots,N$ and $j=1,\ldots,N$ can be found with cost
$\mathcal{O}(N^3)$ by finding the eigendecomposition of
$\tilde{\mathbf{K}}_\epsilon$.

The whitening process is carried out by projecting each residual
$\tilde{\bm{\epsilon}}_k$ onto each eigenvector $\mathbf{v}_i$ and then
multiplying by $1 / \sqrt{\lambda_i}$.  The corresponding $i$-th component for
the $k$-th residual, denoted by $Z_{k,i}$, is $Z_{k,i} = 1 / \sqrt{\lambda_i}
\mathbf{v}_i^\text{T} \tilde{\bm{\epsilon}}_k =  1 / \sqrt{\lambda_i}
\sum_{j=1}^N b_{i,j} \tilde{\bm{\epsilon}}_j^\text{T} \tilde{\bm{\epsilon}}_k$,
and in consequence, the whitened residuals are $\mathbf{Z} =
\tilde{\mathbf{K}}_\epsilon \mathbf{B} \mathbf{D} = N \mathbf{B}
\mathbf{D}^{-1} = \sqrt{N} \tilde{\mathbf{B}}$, where $\mathbf{B}$ is a matrix
whose columns contain each $\mathbf{b}_i$, $\tilde{\mathbf{B}}$ is a matrix
whose columns contain each $\tilde{b}_i$ and $\mathbf{D}$ is a diagonal matrix
whose entries are equal to $1 / \sqrt{\lambda_i}$.  Each row of $\mathbf{Z}$
now contains the whitened residuals. 

\subsection{Inferring the most likely causal direction}
\label{sec:test}

After having trained the model and obtained the matrix of whitened residuals $\mathbf{Z}$ 
in each direction, a suitable Gaussianity test can be used to determine the correct
causal relation between the variables $\mathcal{X}$ and $\mathcal{Y}$. Given the theoretical 
results of Section \ref{sec:asymmetry} one may be tempted to use tests based
on entropy or cumulants estimation. Such tests may perform poorly in practice due to the difficulty of
estimating high-order cumulants or differential entropy.  In particular, the estimators of the cumulants 
involve high-order moments and hence, suffer from high variance. As a consequence, in our experiments we use 
a statistical test for Gaussianity based on the \emph{energy distance} \citep{Szekely2005}, which has 
good power, is robust to noise, and does not have any adjustable hyper-parameters. Furthermore,
in Appendix \ref{Appendix:B} we motivate that in the anti-causal direction one should also expect
a smaller energy distance to the Gaussian distribution.

Assume $\mathcal{X}$ and $\mathcal{Y}$ are two independent random variables whose probability distribution 
functions are $F(\cdot)$ and $G(\cdot)$. The energy distance between these distributions is defined as
\begin{align}
D^2(F,G) &= 2 \mathds{E}[||\mathcal{X}-\mathcal{Y}||] - \mathds{E}[||\mathcal{X} - \mathcal{X}'||] - 
\mathds{E}[||\mathcal{Y} - \mathcal{Y}'||]\,, 
\end{align}
where $||\cdot||$ denotes some norm, typically the $\ell_2$-norm; $\mathcal{X}$ and $\mathcal{X}'$ are 
independent and identically distributed (i.i.d.); $\mathcal{Y}$ and $\mathcal{Y}'$ are i.i.d; and $\mathds{E}$ denotes 
expected value. The energy distance satisfies all axioms of a metric and hence characterizes the
equality of distributions. Namely, $D^2(F,G)=0$ if and only if $F=G$. Furthermore, in the case of 
univariate random variables the energy distance is twice the Cram\'er-von Mises distance given by
$\int \left(F(x) - G(x)\right)^2 dx$. 

Assume $\mathbf{X}=(\mathbf{x}_1,\ldots,\mathbf{x}_N)^\text{T}$ is a matrix that contains 
$N$ random samples (one per each row of the matrix) from a $d$ dimensional 
random variable with probability density $f$. The statistic described for testing for 
Gaussianity, \emph{i.e.}, $\text{H}_0:f=\mathcal{N}(\cdot|\mathbf{0},\mathbf{I})$ vs  $\text{H}_1:f\neq \mathcal{N}(\cdot|\mathbf{0},\mathbf{I})$,
that is described in \citep{Szekely2005} is:
\begin{align}
\text{Energy}(\mathbf{X}) &= N \left(\frac{2}{N} \sum_{j=1}^N \mathds{E}[||\mathbf{x}_j - \mathcal{Y} ||] 
- \mathds{E}[|| \mathcal{Y} - \mathcal{Y}' ||] - \frac{1}{N^2} \sum_{j,k=1}^N || \mathbf{x}_j - \mathbf{x}_k||
\right)\,,
\label{eq:statistic}
\end{align}
where $\mathcal{Y}$ and $\mathcal{Y}'$ are independent random variables distributed as 
$\mathcal{N}(\cdot|\mathbf{0},\mathbf{I})$ and $\mathds{E}$ denotes expected value.
Furthermore, the required expectations with respect to the Gaussian random variables $\mathcal{Y}$ and $\mathcal{Y}'$ can be 
efficiently computed as described in \citep{Szekely2005}. The idea is that if $f$ is similar to a Gaussian density 
$\mathcal{N}(\cdot|0,\mathbf{I})$, then $\text{Energy}(\mathbf{X})$ is close to zero. Conversely, the null 
hypothesis $H_0$ is rejected for large values of $\text{Energy}(\mathbf{X})$.

The data to test for Gaussianity is in our case $\mathbf{Z}$, \emph{i.e.}, the matrix of whitened 
residuals, which has size $N \times N$.  Thus, the whitened residuals have $N$ dimensions. 
The direct introduction of these residuals into a statistical test for Gaussianity is not expected to provide 
meaningful results, as a consequence of the high dimensionality. Furthermore, in our experiments 
we have observed that it is often the case that a large part of the total variance is explained by the first 
principal component (see the supplementary material for evidence supporting this). That is, $\lambda_i$, \emph{i.e.}, 
the eigenvalue associated to the $i$-th principal component, is almost negligible for $i \ge 2$. 
Additionally, we motivate in Appendix \ref{Appendix:C} that one should also obtain more Gaussian residuals, after
projecting the data onto the first principal component, in terms of a reduction of the magnitude of the high-order cumulants.
Thus, in practice, we consider only the first principal component of the estimated residuals in feature space. This is the 
component $i$ with the largest associated eigenvalue $\lambda_i$. We denote such $N$-dimensional vector by $\mathbf{z}$.

Let $\mathbf{z}_{x\rightarrow y}$ be the vector of coefficients of the  first
principal component of the residuals in feature space when the linear 
fit is performed in the direction $\mathcal{X} \rightarrow \mathcal{Y}$. Let
$\mathbf{z}_{y\rightarrow x}$ be the vector of coefficients of the first principal
component of the residuals in feature space when the linear fit is carried out in the
direction $\mathcal{Y} \rightarrow \mathcal{X}$. We define the measure of
Gaussianization of the residuals as $\mathcal{G}=\text{Energy}(\mathbf{z}_{x\rightarrow y}) / N -
\text{Energy}(\mathbf{z}_{y\rightarrow x}) / N$, where $\text{Energy}(\cdot)$
computes the statistic of the energy distance test for Gaussianity described above. 
Note that we divide each statistic by $N$ to cancel the corresponding factor that is 
considered in (\ref{eq:statistic}). Since in this test larger values for the statistic corresponds
to larger deviations from Gaussianity,  if $\mathcal{G}>0$
the direction $\mathcal{X} \rightarrow \mathcal{Y}$ is expected to be
more likely the causal direction. Otherwise, the direction $\mathcal{Y} \rightarrow \mathcal{X}$ is preferred. 

The variance of $\mathcal{G}$ will depend on the sample size $N$.
Thus, ideally one should use the difference between the $p$-values associated to
each statistic as the confidence in the decision taken.  Unfortunately,
computing these $p$-values is expensive since the distribution of the statistic
under the null hypothesis must be approximated via random sampling. 
In our experiments we measure the confidence of the decision in terms of the absolute value of 
$\mathcal{G}$, which is faster to obtain and we have found to perform well in practice.

\subsection{Parameter tuning and error evaluation}

Assume that a squared exponential kernel is employed in the method described
above. This means that $k(x_i,x_j) = \exp \left(-\gamma(x_i-x_j)^2\right)$,
where $\gamma > 0$ is the bandwidth of the kernel.  The same is assumed for
$k(y_i,y_j)$.  Therefore, two hyper-parameters require adjustment in the method
described. These are the ridge regression regularization parameter $\tau$ and
the kernel bandwidth $\gamma$.  They must be tuned in some way to produce the
best possible fit in each direction.  The method chosen to guarantee
this is a grid search guided by a 10-fold cross-validation procedure, which
requires computing the squared prediction error over unseen data. In this
section we detail how to evaluate these errors.

Assume that $M$ new paired data instances are available for validation. Let the
two matrices $\bm{\Phi}_{y^\text{new}}=(\phi(y_1^\text{new}),\ldots,
\phi(y_M^\text{new}))$ and $\bm{\Phi}_{x^\text{new}}=(\phi(x_1^\text{new}),
\ldots,\phi(x_M^\text{new}))$ summarize these data.  Define
$\bm{\epsilon}_i^\text{new} = \phi(y_i^\text{new}) - \hat{\mathbf{A}}
\phi(x_i^\text{new})$.  After some algebra, it is possible to show that the sum
of squared errors for the new instances is:
{\small
\begin{align}
E =  \sum_{i=1}^M (\bm{\epsilon}_i^\text{new})^\text{T} \bm{\epsilon}_i^\text{new}
 & = \text{trace}\bigg(\mathbf{K}_{y^\text{new},y^\text{new}} - 
\mathbf{K}_{y^\text{new},y} \mathbf{V} \mathbf{K}_{x^\text{new},x}^\text{T} -
\nonumber \\
& \quad 
	\mathbf{K}_{x^\text{new},x} \mathbf{V} \mathbf{K}_{y^\text{new},y}^\text{T} + 
\mathbf{K}_{x^\text{new},x}\mathbf{V} \mathbf{K}_{y,y} \mathbf{V} \mathbf{K}_{x^\text{new},x}^\text{T}\bigg)\,.
\end{align}
}
where
$\mathbf{K}_{y^\text{new},y^\text{new}}=\bm{\Phi}_{y^\text{new}}^\text{T}\bm{\Phi}_{y^\text{new}}$,
$\mathbf{K}_{y^\text{new},y}=\bm{\Phi}_{y^\text{new}}^\text{T}\bm{\Phi}_{y}$
and $\mathbf{K}_{x,x^\text{new}}=\bm{\Phi}_x^\text{T}\bm{\Phi}_{x^\text{new}}$.

Of course, the new data must be centered before computing the error estimate.
This process is similar to the one described in the previous section.  In
particular, centering can be simply carried out by working with the modified
kernel matrices:
\begin{align}
\tilde{\mathbf{K}}_{x^\text{new},x} &= 
\mathbf{K}_{x^\text{new},x} - \mathbf{M}_N  \mathbf{K}_{x,x}
- \mathbf{K}_{x^\text{new},x}\mathbf{1}_N + \mathbf{M}_N \mathbf{K}_{x,x}\mathbf{1}_N
\,,
\nonumber \\
\tilde{\mathbf{K}}_{y^\text{new},y} &= 
\mathbf{K}_{y^\text{new},y} - \mathbf{M}_N  \mathbf{K}_{y,y}
- \mathbf{K}_{y^\text{new},y}\mathbf{1}_N + \mathbf{M}_N \mathbf{K}_{y,y}\mathbf{1}_N \,,
\nonumber\\
\tilde{\mathbf{K}}_{y^\text{new},y^\text{new}} &= 
\mathbf{K}_{y^\text{new},y^\text{new}} - \mathbf{M}_N  \mathbf{K}_{y,y^\text{new}}
- \mathbf{K}_{y^\text{new},y}\mathbf{M}_N^\text{T} + \mathbf{M}_N \mathbf{K}_{y,y}\mathbf{M}_N^\text{T} \,,
\end{align}
where $\mathbf{M}_N$ is a matrix of  size $M \times N$ with all components
equal to $1/N$.  In this process, the averages employed for the centering step
are computed using only the observed data.

A disadvantage of the squared error is that it strongly depends on the kernel
bandwidth parameter $\gamma$. This makes it difficult to choose this hyper-parameter 
in terms of such a performance measure. 
A better approach is to choose both $\gamma$ and $\tau$ in terms of the
explained variance by the model.  This is obtained as follows:
$\text{Explained-Variance} = 1 - \text{E} / M\text{Var}_\text{ynew}$, where $E$
denotes the squared prediction error and $\text{Var}_\text{ynew}$ the variance
of the targets. The computation of the error $E$ is done as described
previously and $\text{Var}_\text{ynew}$ is simply the average of the diagonal
entries in  $\tilde{\mathbf{K}}_{y^\text{new},y^\text{new}}$.

\subsection{Finding pre-images for illustrative purposes}

The kernel method described above expresses its solution as feature maps of the
original data points.  Since the feature map $\phi(\cdot)$ is usually
non-linear, we cannot guarantee the existence of a pre-image under
$\phi(\cdot)$. That is, a point $y$ such that
$\phi(y)=\hat{\mathbf{A}}\phi(x)$, for some input point $x$. An alternative to
amend this issue is to find approximate pre-images, which can be useful to
make predictions or plotting results \citep{Scholkopf2002}.  In this section we
describe how to find this approximate pre-images.

Assume that we have a new data instance $x_\text{new}$ for which we would like
to know the associated target value $y_\text{new}$, after our kernel model has
been fitted. The predicted value in feature space is:
\begin{align}
\phi(y_\text{new}) &= \bm{\Phi}_y\mathbf{V} \bm{\Phi}_x^\text{T}\phi(x_\text{new})
= \bm{\Phi}_y\mathbf{V} \mathbf{k}_{x,x_\text{new}}
= \sum_{i=1}^n \alpha_i \phi(y_i)\,,
\end{align}
where  $\mathbf{k}_{x,x_\text{new}}$ contains the kernel evaluations between
each entry in $\mathbf{x}$ (\emph{i.e.}, the observed samples of the random variable
$\mathcal{X}$) and the new instance. Finally, each $\alpha_i$ is given by a
component of the vector $\mathbf{V} \mathbf{k}_{x,x_\text{new}}$.  The
approximate pre-image of $\phi(y_\text{new})$, $y_\text{new}$, is found by
solving the following optimization problem:
\begin{align}
y_\text{new} & = \underset{u}{\text{arg min}}  \quad ||\phi(y_\text{new}) -
\phi(u)||^2_2 
 = \underset{u}{\text{arg min}}    \quad 
	- 2 \bm{\alpha}^\text{T}\mathbf{k}_{y,u} + k(u,u)\,,
\end{align}
where $\mathbf{k}_{y,u}$ is a vector with the kernel values between each $y_i$
and $u$, and $k(u,u)=\phi(u)^\text{T}\phi(u)$. This is a non-linear
optimization problem than can be solved approximately using standard techniques
such as gradient descent. In particular, the computation of the gradient of
$\mathbf{k}_{y,u}$ with respect to $u$ is very simple in the case of the
squared exponential kernel.

\section{Data transformation and detailed causal inference algorithm}
\label{sec:algorithm}

The method for causal inference described in the previous section relies on the
fact that both random variables $\mathcal{X}$ and $\mathcal{Y}$ are equally
distributed. In particular, if this is the case, $\phi(x_i)$ and $\phi(y_i)$, \emph{i.e.},
the maps of $x_i$ and $y_i$ in the expanded feature space, will also be equally distributed.
This means that under such circumstances one should expect residuals that are more Gaussian 
in the anti-causal direction due to a reduction of the magnitude of the high order cumulants and
an increment of the differential entropy. The requirement that $\mathcal{X}$ and $\mathcal{Y}$ are equally
distributed can be easily fulfilled in the case of continuous univariate data by transforming 
$\mathbf{x}$, the samples of $\mathcal{X}$, to have the same empirical distribution as $\mathbf{y}$, 
the samples of $\mathcal{Y}$. 

Consider $\mathbf{x}=(x_1,\ldots,x_N)^\text{T}$ and $\mathbf{y}
= (y_1,\ldots,y_N)^\text{T}$ to be $N$ paired samples of $\mathcal{X}$ and
$\mathcal{Y}$, respectively.  To guarantee the same distribution for these samples
we only have to replace $\mathbf{x}$ by
$\tilde{\mathbf{x}}$, where each component of $\tilde{\mathbf{x}}$,
$\tilde{x}_i$, is given by $\tilde{x}_i= \hat{F}^{-1}_y(\hat{F}_x(x_i))$, with
$\hat{F}_y^{-1}(\cdot)$ the empirical quantile distribution function of the
random variable $\mathcal{Y}$, estimated using $\mathbf{y}$. Similarly,
$\hat{F}_x(\cdot)$ is the empirical cumulative distribution function of
$\mathcal{X}$, estimated using $\mathbf{x}$. This operation is known as the
probability integral transform.

One may wonder why should $\mathbf{x}$ be transformed instead of $\mathbf{y}$.
The reason is that by transforming $\mathbf{x}$ the additive noise hypothesis
made in (\ref{eq:model_one_d})  and (\ref{eq:model_multi}) is preserved.  
In particular, we have that $y_i = f(\hat{F}^{-1}_x(\hat{F}_y(\tilde{x}_i))) + \epsilon_i$.  
On the other hand, if $\mathbf{y}$ is transformed instead, the additive noise
model will generally not be valid anymore. More precisely, the transformation that 
computes $\tilde{\mathbf{y}}$ in such a way that it is distributed as $\mathbf{x}$ is 
$\tilde{y}_i= \hat{F}^{-1}_x(\hat{F}_y(y_i))$, $\forall i$. Thus, under this 
transformation we have that $\tilde{y}_i = \hat{F}^{-1}_x(\hat{F}_y(f(x_i) + \epsilon_i))$,  
which will lead to the violation of the additive noise model.

Of course, transforming $\mathbf{x}$ requires the knowledge of the causal direction.  In 
practice, we will transform both $\mathbf{x}$ and $\mathbf{y}$ and consider that the
correct transformation is the one that leads to the highest level of
Gaussianization of the residuals in the feature space, after fitting the model
in each direction. That is, the transformation that leads to the highest
value of $\mathcal{G}$ is expected to be the correct one. 
We expect that when $\mathbf{y}$ is transformed instead of $\mathbf{x}$, the
Gaussianization effect of the residuals is not as high as when $\mathbf{x}$ is
transformed, as a consequence of the violation of the additive noise model. 
This will allow to determine the causal direction. We do not have a theoretical 
result confirming this statement, but the good results obtained in 
Section \ref{sec:synthetic_exp} indicate that this is the case.

The details of the complete causal inference algorithm proposed are given in
Algorithm \ref{alg1}. Besides a causal direction, \emph{e.g.}, $\mathcal{X}
\rightarrow \mathcal{Y}$ or $\mathcal{Y} \rightarrow \mathcal{X}$, this
algorithm also outputs a confidence level in the decision made which is defined
as $\text{max}(|\mathcal{G}_{\tilde{x}}|, |\mathcal{G}_{\tilde{y}}|)$, where
$\mathcal{G}_{\tilde{x}}=\text{Energy}(\mathbf{z}_{\tilde{x}\rightarrow y}) / N - 
\text{Energy}(\mathbf{z}_{y \rightarrow \tilde{x}}) / N$ 
denotes the estimated level of Gaussianization of the residuals 
when $\mathbf{x}$ is transformed to have the same distribution as $\mathbf{y}$.
Similarly, $\mathcal{G}_{\tilde{y}}=\text{Energy}(\mathbf{z}_{\tilde{y}\rightarrow x}) / N - 
\text{Energy}(\mathbf{z}_{\tilde{y}\rightarrow x}) / N$ 
denotes the estimated level of Gaussianization of the residuals 
when $\mathbf{x}$ and $\mathbf{y}$ are swapped and $\mathbf{y}$ 
is transformed to have the same distribution as $\mathbf{x}$. 
Here $\mathbf{z}_{\tilde{x} \rightarrow y}$ contains the first 
principal component of the residuals in the expanded feature space when trying 
to predict $\mathbf{y}$ using $\tilde{\mathbf{x}}$. The same applies for 
$\mathbf{z}_{y \rightarrow \tilde{x}}$, $\mathbf{z}_{\tilde{y} \rightarrow x}$ and
$\mathbf{z}_{x \rightarrow \tilde{y}}$. However, the residuals are obtained this time 
when trying to predict $\tilde{\mathbf{x}}$ using $\mathbf{y}$, when trying to predict 
$\mathbf{x}$ using $\tilde{\mathbf{y}}$ and when trying to predict $\tilde{\mathbf{y}}$ using $\mathbf{x}$, 
respectively. Recall that the reason for keeping only the first principal component of 
the residuals is described in Section \ref{sec:test}.

Assume $|\mathcal{G}_{\tilde{x}}|> |\mathcal{G}_{\tilde{y}}|$. In this case we prefer the 
transformation of $\mathbf{x}$ to guarantee that the cause and the effect have the same distribution.
The reason is that it leads to a higher level of Gaussianization of the residuals, as estimated by the energy 
statistical test. Now consider that $\mathcal{G}_{\tilde{x}}>0$. We prefer the direction 
$\mathcal{X}\rightarrow\mathcal{Y}$ because the residuals of a fit in that direction 
are less Gaussian and hence have a higher value of the statistic of the energy test.
By contrast, if $\mathcal{G}_{\tilde{x}}<0$ we prefer the direction 
$\mathcal{Y}\rightarrow\mathcal{X}$ for the same reason. In the case that 
$|\mathcal{G}_{\tilde{x}}|<|\mathcal{G}_{\tilde{y}}|$ the reasoning is the same 
and we prefer the transformation of $\mathbf{y}$.
However, because we have swapped $\mathbf{x}$ and $\mathbf{y}$ for computing $\mathcal{G}_{\tilde{y}}$, 
the decision is the opposite as the previous one. Namely, if $\mathcal{G}_{\tilde{y}}>0$ we prefer 
the direction $\mathcal{Y}\rightarrow \mathcal{X}$ and otherwise we prefer 
the direction $\mathcal{X}\rightarrow\mathcal{Y}$. The confidence in the decision (\emph{i.e.}, the estimated level of 
Gaussianization) is always measured by $\text{max}(|\mathcal{G}_{\tilde{x}}|, |\mathcal{G}_{\tilde{y}}|)$.

\RestyleAlgo{boxruled}
\IncMargin{1.2em}
\begin{algorithm}[ht]
\LinesNumbered
{\small
\SetKwData{Left}{left}\SetKwData{This}{this}\SetKwData{Up}{up}
\SetKwFunction{Union}{Union}\SetKwFunction{FindCompress}{FindCompress}
\SetKwInOut{Input}{input}\SetKwInOut{Output}{output}
\KwData{Paired samples $\mathbf{x}$ and $\mathbf{y}$ from the random variables $\mathcal{X}$ and $\mathcal{Y}$.}
\KwResult{An estimated causal direction alongside with a confidence level.}
\BlankLine
Standardize $\mathbf{x}$ and $\mathbf{y}$ to have zero mean and unit variance\;
\BlankLine
Transform $\mathbf{x}$ to compute $\tilde{\mathbf{x}}$ \tcp*[r]{This guarantees that $\tilde{\mathbf{x}}$ is distributed as $\mathbf{y}$.}
$\hat{\mathbf{A}}_{\tilde{x}\rightarrow y} 
\leftarrow \text{FitModel}(\tilde{\mathbf{x}},\mathbf{y})$ \tcp*[r]{This also finds the hyper-parameters $\tau$ and $\gamma$.}
$\mathbf{z}_{\tilde{x}\rightarrow y} \leftarrow \text{ObtainResiduals}(\tilde{\mathbf{x}},\mathbf{y}, \hat{\mathbf{A}}_{\tilde{x}\rightarrow y} )$ 
\tcp*[r]{First PCA component in feature space.}
$\hat{\mathbf{A}}_{y\rightarrow \tilde{x}} \leftarrow \text{FitModel}(\mathbf{y},\tilde{\mathbf{x}})$ 
\tcp*[r]{Fit the model in the other direction}
$\mathbf{z}_{y\rightarrow\tilde{x}} \leftarrow \text{ObtainResiduals}(\mathbf{y}, \tilde{\mathbf{x}},\hat{\mathbf{A}}_{y \rightarrow \tilde{x}})$ 
\tcp*[r]{First PCA component in feature space.}
\BlankLine
$\mathcal{G}_{\tilde{x}} \leftarrow \text{Energy}(\mathbf{z}_{\tilde{x}\rightarrow y}) / N - \text{Energy}(\mathbf{z}_{y \rightarrow \tilde{x}}) / N$
\tcp*[r]{Get the Gaussianization level.}
\BlankLine
Swap $\mathbf{x}$ and $\mathbf{y}$ and repeat lines 2-7 of the algorithm to compute $\mathcal{G}_{\tilde{y}}$.
\BlankLine
\eIf{$|\mathcal{G}_{\tilde{x}}| > |\mathcal{G}_{\tilde{y}}|$}{
	\eIf{$\mathcal{G}_{\tilde{x}} > 0 $}{{\bf Output}: $\mathcal{X} \rightarrow \mathcal{Y}$ with confidence $|\mathcal{G}_{\tilde{x}}|$}
	{{\bf Output}: $\mathcal{Y} \rightarrow \mathcal{X}$ with confidence $|\mathcal{G}_{\tilde{x}}|$}
}{
	\eIf{$\mathcal{G}_{\tilde{y}} > 0 $}{{\bf Output}: $\mathcal{Y} \rightarrow \mathcal{X}$ with confidence $|\mathcal{G}_{\tilde{y}}|$}
	{{\bf Output}: $\mathcal{X} \rightarrow \mathcal{Y}$ with confidence $|\mathcal{G}_{\tilde{y}}|$}
}
}
\caption{{\small Causal Inference Based on the Gaussianity of the Residuals (GR-AN)}} \label{alg1}

\end{algorithm}\DecMargin{1.2em}

The algorithm uses a squared exponential kernel with 
bandwidth parameter $\gamma$ and the actual matrices
$\hat{\mathbf{A}}_{\tilde{x}\rightarrow y}$ and $\hat{\mathbf{A}}_{y
\rightarrow \tilde{x}}$, of potentially infinite dimensions, need not be
evaluated in closed form in practice. 
As indicated in Section \ref{sec:feature_expansion}, all computations 
are carried out efficiently with cost $\mathcal{O}(N^3)$ using inner products, which are
evaluated in terms of the corresponding kernel function. All
hyper-parameters, \emph{i.e.}, $\tau$ and $\gamma$, are chosen using a grid search
method guided by a 10-fold cross-validation process. This search maximizes the
explained variance of the left-out data and 10 potential values are considered
for both $\tau$ and $\gamma$.

\section{Related work}
\label{sec:related_work}

The Gaussianity of residuals was first employed for causal inference by
\citep{hernandezLobato11}.  These authors analyze auto-regressive (AR) processes
and show that a similar asymmetry as the one described in this paper can be
used to determine the temporal direction of a time series in the presence of
non-Gaussian noise. Namely, when fitting an AR process to a reversed time
series, the residuals obtained follow a distribution that is closer to a
Gaussian distribution. Nevertheless, unlike the work described here, the method
proposed by \cite{hernandezLobato11} cannot be used to tackle multidimensional or
non-linear causal inference problems.  In their work, \citet{hernandezLobato11}
show some advantages of using statistical tests based on measures of
Gaussianity to determine the temporal direction of a time series, as a
practical alternative to statistical tests based on the independence of the
cause and the residual. The motivation for these advantages is that the former
tests are one-sample tests while the later ones are two-sample tests.

The previous paper is extended by \citet{morales13} to consider
multidimensional AR processes. However, this work lacks a theoretical result
that guarantees that the residuals obtained when fitting a vectorial AR process in
the reversed (anti-chronological) direction will follow a distribution closer to a Gaussian
distribution.  In spite of this issue, extensive experiments with simulated
data suggest the validity of such conjecture. Furthermore, a series of
experiments show the superior results of the proposed rule to determine the
direction of time, which is based on measures of Gaussianity, and compared with
other state-of-the-art methods based on tests of independence.

The problem of causal inference under continuous-valued data has also been
analyzed by \cite{Shimizu06}.  The authors propose a method called LINGAM that
can identify the causal order of several variables when assuming that (a) the
data generating process is linear, (b) there are no unobserved co-founders, and
(c) the disturbance variables have non-Gaussian distributions with non-zero
variances. These assumptions are required because LINGAM relies on the use of
Independent Component Analysis (ICA).  More specifically, let $\mathbf{x}$
denote a vector that contains the variables we would like to determine the
causal order of.  LINGAM assumes that $\mathbf{x} = \mathbf{B} \mathbf{x} +
\mathbf{e}$, where $\mathbf{B}$ is a matrix that can be permuted to strict
lower triangularity if one knows the actual causal ordering in $\mathbf{x}$,
and $\mathbf{e}$ is a vector of non-Gaussian independent disturbance variables.
Solving for $\mathbf{x}$, one gets $\mathbf{x}=\mathbf{A}\mathbf{e}$, where
$\mathbf{A}=(\mathbf{I}-\mathbf{B})^{-1}$.  The $\mathbf{A}$ matrix can be
inferred using ICA. Furthermore, given an estimate of $\mathbf{A}$,
$\mathbf{B}$ can be obtained to find the corresponding connection strengths
among the observed variables, which can then be used to determine the true
causal ordering. LINGAM has been extended to consider linear relations among 
groups of variables in \citep{entner_hoyer2012,Kawahara2010}. 

In real-world data, causal relationships tend to be non-linear, a fact that
questions the usefulness of linear methods.  \cite{Hoyeretal08} show that a
basic linear framework for causal inference can be generalized to non-linear
models. For non-linear models with additive noise, almost any non-linearities
(invertible or not) will typically yield identifiable models. In particular,
\cite{Hoyeretal08} assume that $y_i = f(x_i) + \epsilon_i$, where $f(\cdot)$
is a possibly non-linear function, $x_i$ is the cause variable, and
$\epsilon_i$ is some independent and random noise. The proposed causal
inference mechanism consists in performing a non-linear regression on the data
to get an estimate of $f(\cdot)$, $\hat{f}(\cdot)$, and then calculate the
corresponding residuals $\hat{\epsilon}_i = y_i - \hat{f}(x_i)$.  Then, one may
test whether $\hat{\epsilon}_i$ is independent of $x_i$ or not. The same
process is repeated in the other direction. The direction with
the highest level of independence is chosen as the causal one.  In practice,
the estimate $\hat{f}(\cdot)$ is obtained using Gaussian
processes for regression, and the HSIC test \citep{NIPS2007} is used as the
independence criterion. This method has obtained good performance results
\citep{Janzing12} and it has been extended in \citep{zhang2009} to address
problems where the model is  $y_i = h(f(x_i) + \epsilon_i)$, for some
invertible function $h(\cdot)$. A practical difficulty is however 
that such a model is significantly harder to fit to the data.

In \citet{NIPS2010}, a method for causal inference is proposed based on a
latent variable model, used to incorporate the effects of un-observed noise. In
this context, it is considered that the effect variable is a function of the
cause variable and an independent noise term, not necessarily additive, that
is, $y_i = f(x_i,\epsilon_i)$, where $x_i$ is the cause variable and
$\epsilon_i$ is some independent and random noise. The causal direction is then
inferred using standard Bayesian model selection. In particular, the preferred
direction is the one under which the corresponding model has the largest
marginal likelihood, where the marginal likelihood is understood as a proxy for
the Kolmogorov complexity.  This method suffers from several implementation
difficulties, including the intractability of the marginal likelihood
computation.  However, it has shown encouraging results on synthetic and
real-world data.

\citet{icml2010} consider the problem of inferring linear causal relations
among multi-dimensional variables. The key point here is to use an asymmetry
between the distributions of the cause and the effect that occurs if the
covariance matrix of the cause and the matrix mapping the cause to the effect
are independently chosen. This method exhibits the nice property that applies
to both deterministic and stochastic causal relations, provided that the
dimensionality of the involved random variables is sufficiently high. The
method assumes that $\mathbf{y}_i = \mathbf{A} \mathbf{x}_i + \bm{\epsilon}_i$,
where $\mathbf{x}_i$ is the cause and $\bm{\epsilon}_i$ is additive noise.
Namely, denote with $\hat{\bm{\Sigma}}$ to the empirical covariance matrix of
the variables in each $\mathbf{x}_i$. Given an estimate of $\mathbf{A}$,
$\hat{\mathbf{A}}$, the method computes
$\bm{\Delta}_{\mathbf{x}\rightarrow\mathbf{y}} = \log
\text{trace}(\hat{\mathbf{A}} \hat{\bm{\Sigma}} \hat{\mathbf{A}}^\text{T}) 
- \log \text{trace}(\hat{\bm{\Sigma}})  + \log \text{trace}(\hat{\mathbf{A}}
\hat{\mathbf{A}}^\text{T}) + d$, where $d$ is the dimension of
$\mathbf{x}_i$. This process is repeated to compute
$\bm{\Delta}_{\mathbf{y}\rightarrow\mathbf{x}}$ where $\mathbf{x}_i$ and
$\mathbf{y}_i$ are swapped.  The asymmetry described states that
$\bm{\Delta}_{\mathbf{x}\rightarrow\mathbf{y}}$ should be close to zero while
$\bm{\Delta}_{\mathbf{y}\rightarrow\mathbf{x}}$ should not.  Thus, if
$|\bm{\Delta}_{\mathbf{x}\rightarrow\mathbf{y}}| >
|\bm{\Delta}_{\mathbf{y}\rightarrow\mathbf{x}}|$, $\mathbf{x}_i$ is expected
to be the cause. Otherwise, the variables in $\mathbf{y}_i$ are predicted to
be cause instead.  Finally, a kernelized version of this method is also
described in \citet{chen2013nonlinear}.

Most of the methods introduced in this section assume some form of noise in the
generative process of the effect.  Thus, their use is not justified in the case
of noiseless data. \citet{Janzing12} describe a method to deal with
these situations. In particular, the method makes use of information geometry
to identify an asymmetry that can be used for causal inference. The asymmetry
relies on the idea that the marginal distribution of the cause variable,
denoted by $p(x)$, is expected to be chosen independently from the mapping
mechanism producing the effect variable, denoted by the conditional
distribution $p(y|x)$. Independence is defined here as orthogonality in the
information space, which allows to describe a dependence that occurs between
$p(y)$ and $p(x|y)$ in the anti-causal direction. This dependence can be
then used to determine the causal order. A nice property of this method
is that this asymmetry between the cause and the effect becomes very simple if
both random variables are deterministically related.  Remarkably, the method
also performs very well in noisy scenarios, although no theoretical guarantees
are provided in this case. 

A similar method for causal inference to the last one is described by
\cite{ChenZCS14}.  These authors also consider that $p(x)$ and $p(y|x)$ fulfil
some sort of independence condition, and that this independence condition does
not hold for the anti-causal direction.  Based on this, they define an
uncorrelatedness criterion between $p(x)$ and $p(y|x)$, and show an asymmetry
between the cause and the effect in terms of a certain complexity metric on
$p(x)$ and $p(y|x)$, which is less than the same complexity metric on $p(y)$
and $p(x|y)$.  The complexity metric is calculated in terms of a reproducing
kernel Hilbert space embedding (EMD) of probability distributions. Based on the
complexity metric, the authors propose an efficient kernel-based algorithm for
causal discovery. 

In Section \ref{sec:information_theory} we have shown that in the multivariate case one 
should expect higher entropies in the anti-causal direction. Similar results have been 
obtained in the case of non-linear relations and the univariate data case 
\citep{HyvarinenS13,KpotufeSJS14}. Assume $x,y\in\mathds{R}$ and the actual causal model to be $y = f(x) + d$, with 
$x \bot d$ and $f(\cdot)$ an arbitrary function. Let $e$ be the residual of a fit performed 
in the anti-causal direction. \cite[Sec. 5.2]{HyvarinenS13} shows that the likelihood ratio $R$ 
of each model (\emph{i.e.}, the model fitted in the causal direction and the model fitted in the anti-causal 
direction) converges in the presence of infinite data to the difference between the sum of the 
entropies of the independent variable and the residual in each direction. Namely, 
$R \rightarrow - H(x) - H(d/\sigma_d) + H(y) + H(e /\sigma_e) + \log \sigma_d - \log\sigma_e$,
where $\sigma_d$ and $\sigma_e$ denote the standard deviation of the errors in each 
direction. If $R > 0$, the causal direction is chosen. By contrast, if $R < 0$ 
the anti-causal direction is preferred. The process of evaluating R involves the estimation of 
the entropies of four univariate random variables, \emph{i.e.}, $x$, $d$, $y$ and $e$ and the 
standard deviation of the errors $d$ and $e$, which need not be equal. The non-linear functions
are estimated as in \citep{Hoyeretal08} using a Gaussian process. The entropies are obtained 
using a maximum entropy approximation under the hypothesis that the distributions of these 
variables are not far from Gaussian \citep{Hyvarinen1998}. The resulting method is called non-linear
maximum entropy (NLME). A practical difficulty is however that the estimation of the entropy 
is a very difficult task, even in one dimension \citep{beirlant++_1997_nonparametric}. Thus, the NLME method is adapted in 
an \emph{ad-hoc} manner with the aim of obtaining better results in certain 
difficult situations with sparse residuals. More precisely, if $H(x)$ and $H(y)$ are 
ignored and Laplacian residuals are assumed $R \rightarrow \log\sigma_e -\log \sigma_d $. That is, the model with the 
minimum error is preferred. The errors are estimated however in terms of the absolute deviations 
(because of the Laplacian assumption). This method is called mean absolute deviation (MAD).
Finally, \cite{KpotufeSJS14} show the consistency of the noise additive model, give a formal proof 
for $R\geq0$ (see Lemma 1), and propose to estimate $H(x)$, $H(y)$, $H(d)$ and $H(e)$ 
using kernel density estimators. Note that if $x$ and $y$ are equally distributed, $H(x)=H(y)$ 
and the condition $R\geq 0$ implies $H(e) \geq H(d)$. Nevertheless, $\sigma_d$ and $\sigma_e$ are in 
general different (see the supplementary material for an illustrative example). This means that 
in the approach of \cite{HyvarinenS13} and \cite{KpotufeSJS14} it is not possible to make a decision 
directly on the basis of a Gaussianization effect on the residuals.

The proposed method GR-AN, introduced in Section \ref{sec:algorithm}, differs from the 
approaches described in the previous paragraph in that it does not have to deal with the estimation 
of four univariate entropies, which can be a particularly difficult task. By contrast, it relies on statistical 
tests of deviation from Gaussianity to infer the causal direction. Furthermore, the tests employed in 
our method need not be directly related to entropy estimation. This is particularly the case of the energy 
test suggested in Section \ref{sec:test}. Not having to estimate differential entropies is an advantage 
of our method confirmed by the results that are obtained in the experiments section. In particular, we have 
empirically observed that that GR-AN performs better than the two methods for causal inference NLME and 
MAD that have been described in the previous paragraph. GR-AN also performs better than GR-ENT, a method
that uses, instead of statistical tests of Gaussianity, a non-parametric estimator of the 
entropy \citep{singh2003nearest}.

\section{Experiments}
\label{sec:experiments}

We carry out experiments to validate the method proposed in this paper, and
empirically verify that the model residuals in the anti-causal direction 
are more Gaussian that the model residuals in the causal direction due to a reduction
of the high-order cumulants and an increment of the differential entropy.  
From now on, we refer to our method as GR-AN (Gaussianity of
the Residuals under Additive Noise).  Furthermore, we compare the performance
of GR-AN with four other approaches from the literature on causal inference,
reviewed in Section \ref{sec:related_work}. First, LINGAM \citep{Shimizu06}, a
method which assumes an additive noise model, but looks for independence
between the cause and the residuals.  Second, IR-AN (Independence of the
Residuals under Additive Noise), by \citet{Hoyeretal08}. Third, a method based on
information geometry, IGCI \citep{Janzing12}. Fourth, a method based on
Reproducing Kernel Hilbert Space Embeddings (EMD) of probability distributions
\citep{ChenZCS14}. Fifth, the two methods for non-linear causal inference 
based on entropy estimation described in \citep{HyvarinenS13}, NLME and MAD.
Sixth, the same GR-AN method, but where we omit the transformation to
guarantee that the random variables $\mathcal{X}$ and $\mathcal{Y}$ are 
equally distributed. This method is called $\text{GR-AN}^\star$.
Last, we also compare results with two variants of GR-AN that are not based on the 
energy distance to measure the level of Gaussianity of the residuals. These are
GR-K4, which uses the empirical estimate of the fourth cumulant (kurtosis) to determine
the causal direction (it chooses the direction with the largest estimated fourth cumulant);
and GR-ENT, which uses a non-parametric estimator of the entropy \citep{singh2003nearest} 
to determine the causal direction (the direction with the smallest entropy is preferred);

The hyper-parameters of the different methods are set as follows. In LINGAM,
we use the parameters recommended by the implementation provided by the
authors.  In IR-AN, NLME and MAD we employ a Gaussian process whose hyper-parameters 
are found by type-II maximum likelihood.  Furthermore, in IR-AN the HSIC test is
used to assess independence between the causes and the residuals. 
In NLME the entropy estimator is the one described in \citep{Hyvarinen1998}.
In IGCI, we test different normalizations (uniform and Gaussian) and different criteria
(entropy or integral) and report the best observed result. In EMD and
synthetic data, we follow \citep{ChenZCS14} to select the hyper-parameters. In
EMD and real-world data, we evaluate different hyper-parameters and report the
results for the best combination found. In GR-AN, GR-K4 and GR-ENT the 
hyper-parameters are found via cross-validation, as described in Section \ref{sec:algorithm}.
The number of neighbors in the entropy estimator of GR-ENT is set to 10, a 
value that we have observed to give a good trade-off between bias and variance.
Finally, in GR-AN, GR-K4, and GR-ENT we transform the data so that both variables are 
equally distributed, as indicated in Section \ref{sec:algorithm}.

The confidence in the decision is computed as indicated in \citep{Janzing12}. 
More precisely, in LINGAM the confidence is given by the maximum absolute value 
of the entries in the connection strength matrix $\mathbf{B}$. In IGCI we employ the 
absolute value of the difference between the corresponding estimates (entropy or integral) 
in each direction. In IR-AN the confidence level is obtained as the maximum of the two 
$p$-values of the HSIC test. In EMD we use the absolute value of the difference between 
the estimates of the corresponding complexity metric in each direction, as described in 
\citep{ChenZCS14}. In NLME and MAD the confidence level is given by the absolute value 
of the difference between the outputs of the entropy estimators in each direction 
\citep{HyvarinenS13}. In GR-K4 we use the absolute difference between the estimated fourth 
cumulants. In GR-ENT we use the absolute difference between the estimates of the entropy.
Finally, in GR-AN we follow the details given in Section \ref{sec:algorithm} 
to estimate the confidence in the decision.

To guarantee the exact reproducibility of the different experiments described
in this paper, the source-code for all methods and datasets is available in the
public repository \url{https://bitbucket.org/dhernand/gr_causal_inference}.

\subsection{Experiments with synthetic data}
\label{sec:synthetic_exp}

We carry out a first batch of experiments on synthetic data. In these
experiments, we employ the four causal mechanisms that map $\mathcal{X}$ to
$\mathcal{Y}$ described by \cite{ChenZCS14}.  They involve linear
and non-linear functions, and additive and multiplicative noise effects:
\begin{itemize}
\item $M_1$: $y_i=0.8x_i + \epsilon_i$.
\item $M_2$: $y_i=x_i \epsilon_i$.
\item $M_3$: $y_i=0.3x_i^3 + \epsilon_i$.
\item $M_4$: $y_i=\text{atan}(x_i)^3 + \epsilon_i$.
\end{itemize}
The noise $\epsilon_i$ can follow four different types of distributions: (i) A
generalized Gaussian distribution with shape parameter equal to 10 (an example
of a sub-Gaussian distribution); (ii) a Laplace distribution (an example of a
super-Gaussian distribution); (iii) a Gaussian distribution; and (iv) a bimodal 
distribution with density $p(\epsilon_i)=0.5 \mathcal{N}(\epsilon_i|m,s) + 0.5 \mathcal{N}(\epsilon_i|-m,s)$, where $m=.63$ 
and $s=.1$. The Laplace distribution and the Gaussian distribution are adjusted to have the 
same variance as the generalized Gaussian distribution. The bimodal distribution already has the same
variance as the generalized Gaussian distribution.

As indicated by \cite{ChenZCS14}, in these experiments, the samples from the
cause variable $\mathcal{X}$ are generated from three potential distributions:
\begin{itemize}
\item $p_1(x) = \frac{1}{\sqrt{2 \pi}} \exp\{-x^2/2\}$.
\item $p_2(x) = \frac{1}{2\sqrt{0.5 \pi}} \exp\{-(x+1)^2/0.5\} +
\frac{1}{2\sqrt{0.5 \pi}} \exp\{-(x-1)^2/0.5\}$.
\item $p_3(x) = \frac{1}{4\sqrt{0.5 \pi}} \exp\{-(x+1.5)^2/0.5\} +
\frac{1}{2\sqrt{0.5 \pi}} \exp\{-x^2/0.5\} + \frac{1}{4\sqrt{0.5 \pi}}
\exp\{-(x-1.5)^2/0.5\}$.
\end{itemize}
These are unimodal, bimodal, and trimodal distributions, respectively.

\begin{figure}[tb]
\begin{center}
\includegraphics[width=\textwidth]{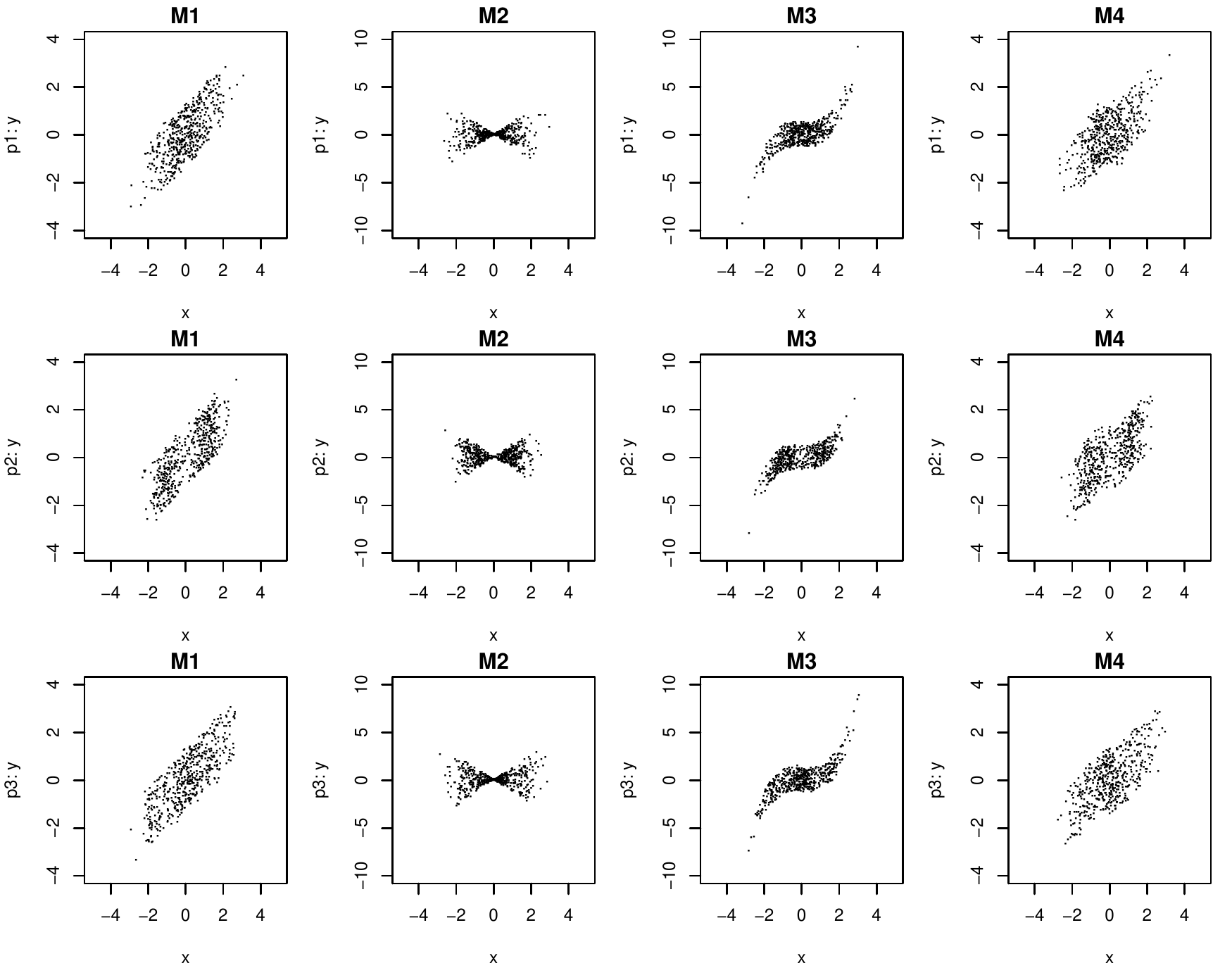} 
\end{center}
\caption{Plots of different distributions and mechanisms when the noise follows
a generalized Gaussian distribution with shape parameter equal to 10.}
\label{fig:samples_mechanisms}
\end{figure}

Figure \ref{fig:samples_mechanisms} displays a representative example of the
plots of different combinations of distributions and mapping mechanisms when
the noise follows a generalized Gaussian distribution with shape parameter
equal to 10. The plots for Laplace, Gaussian or bimodal distributed noise look similar to these ones.
The assumptions made by proposed method, \emph{i.e.}, GR-AN, are valid in the case
of all the causal mechanisms, except for M2, which considers multiplicative 
noise, and in the case of all cause distributions, $p_1$, $p_2$ and $p_3$. The only 
type of noise that violates the assumptions made by GR-AN is the case of Gaussian 
noise. In particular, under Gaussian noise GR-AN cannot infer the causal direction 
using Gaussianity measures because the actual residuals are already Gaussian.

The average results of each method on 100 repetitions of each potential causal
mechanism, distribution for the effect, and noise distribution are displayed in
Table \ref{tab1}. The size of each paired samples of $\mathcal{X}$ and $\mathcal{Y}$ is set to
$500$ in these experiments. We observe that when the assumptions made by the proposed 
method, GR-AN, are satisfied, it identifies the causal direction on a very high fraction 
of the 100 repetitions considered. However, when these assumptions are not valid, \emph{e.g.}, 
in the case of the M2 causal mechanism, which has multiplicative noise, the performance worsens. The same
happens when the distribution of the residuals is Gaussian. In these
experiments, LINGAM tends to fail when the causal relation is strongly
non-linear.  This is the case of the causal mechanism M3. LINGAM also has
problems when all the independent variables are Gaussian.  
Furthermore, all methods generally fail in the case of independent Gaussian variables that are 
linearly related. This corresponds to the causal mechanism M1, the distribution $p_1(x)$ for the cause,
and the Gaussian distribution for the noise. 
The reason for this is that in this particular scenario the causal direction is not identifiable \citep{Shimizu06}.  IGCI and EMD
sometimes fail in the case of the causal mechanism M1 and M4.  However, they
typically correctly identify the causal direction in the case of the mechanism
M2, which has non-additive noise, and where the other methods tend to fail.
Finally, IR-AN performs slightly better than GR-AN, especially in the case of
additive Gaussian noise, where GR-AN is unable to identify the 
causal direction. MAD provides very bad results for some of the mechanism considered, 
\emph{i.e.}, M1 and M4. Surprisingly this is the case even for Laplacian additive noise,
which is the hypothesis made by MAD. This bad behavior is probably a consequence of 
ignoring the entropies of $\mathcal{X}$ and $\mathcal{Y}$ in this method.
NLME and GR-ENT give worse results than GR-AN in some particular cases, \emph{e.g.}, $p1$
and the causal mechanism M3. We believe this is related to the difficulty of estimating
differential entropies in practice. $\text{GR-AN}^\star$ performs very similar
to $\text{GR-AN}$. This indicates that in practice one may ignore the transformation that
guarantees that $\mathcal{X}$ and $\mathcal{Y}$ are equally distributed.
GR-K4 also gives similar results to GR-AN, probably because in 
these experiments the tails of the residuals are not very heavy. 

\begin{figure}[tb]
\begin{center}
\includegraphics[width=0.99\textwidth]{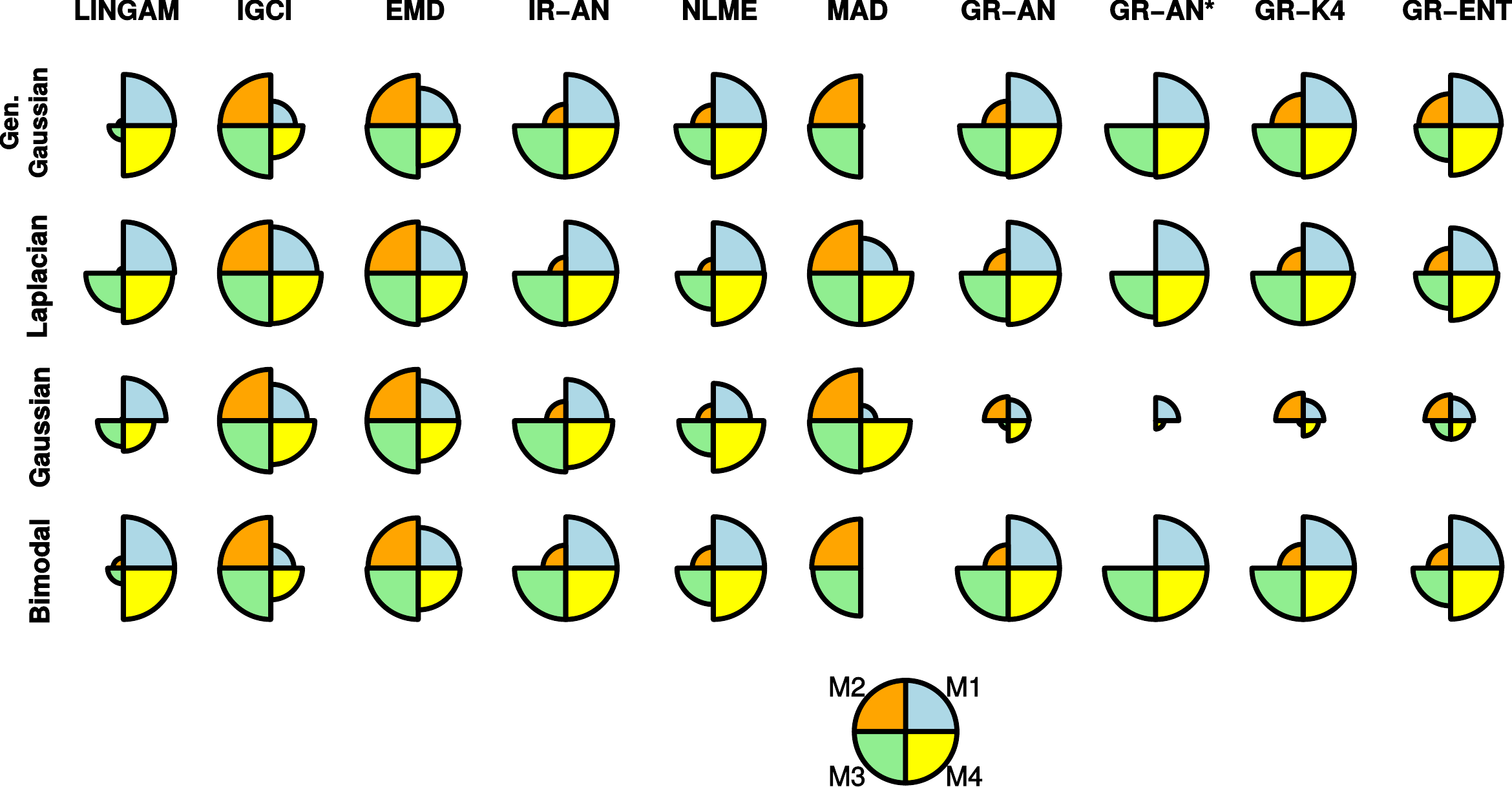}
\end{center}
\caption{Radar charts showing the average accuracy of each method for the different 
types of noise considered and for each mechanism M1, M2, M3 and M4. For a particular 
method and type of noise, the radius of each portion of the pie is proportional to the 
corresponding average accuracy of the method across the distributions $p_1$, $p_2$ and 
$p_3$ for the cause. The pie at the bottom corresponds to 100\% accuracy for each mechanism.
}
\label{fig:radar_chart}
\end{figure}

An overall comparison of the different methods evaluated is shown in Figure \ref{fig:radar_chart}.
This figure displays several radar charts that indicate the average accuracy of each method for the 
different types of noise considered and for each mechanism M1, M2, M3 and M4. In particular, for a 
given method and a given type of noise, the radius of each portion of the pie is proportional to the 
corresponding average accuracy of the method across the distributions $p_1$, $p_2$ and $p_3$ for the cause. 
The pie at the bottom corresponds to 100\% accuracy for each causal mechanism. The conclusions
derived from this figure are similar to the ones obtained from Table \ref{tab1}. 
In particular, IR-AN performs very well, except for multiplicative 
noise (M3), closely followed by GR-AN, $\text{GR-AN}^\star$, and GR-K4, which give similar results. The methods 
perform very poorly in the case of additive Gaussian noise, since they cannot infer the actual causal direction in that 
situation.  NLME and GR-ENT have problems in the case of the causal mechanism M3 and MAD in the case of the mechanisms 
M1 and M4.  LINGAM also performs bad in the case of M3 and IGCI and EMD have problems in the case of the mechanisms M1 and M4.
IGCI, EMD and MAD are the only methods performing well in the case of M2, the mechanism with non-additive noise. 

We have repeated these experiments for other samples sizes, \emph{e.g.}, 100, 200 300 and 1000. The results obtained 
are very similar to the ones reported here, except when the number of samples is small and equal to 100. 
In that case NLME performs slightly better than the proposed approach GR-AN, probably because with
100 samples it is very difficult to accurately estimate the non-linear transformation that is required 
to guarantee that $\mathcal{X}$ and $\mathcal{Y}$ are equally distributed.  These results of these 
additional experiments are found in the supplementary material. 

In summary, the good results provided by GR-AN and its variants in the experiments described indicate 
that (i) when the assumptions made by GR-AN are valid, the method has a good performance and (ii) there 
is indeed a Gaussianization effect in the residuals when the model is fitted under the anti-causal direction. 
Because $\text{GR-AN}^\star$ also performs well in these experiments, this indicates that a Gaussianization of 
the residuals may happen even when $\mathcal{X}$ and $\mathcal{Y}$ do not follow the same distribution.

\begin{sidewaystable}
{\footnotesize
\begin{center}
\caption{Accuracy on the synthetic data for each method, causal mechanisms and type of noise.}
\label{tab1}
\begin{tabular}{l|c@{\hspace{.5mm}}c@{\hspace{.5mm}}c@{\hspace{.5mm}}c|c@{\hspace{.5mm}}c@{\hspace{.5mm}}c@{\hspace{.5mm}}c|c@{\hspace{.5mm}}c@{\hspace{.5mm}}c@{\hspace{.5mm}}c|c@{\hspace{.5mm}}c@{\hspace{.5mm}}c@{\hspace{.5mm}}c}
\hline
{\bf Noise} & \multicolumn{4}{c|}{{\bf Generalized Gaussian}} & \multicolumn{4}{c|}{{\bf Laplacian }} & 
\multicolumn{4}{c|}{{\bf Gaussian }} & \multicolumn{4}{c}{{\bf Bimodal }} \\
\hline
\multirow{2}{2cm}{{\bf Algorithm}} & \multicolumn{4}{c|}{{\bf Mechanism}} & 
\multicolumn{4}{c|}{{\bf Mechanism}} & \multicolumn{4}{c|}{{\bf Mechanism}} & \multicolumn{4}{c}{{\bf Mechanism}} \\
& \multicolumn{1}{c}{{\bf M1}} & \multicolumn{1}{c}{{\bf M2}} & \multicolumn{1}{c}{{\bf M3}} & \multicolumn{1}{c|}{{\bf M4}} & \multicolumn{1}{c}{{\bf M1}} & \multicolumn{1}{c}{{\bf M2}} & \multicolumn{1}{c}{{\bf M3}} & \multicolumn{1}{c|}{{\bf M4}} & \multicolumn{1}{c}{{\bf M1}} & \multicolumn{1}{c}{{\bf M2}} & \multicolumn{1}{c}{{\bf M3}} & \multicolumn{1}{c|}{{\bf M4}} & \multicolumn{1}{c}{{\bf M1}} & \multicolumn{1}{c}{{\bf M2}} & \multicolumn{1}{c}{{\bf M3}} & \multicolumn{1}{c}{{\bf M4}} \\
\hline
$p_1(x)$ & \multicolumn{4}{c|}{} & \multicolumn{4}{c|}{} & \multicolumn{4}{c|}{} &\multicolumn{4}{c}{}  \\
\hspace{0.3cm} LINGAM &                   100\% & 1\% & 7\% & 93\% &      100\% & 0\% & 26\% & 89\% &      58\% & 0\% & 11\% & 12\% &      100\% & 0\% & 3\% & 100\%       \\
\hspace{0.3cm} IGCI   &                   28\% & 100\% & 100\% & 34\% &      73\% & 100\% & 100\% & 96\% &      49\% & 100\% & 100\% & 63\% &      29\% & 98\% & 100\% & 44\%       \\
\hspace{0.3cm} EMD    &                   34\% & 96\% & 100\% & 43\% &      70\% & 100\% & 100\% & 77\% &      48\% & 100\% & 100\% & 50\% &      50\% & 99\% & 100\% & 55\%       \\
\hspace{0.3cm} IR-AN  &                   100\% & 31\% & 100\% & 99\% &      99\% & 26\% & 100\% & 97\% &      45\% & 34\% & 100\% & 75\% &      100\% & 46\% & 100\% & 100\%       \\
\hspace{0.3cm} NLME   &                   100\% & 31\% & 22\% & 100\% &      100\% & 20\% & 15\% & 100\% &      45\% & 22\% & 8\% & 96\% &      100\% & 42\% & 17\% & 100\%       \\
\hspace{0.3cm} MAD    &                   0\% & 94\% & 100\% & 0\% &     100\% & 99\% & 100\% & 100\% &     50\% & 99\% & 100\% & 97\% &     0\% & 92\% & 97\% & 0\%      \\
\hspace{0.3cm} GR-AN  &                   100\% & 46\% & 95\% & 100\% &     100\% & 44\% & 80\% & 100\% &     48\% & 47\% & 6\% & 20\% &     100\% & 51\% & 100\% & 100\%      \\
\hspace{0.3cm} $\text{GR-AN}^\star$ &     100\% & 0\% & 94\% & 100\% &     100\% & 0\% & 94\% & 100\% &     45\% & 0\% & 1\% & 17\% &     100\% & 0\% & 100\% & 100\%      \\
\hspace{0.3cm} GR-K4 &                    100\% & 70\% & 94\% & 100\% &     99\% & 51\% & 96\% & 100\% &     52\% & 56\% & 2\% & 19\% &     100\% & 53\% & 100\% & 100\%      \\
\hspace{0.3cm} GR-ENT &                   100\% & 71\% & 76\% & 97\% &     93\% & 52\% & 51\% & 93\% &     41\% & 58\% & 26\% & 29\% &     100\% & 50\% & 84\% & 99\%      \\
\hline
$p_2(x)$ & \multicolumn{4}{c|}{} & \multicolumn{4}{c|}{} & \multicolumn{4}{c|}{} & \multicolumn{4}{c}{} \\
\hspace{0.3cm} LINGAM &                  100\% & 32\% & 68\% & 100\% &      100\% & 30\% & 99\% & 100\% &      100\% & 13\% & 91\% & 100\% &      100\% & 53\% & 91\% & 100\%       \\
\hspace{0.3cm} IGCI   &                  40\% & 97\% & 100\% & 71\% &      98\% & 100\% & 100\% & 99\% &      72\% & 100\% & 100\% & 97\% &      39\% & 99\% & 100\% & 54\%       \\
\hspace{0.3cm} EMD    &                  96\% & 99\% & 100\% & 98\% &      96\% & 100\% & 100\% & 98\% &      94\% & 100\% & 100\% & 90\% &      92\% & 95\% & 100\% & 95\%       \\
\hspace{0.3cm} IR-AN  &                  100\% & 55\% & 100\% & 100\% &      100\% & 42\% & 100\% & 100\% &      100\% & 44\% & 100\% & 100\% &      100\% & 43\% & 100\% & 100\%       \\
\hspace{0.3cm} NLME   &                  100\% & 47\% & 100\% & 100\% &      95\% & 36\% & 100\% & 100\% &      99\% & 36\% & 100\% & 100\% &      100\% & 38\% & 100\% & 100\%       \\
\hspace{0.3cm} MAD    &                  0\% & 100\% & 100\% & 13\% &     5\% & 100\% & 100\% & 100\% &     2\% & 100\% & 100\% & 95\% &     0\% & 100\% & 85\% & 0\%      \\
\hspace{0.3cm} GR-AN  &                  100\% & 46\% & 100\% & 100\% &     98\% & 32\% & 96\% & 100\% &     29\% & 40\% & 16\% & 54\% &     100\% & 32\% & 100\% & 100\%      \\
\hspace{0.3cm} $\text{GR-AN}^\star$ &    100\% & 0\% & 94\% & 100\% &     100\% & 0\% & 65\% & 100\% &     39\% & 0\% & 0\% & 6\% &     100\% & 0\% & 100\% & 100\%      \\
\hspace{0.3cm} GR-K4 &                   100\% & 50\% & 100\% & 100\% &     87\% & 44\% & 100\% & 98\% &     26\% & 51\% & 23\% & 47\% &     100\% & 30\% & 100\% & 100\%      \\
\hspace{0.3cm} GR-ENT &                  96\% & 56\% & 90\% & 98\% &     78\% & 40\% & 73\% & 92\% &     46\% & 45\% & 42\% & 48\% &     100\% & 36\% & 97\% & 100\%      \\
\hline
$p_3(x)$ & \multicolumn{4}{c|}{} & \multicolumn{4}{c|}{} & \multicolumn{4}{c|}{} & \multicolumn{4}{c}{} \\
\hspace{0.3cm} LINGAM &                   100\% & 0\% & 19\% & 100\% &      100\% & 0\% & 98\% & 100\% &      92\% & 0\% & 56\% & 67\% &      100\% & 8\% & 7\% & 100\%       \\
\hspace{0.3cm} IGCI   &                   76\% & 100\% & 100\% & 83\% &      100\% & 100\% & 100\% & 100\% &      92\% & 100\% & 100\% & 97\% &      67\% & 100\% & 100\% & 87\%       \\
\hspace{0.3cm} EMD    &                   90\% & 100\% & 100\% & 94\% &      98\% & 100\% & 100\% & 100\% &      92\% & 100\% & 100\% & 97\% &      96\% & 100\% & 100\% & 94\%       \\
\hspace{0.3cm} IR-AN  &                   100\% & 40\% & 100\% & 100\% &      100\% & 26\% & 100\% & 100\% &      96\% & 35\% & 100\% & 100\% &      100\% & 45\% & 100\% & 100\%       \\
\hspace{0.3cm} NLME   &                   100\% & 44\% & 100\% & 100\% &      100\% & 26\% & 100\% & 100\% &      74\% & 34\% & 98\% & 100\% &      100\% & 43\% & 99\% & 100\%       \\
\hspace{0.3cm} MAD    &                   0\% & 97\% & 100\% & 1\% &     100\% & 100\% & 100\% & 100\% &     38\% & 98\% & 100\% & 98\% &     0\% & 95\% & 99\% & 0\%      \\
\hspace{0.3cm} GR-AN  &                   100\% & 52\% & 90\% & 100\% &     100\% & 58\% & 100\% & 100\% &     46\% & 53\% & 36\% & 45\% &     100\% & 54\% & 98\% & 100\%      \\
\hspace{0.3cm} $\text{GR-AN}^\star$ &     100\% & 0\% & 100\% & 100\% &     100\% & 0\% & 100\% & 100\% &     51\% & 0\% & 12\% & 26\% &     100\% & 0\% & 100\% & 100\%      \\
\hspace{0.3cm} GR-K4 &                    100\% & 64\% & 94\% & 100\% &     99\% & 47\% & 100\% & 97\% &     43\% & 54\% & 15\% & 24\% &     100\% & 57\% & 100\% & 100\%      \\
\hspace{0.3cm} GR-ENT &                   100\% & 60\% & 42\% & 95\% &     94\% & 54\% & 86\% & 91\% &     47\% & 48\% & 48\% & 32\% &     100\% & 47\% & 43\% & 99\%      \\
\hline
\end{tabular}
\end{center}
}
\end{sidewaystable}

\begin{figure}[tb]
\begin{center}
\begin{tabular}{c|c@{\hspace{1mm}}c|c@{\hspace{1mm}}c}
& \multicolumn{2}{c|}{{\small \bf Causal Direction ($\tilde{\mathcal{X}} \rightarrow \mathcal{Y})$}} & 
\multicolumn{2}{c}{{\small \bf Anti-causal Direction ($\mathcal{Y} \rightarrow \tilde{\mathcal{X}})$}} \\
\hline
\rotatebox{90}{\hspace{.5cm}{\small Gen. Gaussian}} &
\includegraphics[width=0.225\textwidth]{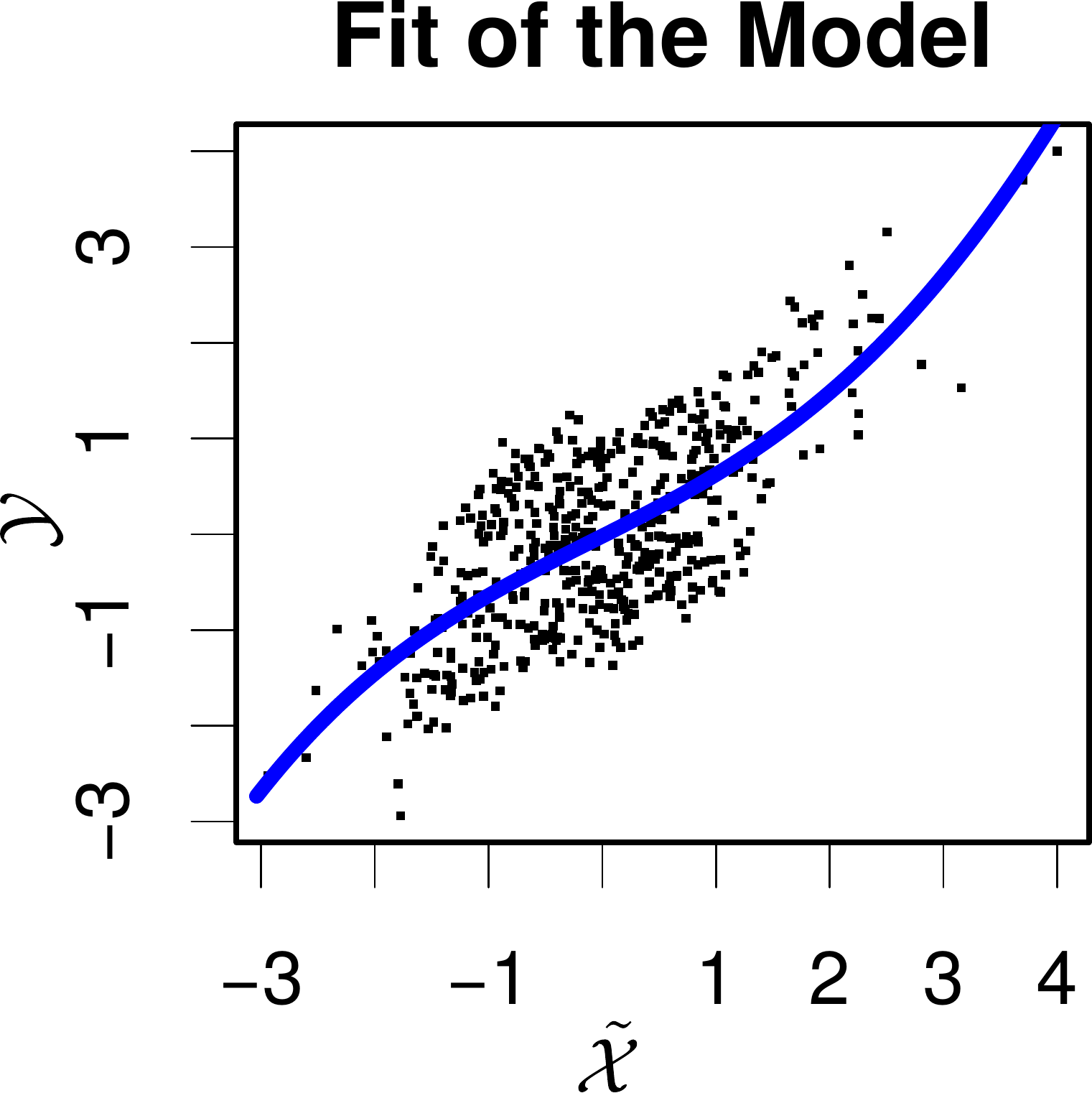} &
\includegraphics[width=0.225\textwidth]{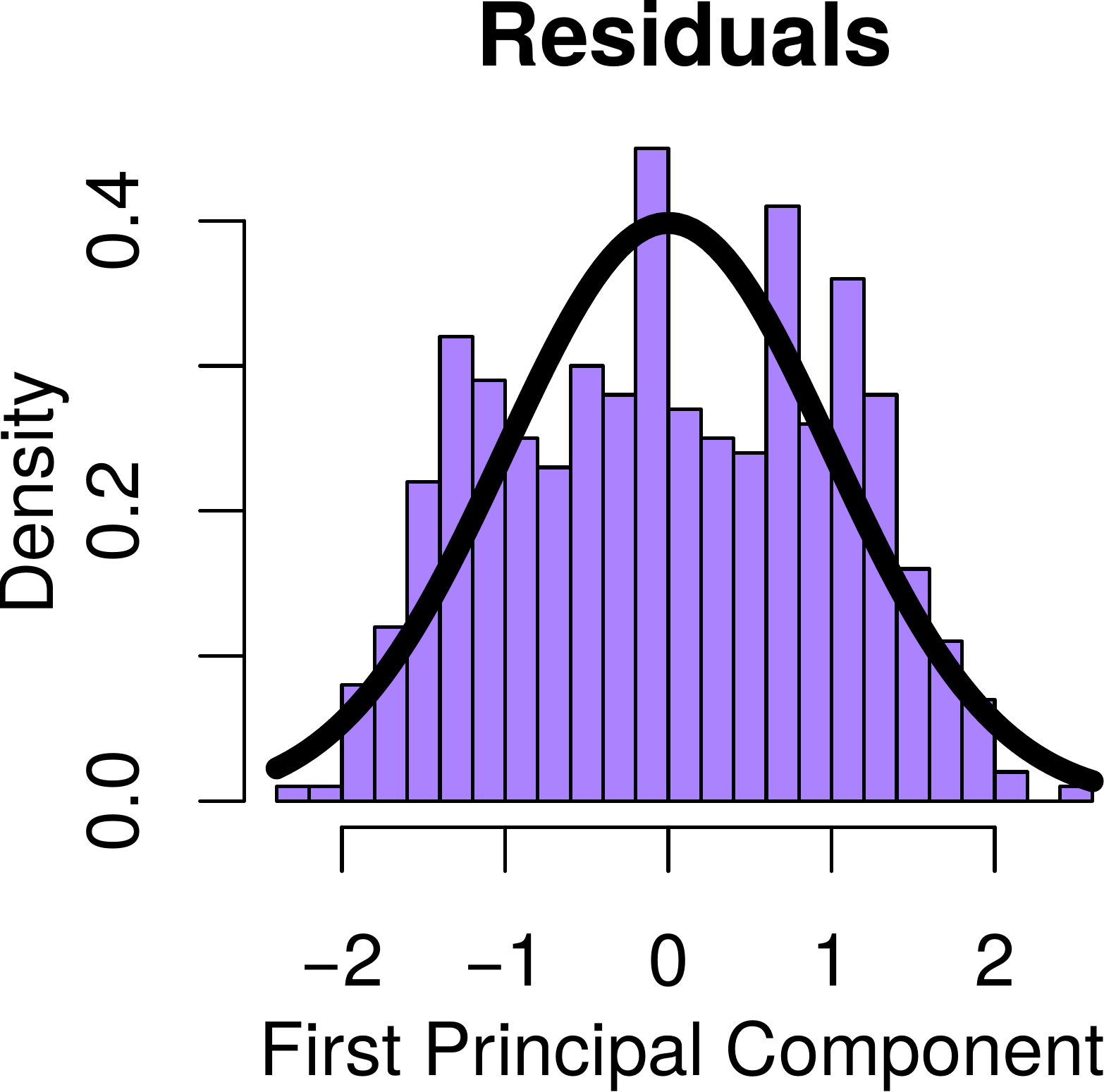} &
\includegraphics[width=0.225\textwidth]{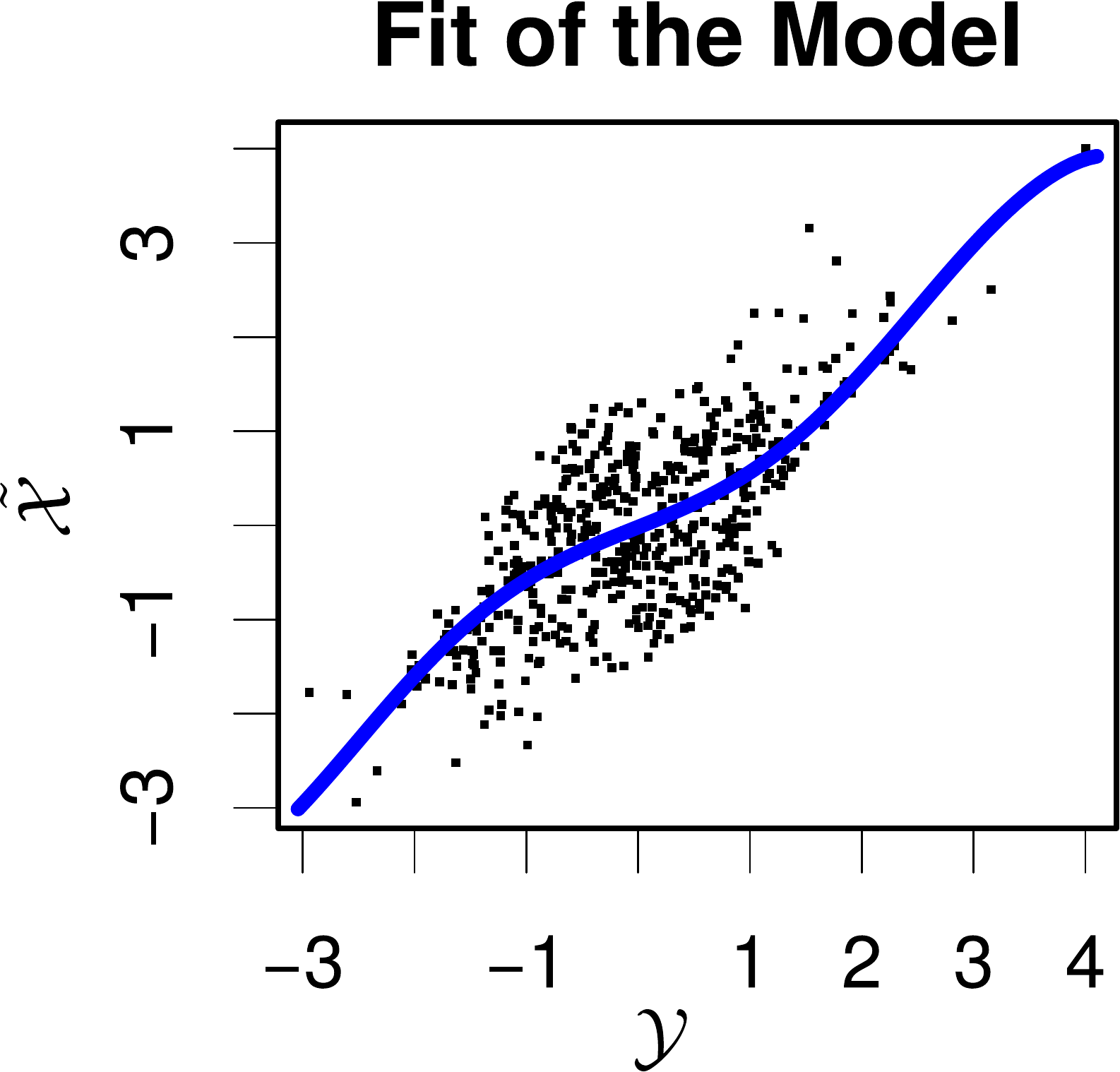} &
\includegraphics[width=0.225\textwidth]{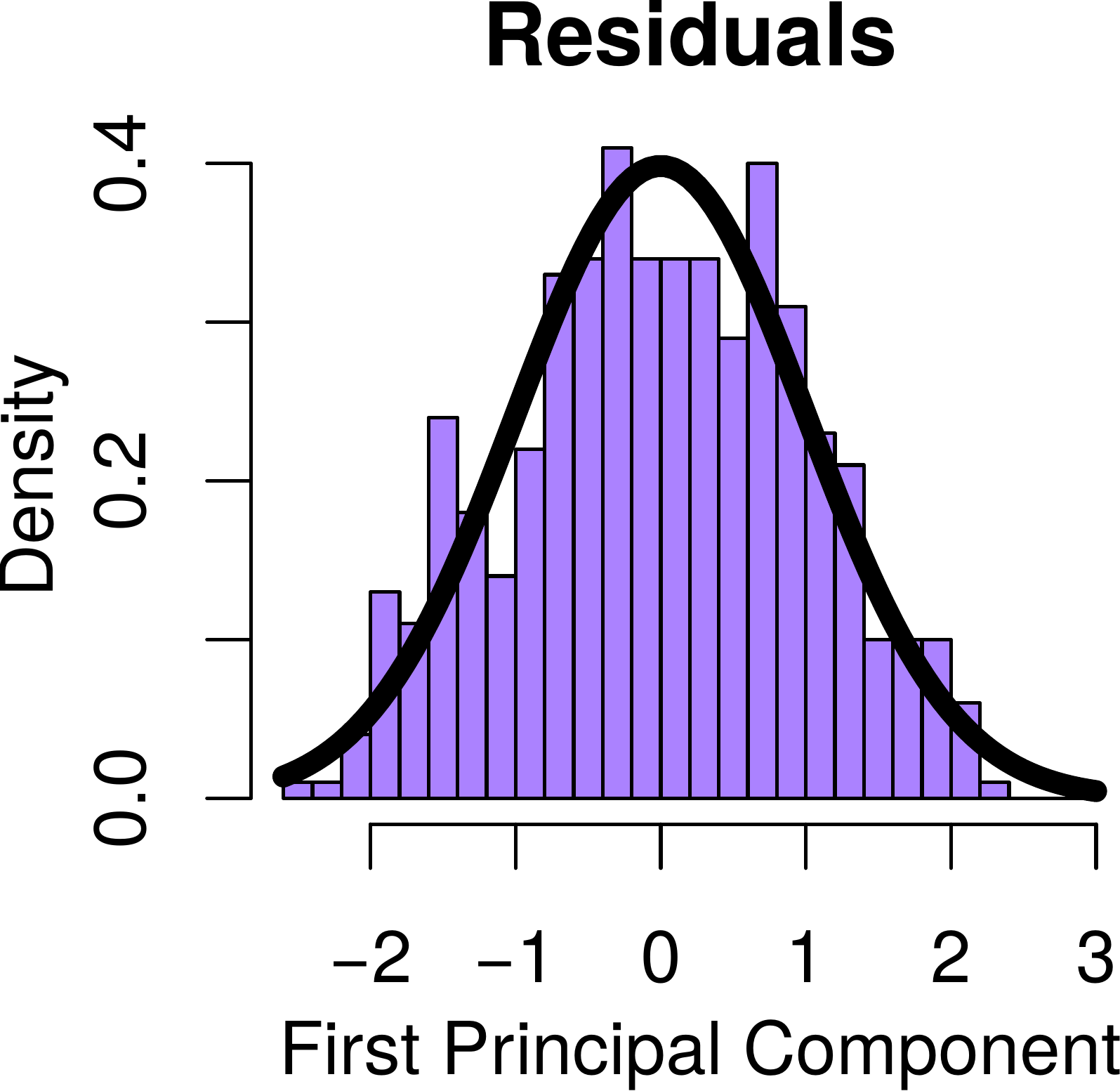} \\
\hline
\rotatebox{90}{\hspace{.75cm}{\small Laplacian}} &
\includegraphics[width=0.225\textwidth]{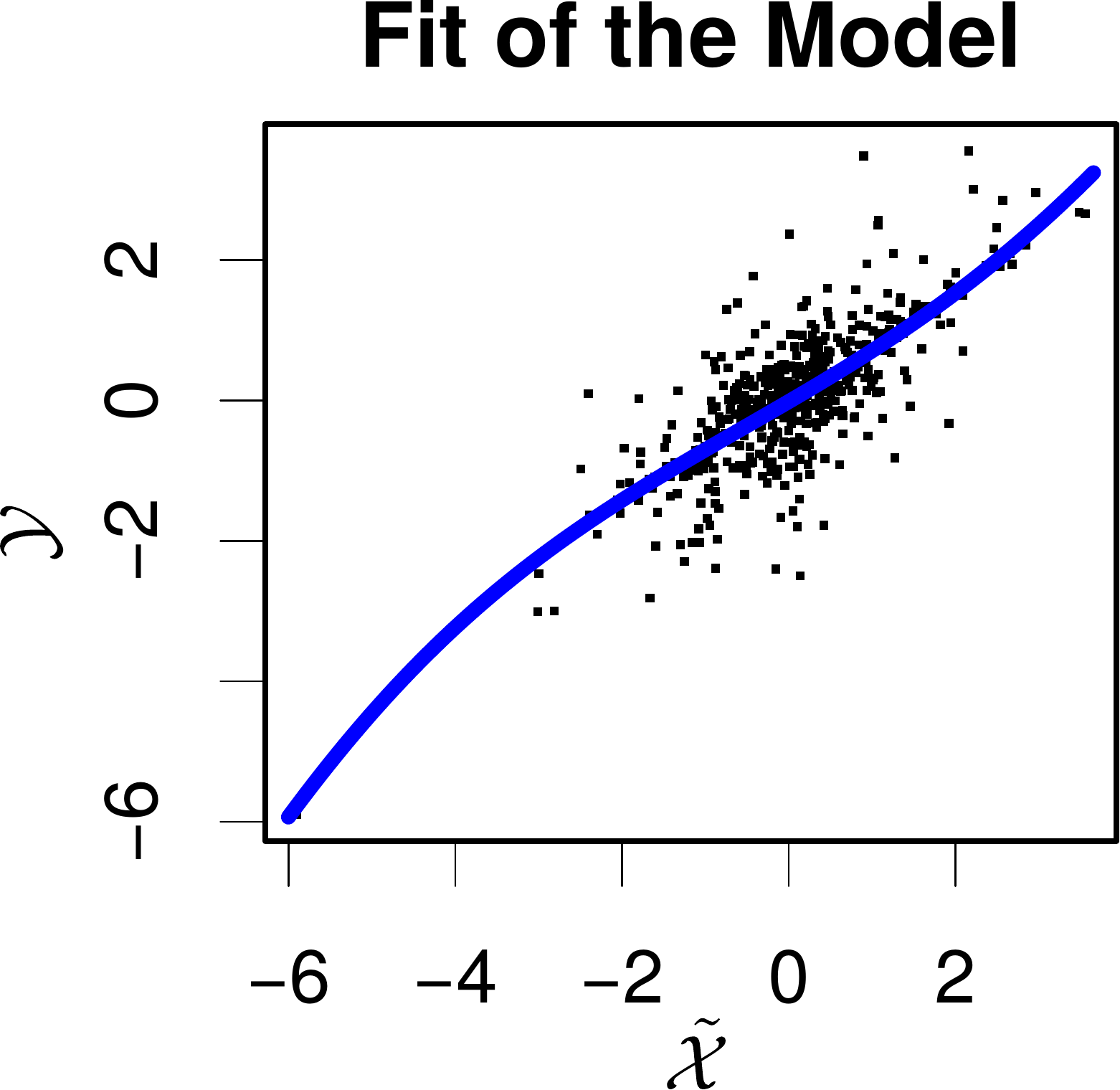} &
\includegraphics[width=0.225\textwidth]{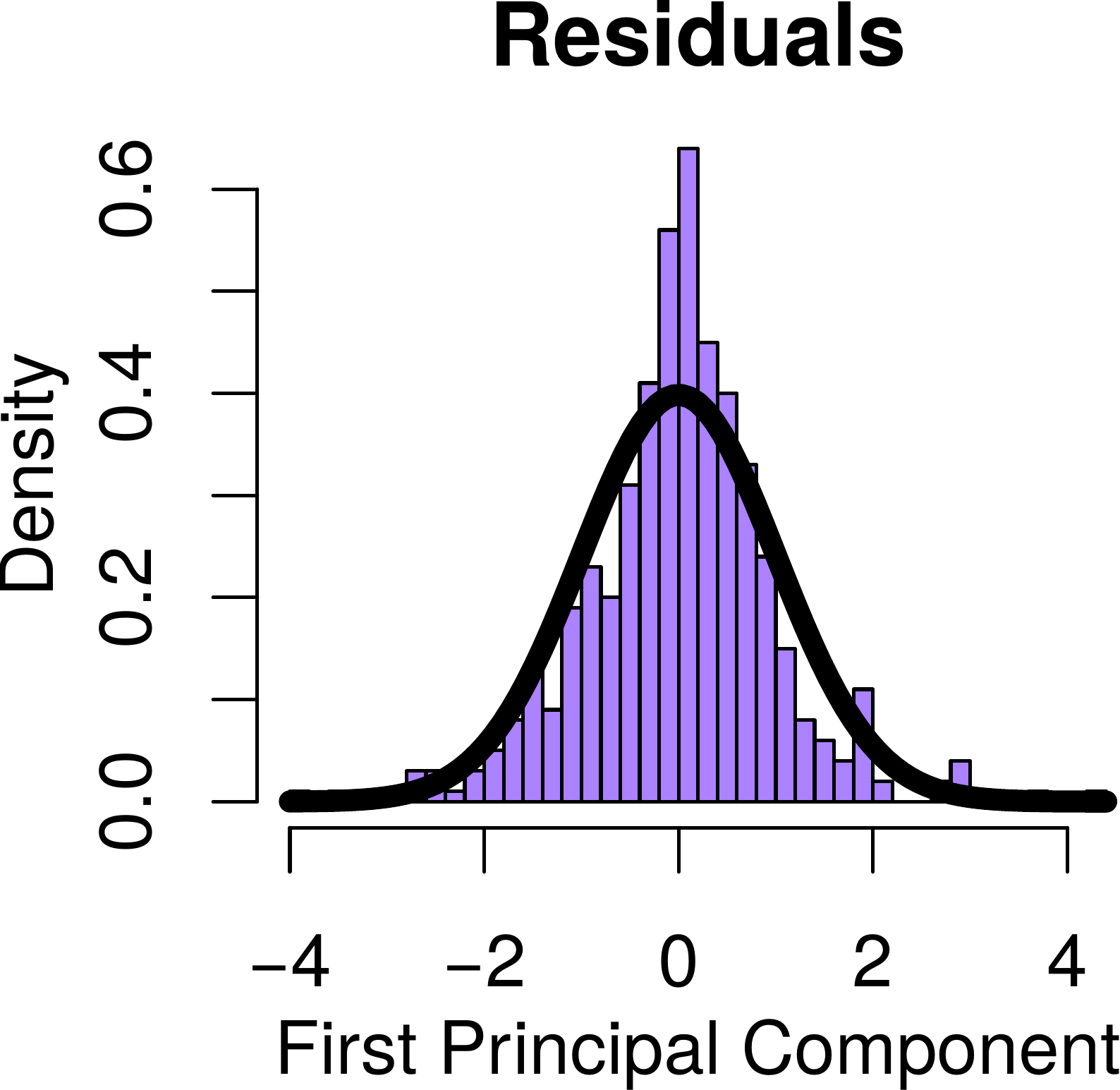} &
\includegraphics[width=0.225\textwidth]{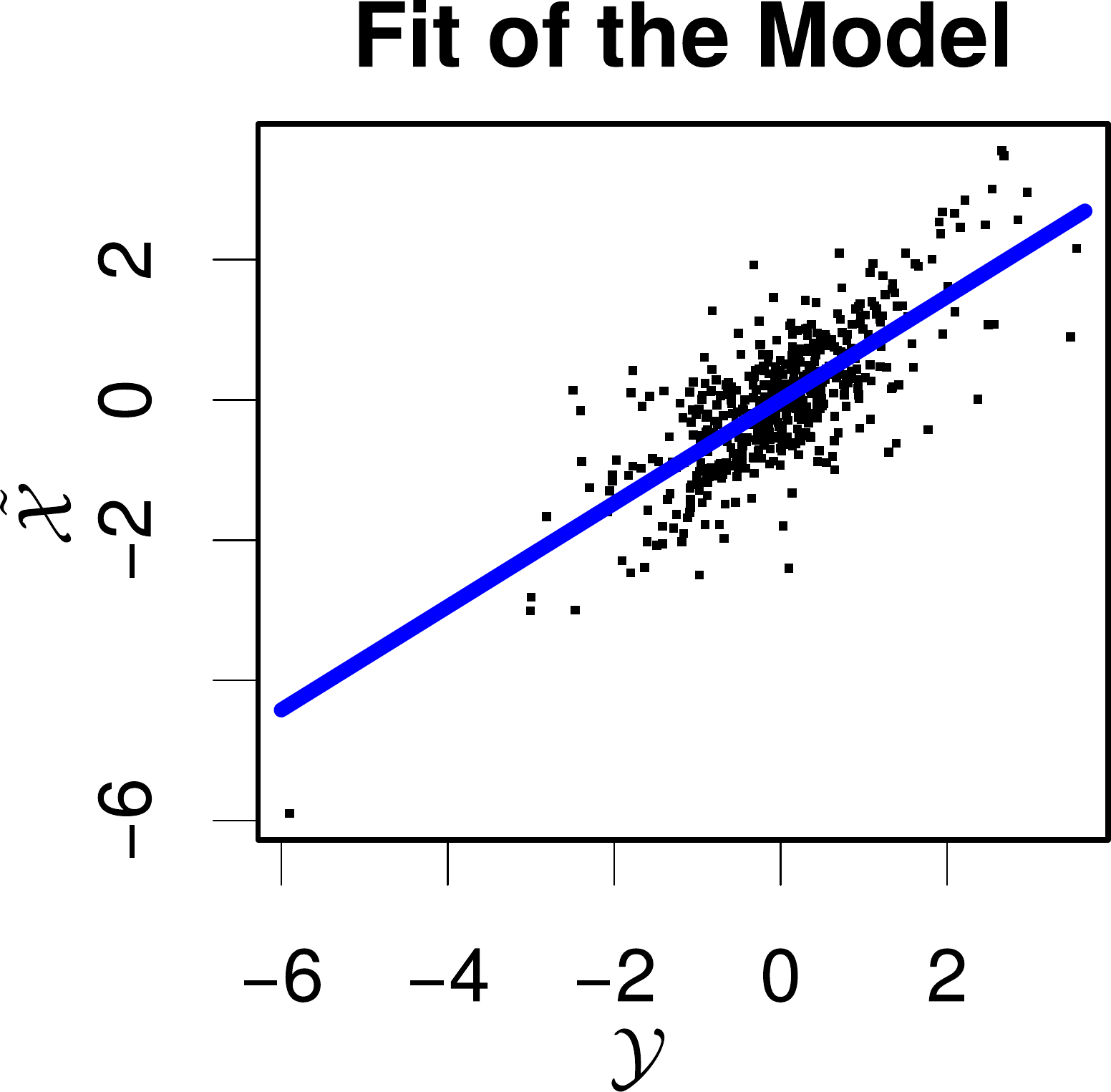} &
\includegraphics[width=0.225\textwidth]{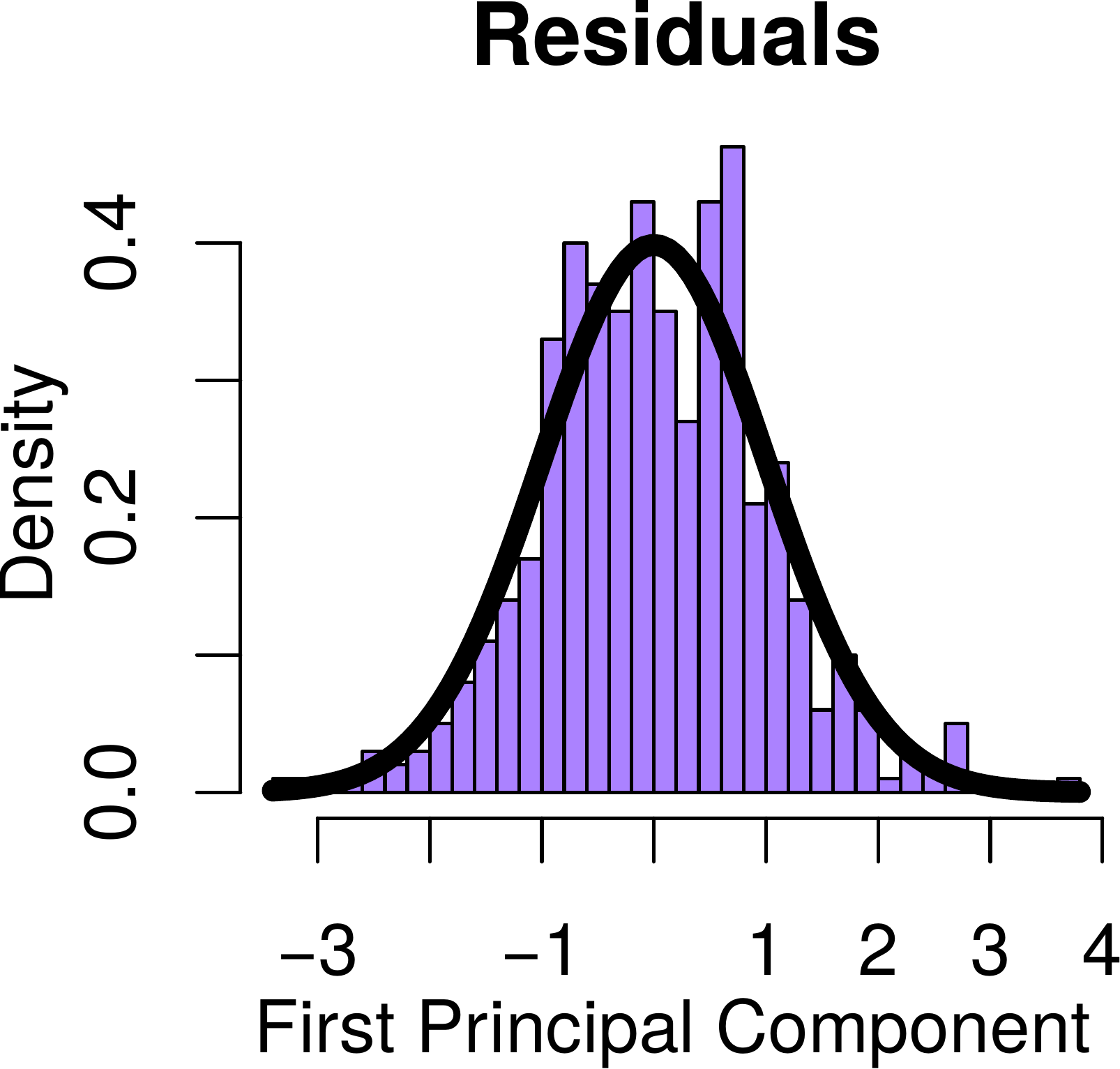} \\
\hline
\rotatebox{90}{\hspace{.8cm}{\small Bimodal}} &
\includegraphics[width=0.225\textwidth]{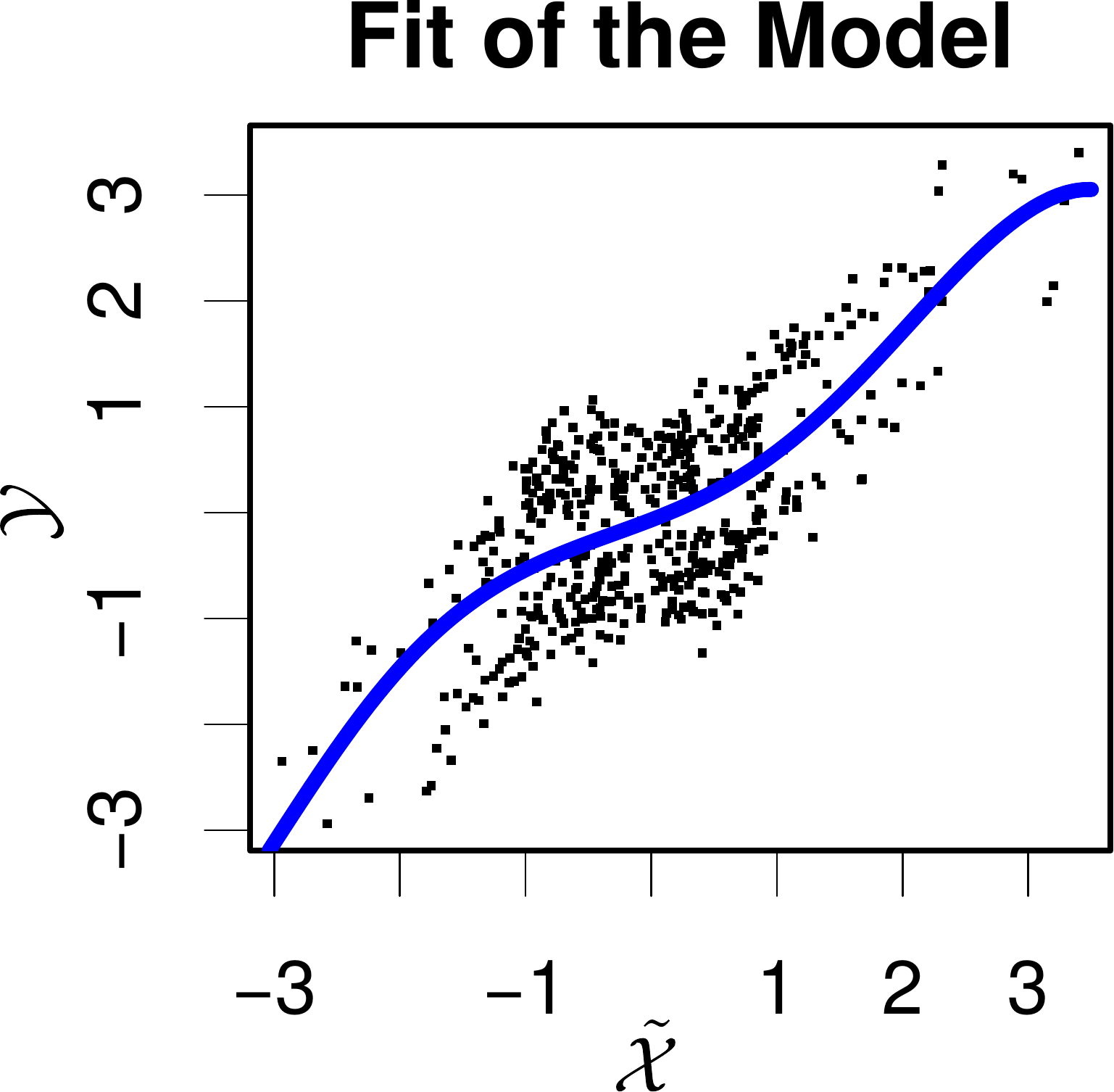} &
\includegraphics[width=0.225\textwidth]{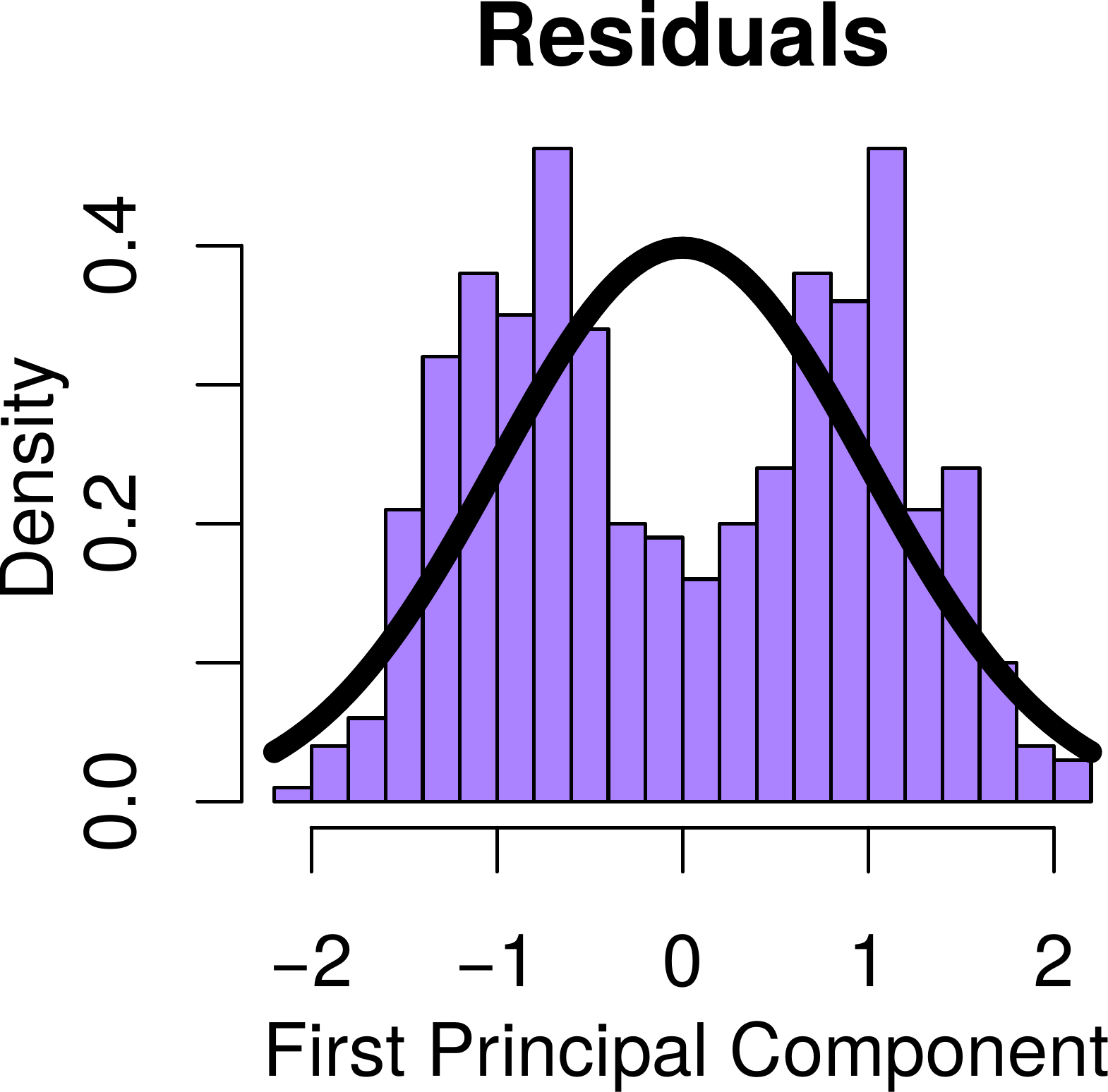} &
\includegraphics[width=0.225\textwidth]{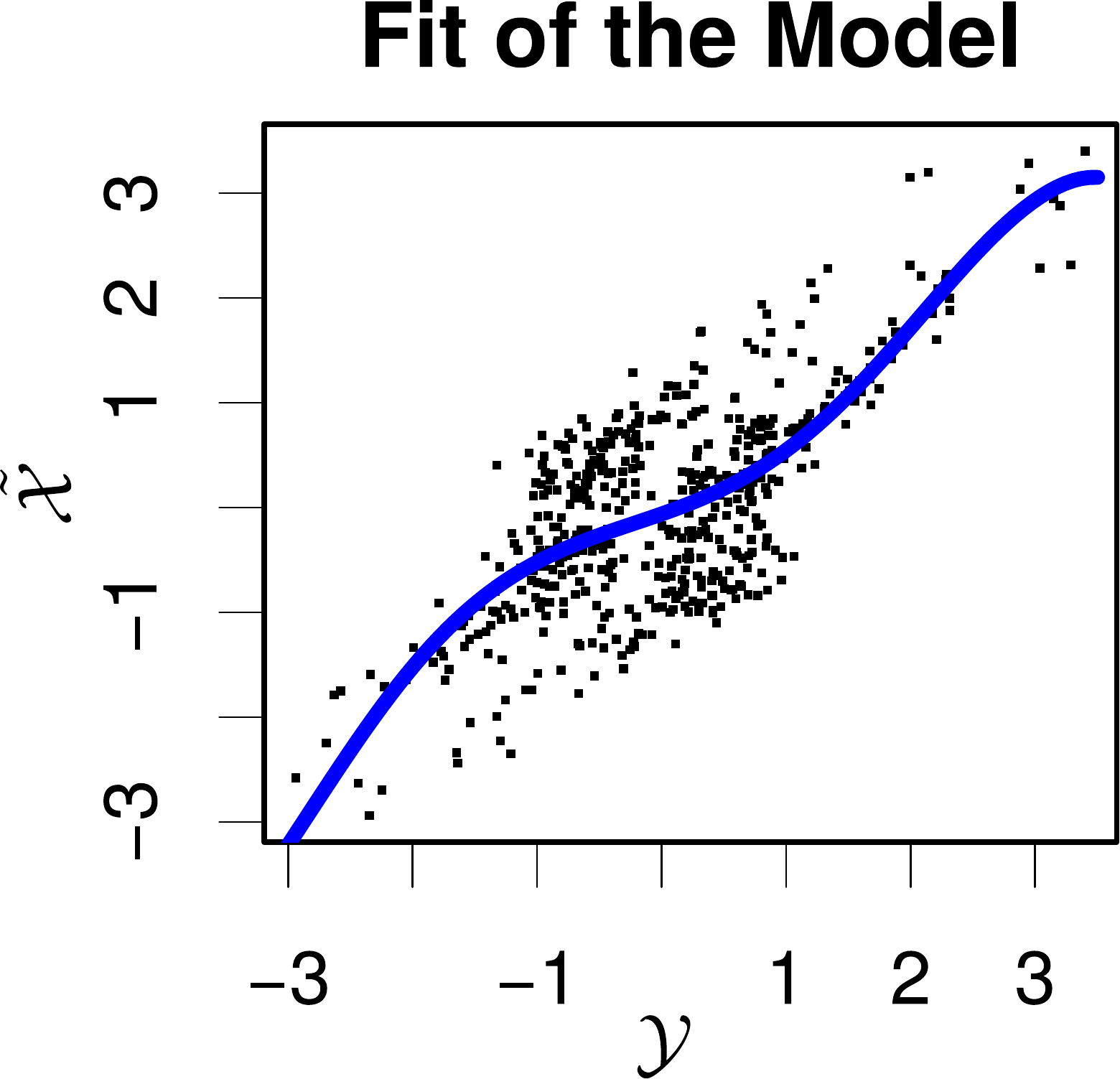} &
\includegraphics[width=0.225\textwidth]{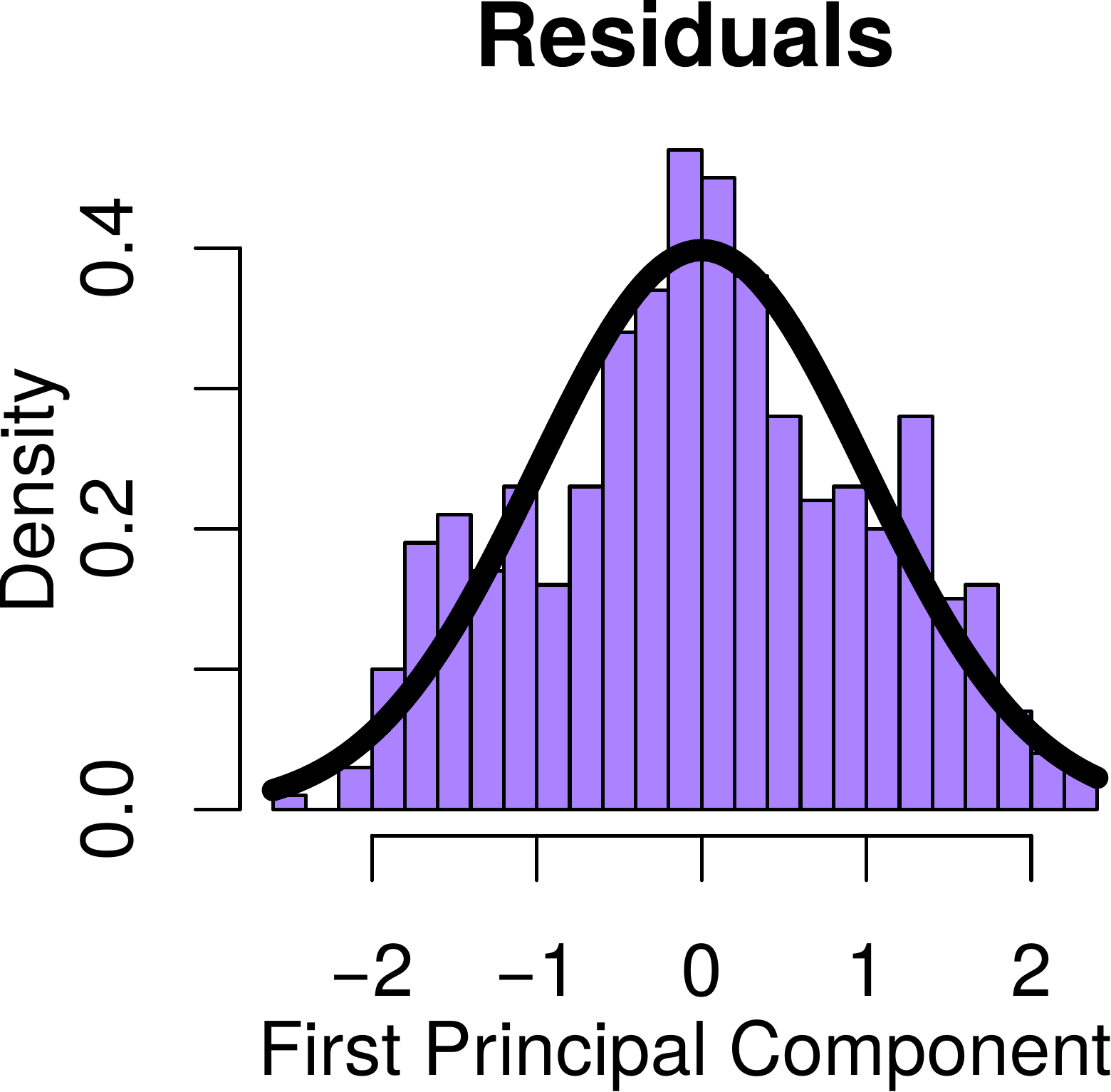} \\
\hline
\end{tabular}
\end{center}
\caption{ (left column) Predicted pre-images obtained in the casual direction
$\tilde{\mathcal{X}}\rightarrow \mathcal{Y}$ alongside with a histogram of the
first principal component of the residuals in feature space. A Gaussian fit is
displayed as a solid black line.  Results are shown for each type of additive noise 
considered. (right column) Same plots for the anti-causal 
direction $\mathcal{Y} \rightarrow \tilde{\mathcal{X}}$.}
\label{fig:sample_synthetic}
\end{figure}

We give further evidence of the Gaussianization of the distribution of the residuals 
obtained when fitting the model under the anti-causal direction. For this, we analyze in
detail three particular cases of GR-AN corresponding to the causal mechanism M3, the 
distribution $p_2(x)$ for the cause and each of the three types of additive noise 
considered. Namely, generalized Gaussian noise, Laplacian noise and bimodal noise. 
Figure \ref{fig:sample_synthetic} shows the predicted pre-images for new data instances 
when the model has been fitted in the causal ($\tilde{\mathcal{X}} \rightarrow \mathcal{Y}$) 
and the anti-causal ($\mathcal{Y} \rightarrow \tilde{\mathcal{X}}$) 
direction alongside with a histogram of the first
principal component of the residuals in feature space. A Gaussian approximation
is also displayed as a solid black line on top of the histogram. In this case 
$\mathbf{x}$, \emph{i.e.}, the samples of $\mathcal{X}$, have been transformed 
to be equally distributed to $\mathbf{y}$, \emph{i.e.}, the samples from $\mathcal{Y}$.
We observe that the distribution of the residuals in the anti-causal
direction ($\mathcal{Y} \rightarrow \tilde{\mathcal{X}}$) is more
similar to a Gaussian distribution. Furthermore, for the direction
$\tilde{\mathcal{X}}\rightarrow \mathcal{Y}$ the statistic of the energy based
Gaussianity test for the first principal component of the residuals is respectively 
$4.02$, $4.64$ and $11.46$, for generalized Gaussian, Laplacian and bimodal noise. 
Recall that the larger the value the larger the deviation from Gaussianity. In the 
case of the direction $\mathcal{Y} \rightarrow \tilde{\mathcal{X}}$, the energy statistic 
associated to the residuals is $0.97$, $0.68$ and $1.31$, respectively. 
When $\mathbf{y}$ is transformed to have the same distribution as
$\mathbf{x}$ similar results are observed (results not shown). However, the
Gaussianization effect is not as strong as in this case, probably because it
leads to the violation of the additive noise assumption. In summary, the figure
displayed illustrates in detail the Gaussianization effect of the residuals
when fitting the model in the anti-causal direction.

\subsection{Experiments with real cause-effect pairs}

A second batch of experiments is performed on the cause-effect pairs from the
ChaLearn challenge\footnote{https://www.codalab.org/competitions/1381}.  This
challenge contains 8073 cause-effect data pairs with a labeled causal
direction. From these pairs, we consider a subset for our experiments. In
particular, we select the 184 pairs that have (i) at least $500$ samples, and
(ii) a fraction of repeated instances for each random variable of at most $1\%$.
The first criterion guarantees that there is enough data to make a decision with high 
confidence.   The second criterion removes the pairs with discrete random variables,
motivated by the transformation required by the GR-AN method to guarantee the
equal distribution of $\mathcal{X}$ and $\mathcal{Y}$. In particular, this 
transformation cannot be carried out on discrete data. Another advantage is that 
this filtering process of the data facilitates the experiments since several of the 
methods considered in the comparison (\emph{i.e.}, GR-AN, GR-ENT, GR-K4, IR-AN, NLME, MAD  and EMD) are computationally 
very expensive. More precisely, they have a cubic cost with respect to the number of 
samples and they require tuning several hyper-parameters. The consequence is that evaluating 
these methods on the 8073 pairs available is therefore not feasible.

\begin{figure}[tb]
\begin{center}
\includegraphics[width=0.99\textwidth]{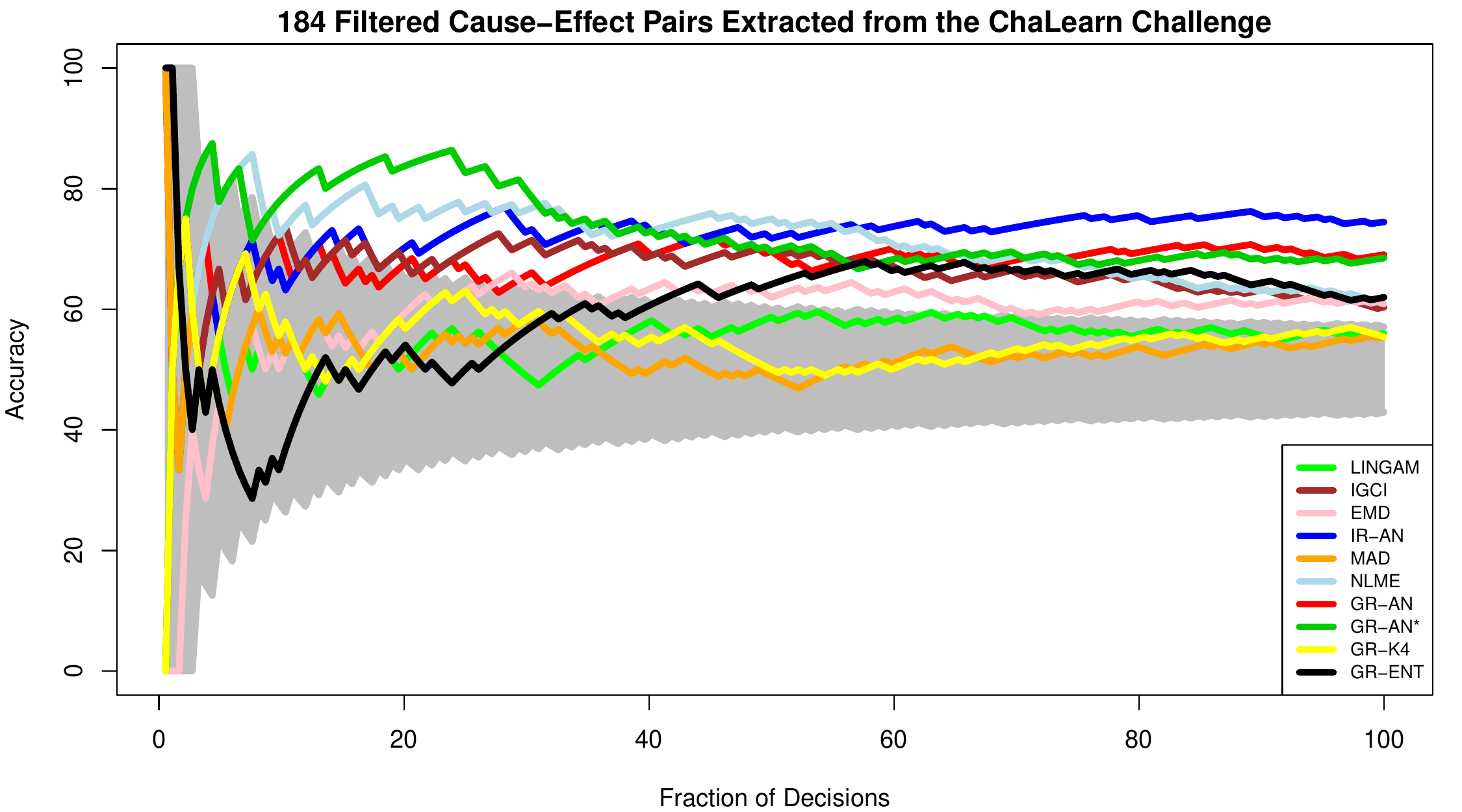} 
\end{center}
\caption{Accuracy of each method, as a fraction of the decisions made, on the
184 filtered cause-effect pairs extracted from the ChaLearn challenge. The number
of samples of each pair is equal to 500. Best seen in color.}
\label{fig:pairs_challenge}
\end{figure}

Using these 184 pairs we evaluate each of the methods considered in the
previous section and report the corresponding accuracy as a function of the
decisions made. In these experiments we sample at random $500$ instances from
each cause-effect pair.  This is a standard number of samples that has been 
previously employed by other authors in their experiments with cause-effect 
pairs \citep{Janzing12}. Furthermore, a threshold value is fixed and the
obtained confidence in the decision by each method is compared to such
threshold. Only if the confidence is above the threshold value, the
cause-effect pair is considered in the evaluation of the accuracy of the
corresponding method. A summary of the results is displayed in Figure
\ref{fig:pairs_challenge}.  This figure shows for each method, as a fraction of
the decisions made, the accuracy on the filtered datasets on which the
confidence on the decision is above the threshold value. A gray area has been
drawn to indicate accuracy values that are not statistically different from
random guessing (accuracy equal to 50\%) using a binomial test (p-value above
5\%). We observe that IR-AN obtains the best results, followed by GR-AN, $\text{GR-AN}^\star$, 
GR-K4, GR-ENT, IGCI, NLME and EMD. NLME, IGCI and EMD perform worse than GR-AN and $\text{GR-AN}^\star$ when 
a high number of decisions are made. The differences in performance between IR-AN 
and GR-AN, when 100\% of the decisions are made, are not statistically significant 
(a paired t-test returns a $p$-value equal to $25\%$). 
The fact that the performance of $\text{GR-AN}^\star$ is similar to the performance of 
GR-AN also indicates that there is some Gaussianization of the residuals 
even though the two random variables $\mathcal{X}$ and $\mathcal{Y}$ are not equally distributed.
We observe that the results of LINGAM and MAD are not statistically different from 
random guessing. This remarks the importance of non-linear models and questions 
the practical utility of the MAD method.  In these experiments, GR-ENT and GR-K4 perform 
worse than GR-AN, which remarks the benefits of using the energy distance to estimate
the deviation from the Gaussian distribution, as a practical alternative to entropy or 
cumulant based measures.

In summary, the results displayed in Figure \ref{fig:pairs_challenge} confirm that the level of 
Gaussianity of the residuals, estimated using statistical tests, is a useful metric that can be 
used to identify the causal order of two random variables. Furthermore, this figure also validates 
the theoretical results obtained in Section \ref{sec:asymmetry} which state that one
should expect residuals whose distribution is closer to the Gaussian distribution when 
performing a fit in the anti-causal direction. 

In the supplementary material we include additional results 
for other sample sizes. Namely, $100$, $200$ and $300$ samples. The results obtained are similar to the 
ones reported in Figure \ref{fig:pairs_challenge}. However, the 
differences between GR-ENT, NLME, GR-AN and $\text{GR-AN}^\star$ are smaller.
Furthermore, when the number of samples is small (\emph{i.e.}, equal to $100$)
GR-AN performs worse than NLME, probably because with such a small number of samples it is 
difficult to estimate the non-linear transformation that guarantees that $\mathcal{X}$ and $\mathcal{Y}$
are equality distributed. 

\begin{figure}[tb]
\begin{center}
\includegraphics[width=0.99\textwidth]{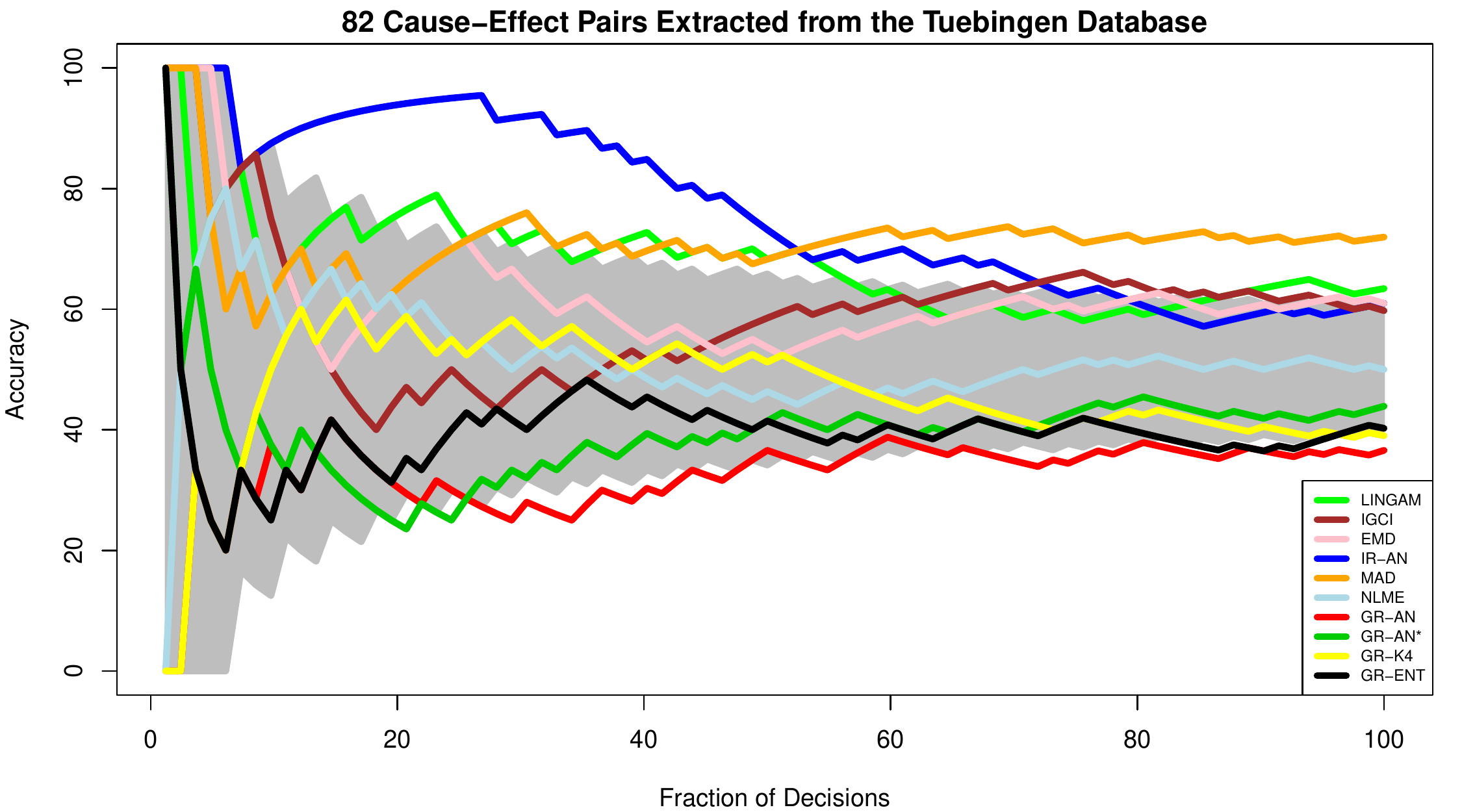} 
\end{center}
\caption{Accuracy of each method, as a fraction of the decisions made, on the
82 cause-effect pairs extracted from the Tuebingen database. Best seen in color.}
\label{fig:pairs_tub}
\end{figure}

In this section we have also evaluated the different methods compared in the previous
experiments on a random subset of 184 cause-effect pairs chosen across the 8073
pairs of the ChaLearn challenge (results not shown). In this case, the 
ranking of the curves obtained looks similar to the ranking displayed in Figure
\ref{fig:pairs_challenge}, \emph{i.e.}, IR-AN performs best followed by GR-AN,
$\text{GR-AN}^\star$, NLME and EMD. However, all methods obtain worse results in general 
and none of them, except IR-AN, perform significantly different from random guessing.

Finally, we also have evaluated the different methods in a subset of 82
cause-effect pairs  extracted from the T\"ubingen cause-effect
pairs\footnote{http://webdav.tuebingen.mpg.de/cause-effect/}. We only
considered those pairs with scalar cause and effect. The results obtained are
displayed in Figure \ref{fig:pairs_tub}. In this case, the performance of the
different methods is worse than the one displayed in Figure
\ref{fig:pairs_challenge}. Only IR-AN, IGCI and MAD perform significantly better
than random guessing. Furthermore, GR-AN and $\text{GR-AN}^\star$ do not perform 
well in this set of cause-effect pairs. This is also the case of NLME.
We believe that the reason for this bad performance is that
in most of these pairs some of the variables take discrete or repeated values.
In the case of $\text{GR-AN}$ this makes infeasible to transform the two 
random variables, $\mathcal{X}$ and $\mathcal{Y}$, so that they are equally 
distributed. Furthermore, the discrete random variables may have a strong
impact in the tests for Gaussianity and in the estimation of the differential
entropy. This could explain the bad performance of $\text{GR-AN}^\star$, NLME and GR-ENT.

In summary, the results reported in this section have shown that in some
cause-effect pairs, when the assumptions made by the proposed method are
satisfied, there is indeed a Gaussianization effect in the residuals obtained
when fitting the model in the anti-causal direction, and this asymmetry is
useful to carry out causal inference on both synthetic and real inference problems.
Our experiments also show that the transformation employed to guarantee that $\mathcal{X}$ and
$\mathcal{Y}$ are equally distributed can be ignored in some cases without 
decreasing the performance. This indicates that our statement about the increased
level of Gaussianity of the residuals, in terms of the increase of the entropy and the
reduction of the high-order cumulants, may be true under more general assumptions.

\section{Conclusions}
\label{sec:conclusions}

In this paper we have shown that in the case of cause-effect pairs with
additive non-Gaussian noise there is an asymmetry that can be used for causal
inference. In particular, assuming that the cause and the effect are 
equally distributed random variables, 
that are linearly related, the residuals of a least 
squares fit in the anti-causal direction are more Gaussian than the 
residuals of a linear fit in the causal direction due a reduction of the magnitude of the high-order cumulants. 
Furthermore, by extending the results of \cite{HyvarinenS13} based on
information theory, we have shown that this Gaussianization 
effect is also present when the two random variables are 
multivariate due to an increment of the differential entropy. This motivates the 
use of kernel methods to work in an expanded feature space. This enables addressing 
non-linear cause-effect inference problems using simple techniques. Specifically, 
kernel methods allow to fit a linear model in an expanded feature space which will be 
non-linear in the original input space. 

Taking advantage of the asymmetry described,  
we have designed a method for non-linear causal 
inference, GR-AN (Gaussianity of the Residuals under Additive Noise). The method consists in 
computing the residuals of a linear model in an expanded feature space in both directions, \emph{i.e.},
$\mathcal{X} \rightarrow \mathcal{Y}$ and $\mathcal{Y}\rightarrow\mathcal{X}$. The 
expected causal direction is the one in which the residuals appear to be more Gaussian (\emph{i.e.}, the
magnitude of the high-order cumulants is reduced and the entropy is increased). 
Thus, a suitable  statistical test that measures the level of non-Gaussianity of the 
residuals can be used to determine the causal direction. In principle, one may be tempted to 
use statistical tests based on entropy or cumulant estimation. However, our experiments show 
that one can obtain better results by using a test based on an \emph{energy distance} to 
quantify the Gaussianity of the residuals. In particular, entropy estimation is an arguably difficult
task and the estimators of the cumulants involve high-order moments which can lead to high variance.

The effectiveness of the proposed method GR-AN has been illustrated in both synthetic and real-world 
causal inference problems. We have shown that in certain problems GR-AN is competitive with state-of-the-art 
methods and that it performs better than related methods based on entropy
estimation \citep{HyvarinenS13}. The entropy can be understood as a measure of non-Gaussianity. 
Nevertheless, it is very difficult to estimate in practice. 
By contrast, the statistical test 
employed by GR-AR is not directly related to entropy estimation. This may explain the improvements observed. 
A limitation of the current formulation of GR-AN is that the distributions of the cause 
and the effect have to be equal. In the case of continuous univariate variables 
finding a transformation to make this possible is straightforward. Additionally, our 
experiments show that such a transformation can be side-stepped in some cases without 
a deterioration in performance. In any case, further research is needed to 
extend this analysis to remove this restriction.

Finally, the performance of GR-AN on cause-effect pairs with
discretized values is rather poor.
We believe this is due to the fact that in this case, finding 
a transformation so that the cause and the effect are 
equally distributed is infeasible.
Furthermore, the discretization process probably has
a strong impact on the Gaussianity tests.  
Further evidence that make these observations more plausible is the
fact that discretization has also a strong negative impact 
in the performance of the methods based on entropy estimation.

\section*{Acknowledgments}

Daniel Hern\'andez-Lobato and Alberto Su\'arez gratefully acknowledges the use of the facilities of Centro de 
Computación Científica (CCC) at Universidad Aut\'onoma de Madrid. These authors also acknowledge
financial support from the Spanish Plan Nacional I+D+i, Grants TIN2013-42351-P and TIN2015-70308-REDT, 
and from Comunidad de Madrid, Grant S2013/ICE-2845 CASI-CAM-CM. 

\appendix

\section{}
\label{Appendix:A}

In this appendix we show that if $\mathcal{X}$ and $\mathcal{Y}$ follow the same distribution and they have 
been centered, then the determinant of the covariance matrix of the random variable corresponding to $\bm{\epsilon}_i$, 
denoted with $\text{Cov}(\bm{\epsilon}_i)$, coincides with the determinant of the covariance
matrix corresponding to the random variable $\tilde{\bm{\epsilon}}_i$, denoted with $\text{Cov}(\tilde{\bm{\epsilon}}_i)$. 

From the causal model, \emph{i.e.}, $\mathbf{y}_i = \mathbf{A} \mathbf{x}_i + \bm{\epsilon}_i$, we have that:
\begin{align}
\text{Cov}(\mathcal{Y}) &= \mathbf{A} \text{Cov}(\mathcal{X}) \mathbf{A}^\text{T} + \text{Cov}(\bm{\epsilon}_i)\,.
\end{align}
Since $\mathcal{X}$ and $\mathcal{Y}$ follow the same distribution we have that $\text{Cov}(\mathcal{Y})=\text{Cov}(\mathcal{X})$.
Furthermore, we know from the causal model that $\mathbf{A}=\text{Cov}(\mathcal{Y},\mathcal{X})\text{Cov}(\mathcal{X})^{-1}$.
Then,
\begin{align}
\text{Cov}(\bm{\epsilon}_i) & = 
\text{Cov}(\mathcal{X}) -  \text{Cov}(\mathcal{Y},\mathcal{X}) \text{Cov}(\mathcal{X})^{-1} \text{Cov}(\mathcal{X},\mathcal{Y}) \,.
\end{align}
In the case of $\tilde{\bm{\epsilon}}_i$ we know that the relation $\tilde{\bm{\epsilon}}_i = 
(\mathbf{I}-\tilde{\mathbf{A}} \mathbf{A})\mathbf{x}_i - \tilde{\mathbf{A}} \bm{\epsilon}_i$ must be satisfied, 
where $\tilde{\mathbf{A}}=\text{Cov}(\mathcal{X},\mathcal{Y})\text{Cov}(\mathcal{Y})^{-1}=
\text{Cov}(\mathcal{X},\mathcal{Y})\text{Cov}(\mathcal{X})^{-1}$.
Thus, we have that:
\begin{align}
\text{Cov}(\tilde{\bm{\epsilon}}_i) &= (\mathbf{I}-\tilde{\mathbf{A}} \mathbf{A})  
	\text{Cov}(\mathcal{X})
	(\mathbf{I}- \mathbf{A}^\text{T}\tilde{\mathbf{A}}^\text{T})  +  \tilde{\mathbf{A}} \text{Cov}(\bm{\epsilon}_i) 
	\tilde{\mathbf{A}}^\text{T}
\nonumber \\
& = \text{Cov}(\mathcal{X}) - \text{Cov}(\mathcal{X}) \mathbf{A}^\text{T}\tilde{\mathbf{A}}^\text{T} - 
	 \tilde{\mathbf{A}}\mathbf{A} \text{Cov}(\mathcal{X})
	+   \tilde{\mathbf{A}}\mathbf{A}\text{Cov}(\mathcal{X}) \mathbf{A}^\text{T}\tilde{\mathbf{A}}^\text{T} 
\nonumber \\
& \quad +  \tilde{\mathbf{A}} \text{Cov}(\mathcal{X}) \tilde{\mathbf{A}}^\text{T} -  
	\tilde{\mathbf{A}}  \text{Cov}(\mathcal{Y},\mathcal{X}) \text{Cov}(\mathcal{X})^{-1} 
	\text{Cov}(\mathcal{X},\mathcal{Y}) \tilde{\mathbf{A}}^\text{T}
\nonumber \\
& = \text{Cov}(\mathcal{X}) - \text{Cov}(\mathcal{X}) \mathbf{A}^\text{T}\tilde{\mathbf{A}}^\text{T} - 
	 \tilde{\mathbf{A}}\mathbf{A} \text{Cov}(\mathcal{X})
+  \tilde{\mathbf{A}} \text{Cov}(\mathcal{X}) \tilde{\mathbf{A}}^\text{T}
\nonumber \\
& = \text{Cov}(\mathcal{X}) - \text{Cov}(\mathcal{X},\mathcal{Y}) \text{Cov}{\mathcal{X}}^{-1}  \text{Cov}(\mathcal{Y},\mathcal{X}) 
 - \text{Cov}(\mathcal{X},\mathcal{Y}) \text{Cov}{\mathcal{X}}^{-1}  \text{Cov}(\mathcal{Y},\mathcal{X}) 
\nonumber \\
& \quad
+  \text{Cov}(\mathcal{X},\mathcal{Y}) \text{Cov}{\mathcal{X}}^{-1}  \text{Cov}(\mathcal{Y},\mathcal{X}) 
\nonumber \\
& = \text{Cov}(\mathcal{X}) - \text{Cov}(\mathcal{X},\mathcal{Y}) \text{Cov}(\mathcal{X})^{-1}  \text{Cov}(\mathcal{Y},\mathcal{X}) \,.
\end{align}
By the matrix determinant theorem we have that 
$\text{det} \text{Cov}(\tilde{\bm{\epsilon}}_i)= \text{det} \text{Cov}(\bm{\epsilon}_i)$.
See \citep[p. 117]{Murphy2012} for further details.

\section{}
\label{Appendix:B}

In this Appendix we motivate that, if the distribution of the residuals is not Gaussian, but is close to Gaussian,
one should also expected more Gaussian residuals in the anti-causal direction in terms of the energy distance 
described in Section \ref{sec:test}. For simplicity we will consider the univariate case.  We use the fact that 
the energy distance in the one-dimensional case is the squared distance between the cumulative distribution 
functions of the residuals and a Gaussian distribution \citep{szekely2013energy}.  Thus,
\begin{align}
\tilde{D}^2 &= \int_{-\infty}^\infty \left[\tilde{F}(x) - \Phi(x)\right]^2 d x\,,
& 
D^2 &= \int_{-\infty}^\infty \left[F(x) - \Phi(x)\right]^2 d x\,,
\end{align}
where $\tilde{D}^2$ and $D^2$ are the energy distances to the Gaussian distribution in the anti-causal and the causal direction
respectively; $\tilde{F}(x)$ and $F(x)$ are the c.d.f. of the residuals in the anti-causal and the causal direction, respectively;
and finally, $\Phi(x)$ is the c.d.f. of a standard Gaussian.

One should expect that $\tilde{D}^2 \leq D^2$. To motivate this, we use the Gram-Charlier
series and compute an expansion of $\tilde{F}(x)$ and $F(x)$ around the standard Gaussian distribution 
\citep{patel1996handbook}. Such an expansion only converges in the case of distributions that are
close to be Gaussian (see Section 17.6.6a of \citep{cramer1946} for further details). Namely,
\begin{align}
\tilde{F}(x) & = \Phi(x) - \phi(x) \left( \frac{\tilde{a}_3}{3!} H_2(x) + \frac{\tilde{a}_4}{4!} H_3(x) + \cdots \right)\,,
\nonumber \\
F(x) & = \Phi(x) - \phi(x) \left( \frac{a_3}{3!} H_2(x) + \frac{a_4}{4!} H_3(x) + \cdots \right)\,,
\end{align}
where $\phi(x)$ is the p.d.f. of a standard Gaussian, $H_n(x)$ are Hermite polynomials and $\tilde{a}_n$ and $a_n$ are
coefficients that depend on the cumulants, \emph{e.g.}, $a_3 = \kappa_3$, $a_4 = \kappa_4$, $\tilde{a}_3=\tilde{\kappa}_3$,
$\tilde{a}_4=\tilde{\kappa}_4$. Note, however, that coefficients $a_n$ and $\tilde{a}_n$ for $n>5$ depend on 
combinations of the cumulants. Using such an expansion we find:
\begin{align}
\tilde{D}^2 &= \int_{-\infty}^\infty \phi(x)^2\left[ - \sum_{n=3}^{\infty} \frac{\tilde{a}_n}{n!} H_{n-1}(x) \right]^2 d x
\approx \int_{-\infty}^\infty \phi(x)^2\left[ - \sum_{n=3}^4 \frac{\tilde{\kappa}_n}{n!} H_{n-1}(x) \right]^2 d x \nonumber \\
& \approx  \frac{\tilde{\kappa}_3^2}{36} \mathds{E}[H_2(x)^2\phi(x)] + \frac{\tilde{\kappa}_4^2}{576} \mathds{E}[H_3(x)^2\phi(x)] \,,
\end{align}
where $\mathds{E}[\cdot]$ denotes expectation with respect to a standard Gaussian and we have truncated the 
Gram-Charlier expansion after $n=4$. Truncation of the Gram-Charlier expansion after $n=4$ is a standard procedure 
that is often done in the ICA literature for entropy approximation. See for example Section 5.5.1 of 
\citep{hyvarinen2004independent}. We have also used the fact that $\mathds{E}[H_3(x)H_2(x)\phi(x)] = 0$. The 
same approach can be followed in the case of $D^2$, the energy distance in the causal direction. The consequence is that 
$D^2\approx\kappa_3^2/36 \cdot \mathds{E}[H_2(x)^2\phi(x)] + \kappa_4^2/576 \cdot \mathds{E}[H_3(x)^2\phi(x)]$.
Finally, the fact that one should expect $\tilde{D}^2\leq D^2$ is obtained by noting that $\tilde{\kappa}_n = c_n \kappa_n$, where $c_n$ is 
some constant that lies in the interval $(-1,1)$, as indicated in Section \ref{sec:asym_univa}.
We expect that this result extends to the multivariate case.

\section{}
\label{Appendix:C}

In this Appendix we motivate that one should expect also more Gaussian residuals in the anti-causal direction,
based on a reduction of the cumulants, when the residuals in feature space are projected onto the first principal component. 
That is, when they are  multiplied by the first eigenvector of the covariance matrix of the residuals, and scaled by the corresponding eigenvalue.
Recall from Section \ref{sec:mul-cum} that these covariance matrices are 
$\mathbf{C}=\mathbf{I} - \mathbf{A}\mathbf{A}^\text{T} $ and $\tilde{\mathbf{C}}=\mathbf{I} - \mathbf{A}^\text{T} \mathbf{A}$,
in the causal and anti-causal direction respectively. Note that both matrices have the same eigenvalues. 

If $\mathbf{A}$ is symmetric we have that both $\mathbf{C}$ and $\tilde{\mathbf{C}}$ have the same matrix of 
eigenvectors $\mathbf{P}$. Let $\mathbf{p}_1^n$ be the first eigenvector multiplied $n$ times using the Kronecker
product. The cumulants in the anti-causal and the causal direction, after projecting the data onto the first eigenvector
are $\tilde{\kappa}_n^\text{proj} = (\mathbf{p}_1^\text{n})^\text{T} \mathbf{M}_n \text{vect}(\tilde{\kappa}_n) =
c(\mathbf{p}_1^\text{n})^\text{T} \text{vect}(\tilde{\kappa}_n)$ and
$\kappa_n^\text{proj} = (\mathbf{p}_1^\text{n})^\text{T} \text{vect}(\kappa_n)$, respectively, 
where $\mathbf{M}_n$ is the matrix that
relates the cumulants in the causal and the anti-causal direction (see Section \ref{sec:mul-cum}) and $c$
is one of the eigenvalues of $\mathbf{M}_n$. In particular, if $\mathbf{A}$ is symmetric, it is not difficult to show 
that $\mathbf{p}_1^\text{n}$ is one of the eigenvectors of $\mathbf{M}_n$. Furthermore, we also showed in that case that 
$||\mathbf{M}_n||_\text{op} < 1$ for $n\geq 3$ (see Section \ref{sec:mul-cum}). The consequence is that $c \in (-1,1)$,
which combined with the fact that $||\mathbf{p}_1^n||=1$ leads to smaller cumulants in magnitude in the case of the 
projected residuals in the anti-causal direction.

If $\mathbf{A}$ is not symmetric we motivate that one should also expect more Gaussian residuals in 
the anti-causal direction due to a reduction in the magnitude of the cumulants. For this, we derive 
a smaller upper bound on their magnitude. 
This smaller upper bound is based on an argument that uses the operator norm of vectors.

\begin{definition}
The operator norm of a vector $\mathbf{w}$ induced by the $\ell_p$ norm is
$||\mathbf{w}||_\text{op}=\text{min}\{c \geq 0 : ||\mathbf{w}^\text{T}\mathbf{v}||_p \leq c 
||\mathbf{v}||_p, \forall \mathbf{v} \}$.
\end{definition}

The consequence is that $||\mathbf{w}||_\text{op} \geq ||\mathbf{w}^\text{T} \mathbf{v}||_p / ||\mathbf{v}||_p$, $\forall \mathbf{v}$.
Thus, the smallest the operator norm of $\mathbf{w}$, the smallest the expected value obtained when
multiplying any vector by the vector $\mathbf{w}$. Furthermore, it is clear that 
$||\mathbf{w}||_\text{op}=||\mathbf{w}||_2$, in the case of the $\ell_2$-norm.  From the previous paragraph, in the anti-causal 
direction we have $||\tilde{\kappa}_n^\text{proj}||_2= ||(\tilde{\mathbf{p}}_1^n)^\text{T} \mathbf{M}_n \text{vect}(\kappa_n)||_2$, 
where $\tilde{\mathbf{p}}_1$ is the first eigenvector of $\tilde{\mathbf{C}}$,
while in the causal direction we have $||\kappa_n^\text{proj}||_2= ||(\mathbf{p}_1^n)^\text{T} \text{vect}(\kappa_n)||_2$,
where $\mathbf{p}_1$ is the first eigenvector of $\mathbf{C}$.
Thus, because the norm of each vector $\tilde{\mathbf{p}}_1^n$ and $\mathbf{p}_1^n$ is one, we have that
$||\mathbf{p}_1^n||_\text{op}=1$. However, because we expect $\mathbf{M}_n$, to reduce the norm of
$(\tilde{\mathbf{p}}_1^n)^\text{T}$, as motivated in Section \ref{sec:mul-cum}, $||(\tilde{\mathbf{p}}_1^n)^\text{T}\mathbf{M}_n||_\text{op} < 1$
should follow. This is expected to lead to smaller cumulants in magnitude in the anti-causal direction.

\bibliography{references}

\begin{thebibliography}{33}
\providecommand{\natexlab}[1]{#1}
\providecommand{\url}[1]{\texttt{#1}}
\expandafter\ifx\csname urlstyle\endcsname\relax
  \providecommand{\doi}[1]{doi: #1}\else
  \providecommand{\doi}{doi: \begingroup \urlstyle{rm}\Url}\fi

\bibitem[Beirlant et~al.(1997)Beirlant, Dudewicz, Gy\"{o}rfi, and {Van Der
  Meulen}]{beirlant++_1997_nonparametric}
J.~Beirlant, E.~J. Dudewicz, L.~Gy\"{o}rfi, and E.~C. {Van Der Meulen}.
\newblock {Nonparametric entropy estimation: An overview}.
\newblock \emph{International Journal of Mathematical and Statistical
  Sciences}, 6\penalty0 (1):\penalty0 17--39, 1997.

\bibitem[Chen et~al.(2013)Chen, Zhang, and Chan]{chen2013nonlinear}
Zhitang Chen, Kun Zhang, and Laiwan Chan.
\newblock Nonlinear causal discovery for high dimensional data: A kernelized
  trace method.
\newblock In \emph{IEEE 13th International Conference on Data Mining}, pages
  1003--1008, 2013.

\bibitem[Chen et~al.(2014)Chen, Zhang, Chan, and Sch{\"o}lkopf]{ChenZCS14}
Zhitang Chen, Kun Zhang, Laiwan Chan, and Bernhard Sch{\"o}lkopf.
\newblock Causal discovery via reproducing kernel {H}ilbert space embeddings.
\newblock \emph{Neural Computation}, 26\penalty0 (7):\penalty0 1484--1517,
  2014.

\bibitem[Cornish and Fisher(1938)]{cornish+fisher_1938_moments}
E.~A. Cornish and R.~A. Fisher.
\newblock Moments and cumulants in the specification of distributions.
\newblock \emph{Revue de l'Institut International de Statistique / Review of
  the International Statistical Institute}, 5\penalty0 (4):\penalty0 307--320,
  1938.

\bibitem[Cram{\'e}r(1946)]{cramer1946}
Harald Cram{\'e}r.
\newblock \emph{Mathematical methods of statistics}.
\newblock PhD thesis, 1946.

\bibitem[Entner and Hoyer(2012)]{entner_hoyer2012}
Doris Entner and PatrikO. Hoyer.
\newblock Estimating a causal order among groups of variables in linear models.
\newblock In \emph{Artificial Neural Networks and Machine Learning – ICANN
  2012}, pages 84--91. 2012.

\bibitem[Gretton et~al.(2008)Gretton, Fukumizu, Teo, Song, Sch\"{o}lkopf, and
  Smola]{NIPS2007}
Arthur Gretton, Kenji Fukumizu, Choon~H. Teo, Le~Song, Bernhard Sch\"{o}lkopf,
  and Alex~J. Smola.
\newblock A kernel statistical test of independence.
\newblock In J.C. Platt, D.~Koller, Y.~Singer, and S.T. Roweis, editors,
  \emph{Advances in Neural Information Processing Systems 20}, pages 585--592.
  2008.

\bibitem[Gretton et~al.(2012)Gretton, Borgwardt, Rasch, Sch\"{o}lkopf, and
  Smola]{Gretton2012}
Arthur Gretton, Karsten~M. Borgwardt, Malte~J. Rasch, Bernhard Sch\"{o}lkopf,
  and Alexander Smola.
\newblock A kernel two-sample test.
\newblock \emph{Journal of Machine Learning Research}, 13:\penalty0 723--773,
  2012.

\bibitem[Hern{\'a}ndez-Lobato et~al.(2011)Hern{\'a}ndez-Lobato,
  Morales-Mombiela, and Su{\'a}rez]{hernandezLobato11}
Jos{\'e}~Miguel Hern{\'a}ndez-Lobato, Pablo Morales-Mombiela, and Alberto
  Su{\'a}rez.
\newblock {G}aussianity measures for detecting the direction of causal time
  series.
\newblock In \emph{Proceedings of the 22nd International Joint Conference on
  Artificial Intelligence}, pages 1318--1323, 2011.

\bibitem[Hoyer et~al.(2009)Hoyer, Janzing, Mooij, Peters, and
  Sch\"{o}lkopf]{Hoyeretal08}
Patrik~O. Hoyer, Dominik Janzing, Joris~M. Mooij, Jonas Peters, and Bernhard
  Sch\"{o}lkopf.
\newblock Nonlinear causal discovery with additive noise models.
\newblock In D.~Koller, D.~Schuurmans, Y.~Bengio, and L.~Bottou, editors,
  \emph{{A}dvances in {N}eural {I}nformation {P}rocessing {S}ystems 21}, pages
  689--696, 2009.

\bibitem[Hyv\"{a}rinen(1998)]{Hyvarinen1998}
Aapo Hyv\"{a}rinen.
\newblock New approximations of differential entropy for independent component
  analysis and projection pursuit.
\newblock In \emph{Proceedings of the 1997 Conference on Advances in Neural
  Information Processing Systems 10}, pages 273--279. MIT Press, 1998.

\bibitem[Hyv{\"{a}}rinen and Smith(2013)]{HyvarinenS13}
Aapo Hyv{\"{a}}rinen and Stephen~M. Smith.
\newblock Pairwise likelihood ratios for estimation of non-gaussian structural
  equation models.
\newblock \emph{Journal of Machine Learning Research}, 14\penalty0
  (1):\penalty0 111--152, 2013.

\bibitem[Hyv{\"a}rinen et~al.(2004)Hyv{\"a}rinen, Karhunen, and
  Oja]{hyvarinen2004independent}
Aapo Hyv{\"a}rinen, Juha Karhunen, and Erkki Oja.
\newblock \emph{Independent component analysis}, volume~46.
\newblock John Wiley \& Sons, 2004.

\bibitem[Janzing et~al.(2010)Janzing, Hoyer, and Sch{\"o}lkopf]{icml2010}
Dominik Janzing, Patrik~O. Hoyer, and Bernhard Sch{\"o}lkopf.
\newblock Telling cause from effect based on high-dimensional observations.
\newblock In Johannes F{\"u}rnkranz and Thorsten Joachims, editors,
  \emph{Proceedings of the 27th International Conference on Machine Learning
  (ICML-10)}, pages 479--486, 2010.

\bibitem[Janzing et~al.(2012)Janzing, Mooij, Zhang, Lemeire, Zscheischler,
  Daniu{\v s}is, Steudel, and Sch{\"o}lkopf]{Janzing12}
Dominik Janzing, Joris~M. Mooij, Kun Zhang, Jan Lemeire, Jacob Zscheischler,
  Povilas Daniu{\v s}is, Bastian Steudel, and Bernhard Sch{\"o}lkopf.
\newblock Information-geometric approach to inferring causal directions.
\newblock \emph{Artificial Intelligence}, 182-183:\penalty0 1--31, 2012.

\bibitem[Kawahara et~al.(2012)Kawahara, Bollen, Shimizu, and
  Washio]{Kawahara2010}
Yoshinobu Kawahara, Kenneth Bollen, Shohei Shimizu, and Takashi Washio.
\newblock {GroupLiNGAM}: Linear non-gaussian acyclic models for sets of
  variables.
\newblock 2012.
\newblock arXiv:1006.5041.

\bibitem[Kpotufe et~al.(2014)Kpotufe, Sgouritsa, Janzing, and
  Sch{\"{o}}lkopf]{KpotufeSJS14}
Samory Kpotufe, Eleni Sgouritsa, Dominik Janzing, and Bernhard Sch{\"{o}}lkopf.
\newblock Consistency of causal inference under the additive noise model.
\newblock In \emph{Proceedings of the 31th International Conference on Machine
  Learning, {ICML} 2014, Beijing, China, 21-26 June 2014}, pages 478--486,
  2014.

\bibitem[Laub(2004)]{Laub2004}
Alan~J. Laub.
\newblock \emph{Matrix Analysis For Scientists And Engineers}.
\newblock Society for Industrial and Applied Mathematics, Philadelphia, PA,
  USA, 2004.
\newblock ISBN 0898715768.

\bibitem[Marcinkiewicz(1938)]{marcinkiewicz_1938_propriete}
J.~T. Marcinkiewicz.
\newblock Sur une propri{\'e}t{\'e} de la loi de gauss.
\newblock \emph{Mathematische Zeitschrift}, 44:\penalty0 612--618, 1938.

\bibitem[McCullagh(1987)]{mccullagh87}
{Peter} McCullagh.
\newblock \emph{Tensor methods in statistics}.
\newblock Chapman and Hall, 1987.

\bibitem[Mooij et~al.(2010)Mooij, Stegle, Janzing, Zhang, and
  Sch\"{o}lkopf]{NIPS2010}
Joris~M. Mooij, Oliver Stegle, Dominik Janzing, Kun Zhang, and Bernhard
  Sch\"{o}lkopf.
\newblock Probabilistic latent variable models for distinguishing between cause
  and effect.
\newblock In J.D. Lafferty, C.K.I. Williams, J.~Shawe-Taylor, R.S. Zemel, and
  A.~Culotta, editors, \emph{Advances in Neural Information Processing Systems
  23}, pages 1687--1695. 2010.

\bibitem[Morales-Mombiela et~al.(2013)Morales-Mombiela, Hern{\'a}ndez-Lobato,
  and Su{\'a}rez]{morales13}
Pablo Morales-Mombiela, Daniel Hern{\'a}ndez-Lobato, and Alberto Su{\'a}rez.
\newblock Statistical tests for the detection of the arrow of time in vector
  autoregressive models.
\newblock In \emph{Proceedings of the 23rd International Joint Conference on
  Artificial Intelligence}, 2013.

\bibitem[Murphy(2012)]{Murphy2012}
Kevin Murphy.
\newblock \emph{Machine Learning: a Probabilistic Perspective}.
\newblock The MIT Press, 2012.

\bibitem[Patel and Read(1996)]{patel1996handbook}
Jagdish~K Patel and Campbell~B Read.
\newblock \emph{Handbook of the normal distribution}, volume 150.
\newblock CRC Press, 1996.

\bibitem[Pearl(2000)]{Pearl2000}
Judea Pearl.
\newblock \emph{Causality: Models, Reasoning, and Inference}.
\newblock Cambridge University Press, New York, NY, USA, 2000.
\newblock ISBN 0-521-77362-8.

\bibitem[Sch{\"o}lkopf and Smola(2002)]{Scholkopf2002}
Bernhard Sch{\"o}lkopf and Alexander~J. Smola.
\newblock \emph{Learning with Kernels: Support Vector Machines, Regularization,
  Optimization, and Beyond}.
\newblock MIT Press, Cambridge, MA, USA, 2002.
\newblock ISBN 0262194759.

\bibitem[Sch{\"o}lkopf et~al.(1997)Sch{\"o}lkopf, Smola, and
  M{\"u}ller]{scholkopf1997kernel}
Bernhard Sch{\"o}lkopf, Alexander Smola, and Klaus-Robert M{\"u}ller.
\newblock Kernel principal component analysis.
\newblock In \emph{International Conference on Artificial Neural Networks},
  pages 583--588. Springer, 1997.

\bibitem[Shimizu et~al.(2006)Shimizu, Hoyer, Hyv\"{a}rinen, and
  Kerminen]{Shimizu06}
Shohei Shimizu, Patrik~O. Hoyer, Aapo Hyv\"{a}rinen, and Antti Kerminen.
\newblock A linear non-gaussian acyclic model for causal discovery.
\newblock \emph{Journal of Machine Learning Research}, 7:\penalty0 2003--2030,
  2006.

\bibitem[Singh et~al.(2003)Singh, Misra, Hnizdo, Fedorowicz, and
  Demchuk]{singh2003nearest}
Harshinder Singh, Neeraj Misra, Vladimir Hnizdo, Adam Fedorowicz, and Eugene
  Demchuk.
\newblock Nearest neighbor estimates of entropy.
\newblock \emph{American journal of mathematical and management sciences},
  23\penalty0 (3-4):\penalty0 301--321, 2003.

\bibitem[Song et~al.(2013)Song, Fukumizu, and Gretton]{song2013}
Le~Song, K.~Fukumizu, and A~Gretton.
\newblock Kernel embeddings of conditional distributions: A unified kernel
  framework for nonparametric inference in graphical models.
\newblock \emph{Signal Processing Magazine, IEEE}, 30:\penalty0 98--111, 2013.

\bibitem[Sz\'ekely and Rizzo(2005)]{Szekely2005}
G\'abor~J. Sz\'ekely and Maria~L. Rizzo.
\newblock A new test for multivariate normality.
\newblock \emph{Journal of Multivariate Analysis}, 93\penalty0 (1):\penalty0
  58--80, 2005.

\bibitem[Sz{\'e}kely and Rizzo(2013)]{szekely2013energy}
G{\'a}bor~J Sz{\'e}kely and Maria~L Rizzo.
\newblock Energy statistics: A class of statistics based on distances.
\newblock \emph{Journal of Statistical Planning and Inference}, 143\penalty0
  (8):\penalty0 1249--1272, 2013.

\bibitem[Zhang and Hyv{\"a}rinen(2009)]{zhang2009}
Kun Zhang and Aapo Hyv{\"a}rinen.
\newblock On the identifiability of the post-nonlinear causal model.
\newblock In \emph{Proceedings of the International Conference on Uncertainty
  in Artificial Intelligence}, pages 647--655, 2009.

\end{thebibliography}
\end{document}